\documentclass[10pt,twocolumn,letterpaper]{article}

\usepackage[pagenumbers]{cvpr} % To force page numbers, e.g. for an arXiv version

\newlength\mytmplen
\usepackage{tikz}          % for TikZ drawing
\usetikzlibrary{spy}       % for the "spy" library used in zoomin

\usepackage{caption}       % for \captionof
\DeclareRobustCommand{\zoomin}[9]{ %
\begin{tikzpicture}[spy using outlines={rectangle,#9,magnification=#8,size=#6}]   
	\node[anchor=south west,inner sep=0]  {\includegraphics[width=#7]{#1}};
	\spy on (#2, #3) in node at (#4,#5);
\end{tikzpicture}
}

\usepackage{cuted}   % for full-width strips in two-column
\usepackage{graphicx}      % for \resizebox
\usepackage{booktabs}      % for \midrule
\usepackage[table]{xcolor} % for \cellcolor (and named colors)
\usepackage{colortbl}      % optional, also provides \cellcolor
\usepackage{mwe} 
\usepackage{comment}
\newcommand{\best}{\cellcolor{tablered}}

\definecolor{tablered}{rgb}{1, 0.7, 0.7}

\usepackage{soul}
\setuldepth{foobar}

\usepackage{xspace}
\providecommand{\methodname}{MeshSplatting\xspace}

\usepackage[table]{xcolor} % for \cellcolor
\usepackage{pifont}        % for \ding
\usepackage{booktabs}
\newcommand{\cmark}{\ding{51}}  % ✓
\newcommand{\xmark}{\ding{55}}  % ✗

\usepackage{subcaption} % gives (a), (b), ...
\renewcommand{\paragraph}[1]{\vspace{.25em}\noindent\textbf{#1}.}

\usepackage{makecell}

\definecolor{cvprblue}{rgb}{0.21,0.49,0.74}
\usepackage[pagebackref,breaklinks,colorlinks,allcolors=cvprblue]{hyperref}

\title{MeshSplatting: Differentiable Rendering with Opaque Meshes}

\author{
{Jan Held}$^{1,2}$\quad
{Sanghyun Son}$^{3}$\quad 
{Renaud Vandeghen}$^{1}$\quad \\
{Daniel Rebain}$^{4}$\quad 
{Matheus Gadelha}$^{6}$\quad
{Yi Zhou}$^{6}$\quad 
{Anthony Cioppa}$^{1}$\quad 
\\
{Ming C. Lin}$^{3}$\quad
{Marc Van Droogenbroeck}$^{1}$\quad
{Andrea Tagliasacchi}$^{2,5}$ \\[0.5em]
$^1$University of Liège \quad
$^2$Simon Fraser University \quad
$^3$University of Maryland \quad \\
$^4$University of British Columbia \quad
$^5$University of Toronto \quad
$^6$Adobe Research \\
}

\begin{document}
\maketitle
\begin{strip}
\vspace*{-4em}
\centering
\includegraphics[width=\linewidth]{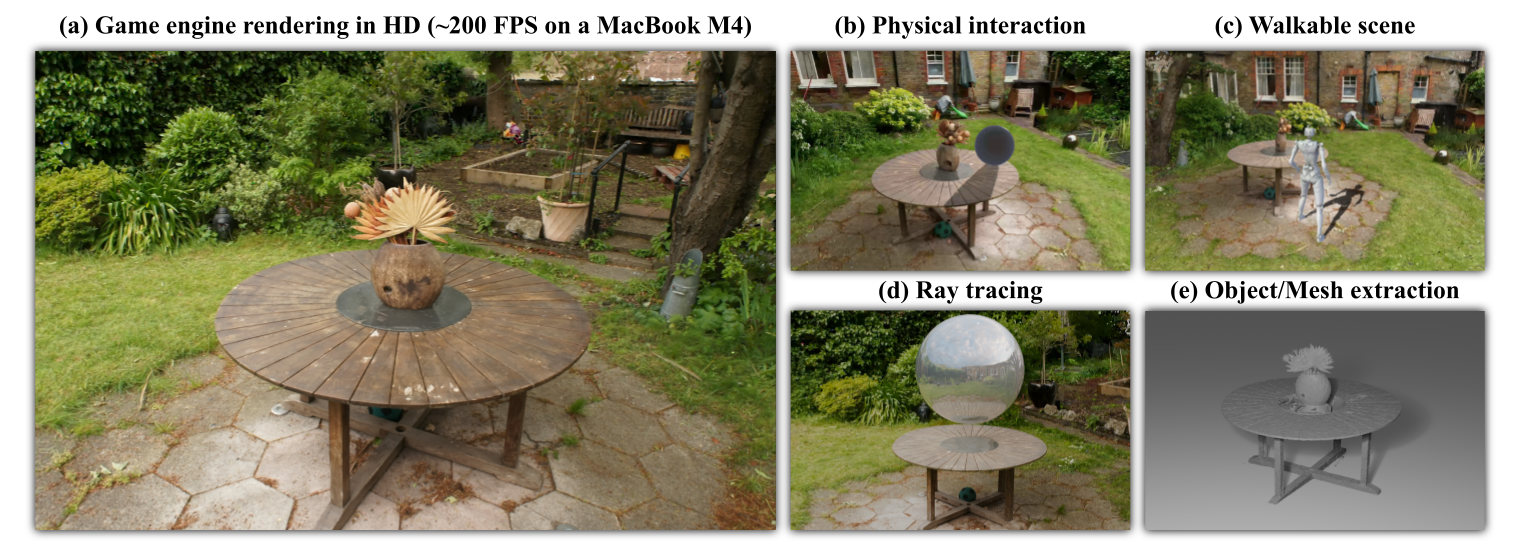}
\vspace*{-2em}
\captionof{figure}{
\textbf{\methodname} produces a \textit{connected mesh} composed only of \textit{opaque triangles}, achieving high-quality novel view synthesis through end-to-end optimization, with a $2\times$ training speed-up and $2\times$ lower memory usage over current state-of-the-art methods. %of under 50 minutes. 
(a) Our representation is compatible with standard game engines, requiring no a-posteriori conversion and/or custom rendering routines for transparency, and natively supports (b) physical interactions, (c) interactive walkthroughs, and (d) ray tracing.
(e) \methodname enables straightforward object extraction, allowing scene elements to be directly exported and imported into game engines.
}
\label{fig:teaser}
\vspace*{-1em}
\end{strip}

\vspace*{-2em}
\begin{center}
{\bf\large Abstract} 
\vspace*{-0.1em}
\end{center} 
Primitive-based splatting methods like 3D Gaussian Splatting have revolutionized novel view synthesis with real-time rendering. However, their point-based representations remain incompatible with mesh-based pipelines that power AR/VR and game engines. We present MeshSplatting, a mesh-based reconstruction approach that jointly optimizes geometry and appearance through differentiable rendering. By enforcing connectivity via restricted Delaunay triangulation and refining surface consistency, MeshSplatting creates end-to-end smooth, visually high-quality meshes that render efficiently in real-time 3D engines. On Mip-NeRF360, it boosts PSNR by +0.69 dB over the current state-of-the-art MiLo for mesh-based novel view synthesis, while training 2× faster and using 2× less memory, bridging neural rendering and interactive 3D graphics for seamless real-time scene interaction. The project page is available at \url{https://meshsplatting.github.io/}.
\vspace*{0.25em}

\section{Introduction}

\label{sec:intro}
Recent advances in novel view synthesis, like 3D Gaussian Splatting~\cite{Kerbl20233DGaussian}, have enabled photo-realistic reconstruction of extremely complex scenes. 
3D Gaussian Splatting encodes scenes with millions of 3D Gaussian primitives, achieving real-time rendering while also training efficiently.
However, while 3DGS renders with high visual fidelity, Gaussian primitives are not immediately compatible with classical graphics pipelines used in simulators, games, and AR/VR applications, as these are typically based on \textit{polygonal meshes}.

\noindent Therefore, integrating 3DGS in classical pipelines either requires \textit{engineering} rendering engines \cite{KIRI3DGSRBlender, XSceneUEPlugin2024, Pranckevicius2023UnityGS} and simulators~\cite{Modi2024Simplicits} to support them. 
However, this is non-trivial since 3DGS relies on \textit{sorting} and \textit{alpha blending}, preventing the use of standard techniques like depth buffers and occlusion culling \cite{AkenineMoller2018RealTimeRendering, Hughes2014CGPP}.
Another line of work is \textit{converting} Gaussian radiance fields into meshes~\cite{Huang20242DGaussian, Zhang2024RaDeGS-arxiv, Guedon2025MILo-arxiv}.
Although somewhat effective, all of these conversion approaches rely on complex post-processing and typically lead to loss of visual quality, as the conversion step is non-differentiable.
\noindent Rather than relying on conversion a-posteriori, we consider optimizing a mesh directly by making the rasterization process differentiable.
Early approaches, such as \citet{Kato2018Neural} and \citet{Liu2019SoftRasterizer}, enabled differentiable optimization of solid polygonal surfaces, but they required very careful initialization for optimization to converge effectively and are mainly limited to object-level setups rather than large realistic scenes.

\noindent Instead, \citet{Held2025Triangle-arxiv} recently proposed to optimize (potentially transparent) triangles via volumetric rendering, therefore effectively replacing Gaussians with \textit{triangles}.
However, when their triangles are rendered in a game engine, a noticeable drop in visual quality occurs, as game engines assume triangles to be opaque.
Moreover, \citet{Held2025Triangle-arxiv} outputs a \textit{soup of triangles}, rather than a connected polygonal mesh, often needed for physics-based simulation.

\paragraph{\bf Key Contributions}
\noindent We introduce {\em \methodname} to address all the aforementioned limitations:
\begin{enumerate*}[label=\textbf{(\roman*)}]
\item an end-to-end optimization of mesh-based scene representations that retains visual quality while training $2\times$ faster than current state-of-the-art methods;
\item rather than a polygon soup, we generate a connected mesh by refining the vertex locations of a restricted Delaunay triangulation;
\item triangles are naturally connected to each other, and quantities stored within vertices are smoothly interpolated across each triangle;
\item the optimization is aware that the triangles should be opaque, and therefore allowing direct high-quality rendering in standard game engines (see \cref{fig:teaser}), opening the door for classical techniques like the use of depth buffers and occlusion culling~\cite{AkenineMoller2018RealTimeRendering, Hughes2014CGPP}.
\end{enumerate*}

\noindent \methodname achieves higher visual fidelity and captures finer geometric detail compared to modern \textit{mesh-based} novel-view synthesis approaches~\cite{Guedon2025MILo-arxiv, Huang20242DGaussian, Yu2024Gaussian, Zhang2024RaDeGS-arxiv}. 
It is the first method to reconstruct large-scale real-world meshes end-to-end, directly producing connected, opaque, and colored triangle meshes \textit{without} post-hoc extraction.
Our representation can be \textit{directly} imported into standard game engines, enabling a wide range of downstream applications including physics-based simulation, interactive walkthroughs, ray tracing, and scene editing, some of which are illustrated in \Cref{fig:teaser}.

\section{Related work}
\label{sec:related_work}

Differentiable rendering enables end-to-end optimization by propagating image-based losses back to scene parameters, allowing for the learning of explicit representations such as point clouds~\cite{Gross2007Point,Kato2018Neural}, voxel grids~\cite{FridovichKeil2022Plenoxels}, polygonal meshes~\cite{Kato2018Neural, Liu2019SoftRasterizer, Loper2014OpenDR}, and more recently, Gaussian primitives~\cite{Kerbl20233DGaussian}.
The advent of 3D Gaussian Splatting~\cite{Kerbl20233DGaussian} showed that it is possible to fit millions of anisotropic Gaussians in minutes, enabling real-time rendering with high fidelity.
Since then, various directions have been explored to improve the Gaussian primitive, including the use of 2D Gaussians~\cite{Huang20242DGaussian}, generalized Gaussians~\cite{Hamdi2024GES, Taktasheva20253DGaussian-arxiv}, alternative kernels~\cite{Huang2025Deformable}, and learnable basis functions~\cite{Chen2024Beyond-arxiv}.
Other works moved beyond Gaussians entirely, investigating different primitives such as smooth 3D convexes~\cite{Held20253DConvex}, linear primitives~\cite{vonLutzow2025LinPrim-arxiv}, sparse voxel fields~\cite{Sun2025Sparse}, or radiance foams~\cite{Govindarajan2025Radiant-arxiv}.
More recently, \citet{Held2025Triangle-arxiv} advocated for the comeback of triangles, the most classical primitive in computer graphics. 
Several researchers have explored this direction, proposing triangle-based representations for efficient scene modeling~\cite{Burgdorfer2025Radiant-arxiv, Jiang2025HaloGS-arxiv}.
However, existing triangle-based methods usually result in an unstructured \textit{triangle soup}, with no connectivity between adjacent triangles, and they fail to produce opaque triangles at the end of training, limiting usability in downstream applications.
We propose a method that enforces \textit{connectivity}, resulting in a mesh of opaque triangles directly compatible with game engines.

\paragraph{Mesh reconstruction from images}
Implicit and explicit methods have made significant progress in reconstructing 3D scenes, but they remain largely incompatible with traditional game engines that primarily rely on mesh-based rasterization with depth buffers.
Some methods propose strategies to convert implicit radiance fields into meshes. BakedSDF~\cite{Yariv2023BakedSDF} learns a neural signed distance field and appearance and then bakes them into a textured triangle mesh. Binary Opacity Fields~\cite{Reiser2024Binary} drives densities toward near binary opacities so surfaces can be extracted as a mesh, and MobileNeRF~\cite{Chen2023MobileNeRF} distills a NeRF into a compact set of textured polygons.
However, these methods introduce overhead and increase the overall training time.

Several methods have built upon 3DGS and proposed ways to extract a mesh from an optimized Gaussian scene. 2DGS~\cite{Huang20242DGaussian} and RaDe-GS~\cite{Zhang2024RaDeGS-arxiv} rely on Truncated Signed Distance Fields for mesh extraction. Other approaches extract meshes by sampling a surface-aligned Gaussian level set followed by Poisson reconstruction~\cite{Guedon2024SuGaR}, or by defining a Gaussian opacity level set and applying Marching Tetrahedra on Gaussian-induced tetrahedral grids~\cite{Yu2024Gaussian}. All of these, however, treat mesh extraction as a \textit{separate} post-processing step, decoupled from the optimization process. 

More recently, MiLo~\cite{Guedon2025MILo-arxiv} integrates surface mesh extraction directly into the optimization, jointly refining both the mesh and the Gaussian representation. However, while MiLo optimizes the mesh geometry during training, color still has to be learned separately. 
Another line of work employs differentiable meshes for 3D reconstruction~\cite{Son2024DMesh, Son2024DMesh++-arxiv}, but these methods primarily target synthetic objects and do not generalize to real-world scenes.
In contrast, \methodname directly optimizes opaque triangles together with their vertex colors, making the result immediately compatible with any game engine without additional post-processing steps.

\section{Methodology}
\label{sec:methodology}
We now overview the key components of our method.
\Cref{sec:background} reviews Triangle Splatting~\cite{Held2025Triangle-arxiv}, which is used as volume-renderable primitives for differentiable rendering in this work.
We give an overview of our \methodname representation in \Cref{sec:vertex_based}, and describe the optimization stages to convert the triangle soup into a connected mesh in \Cref{sec:from_soup_to_mesh}.
For the representation to be \textit{natively} compatible with game engines, triangles need to be \textit{opaque}, and the process to achieve this objective is detailed in \Cref{sec:towards_opaque}.
We conclude by detailing other optimization details in \Cref{sec:optimization}, such as densification/pruning, and training losses.

\subsection{Background}
\label{sec:background}
In Triangle Splatting~\cite{Held2025Triangle-arxiv}, each triangle $\mathbf{T_m}$ is defined by three vertices $\mathbf{v}_i{\in} \mathbb{R}^3$, a color $\mathbf{c_m}$, a smoothness parameter $\sigma_m$ and an opacity~$o_m$.
The rasterization process begins by projecting each 3D vertex $\mathbf{v}_i$ of a triangle $\mathbf{T_m}$ onto the image plane using a standard pinhole camera model\footnote{Unlike 3DGS, linearization of Gaussian projections is required~\cite{Zwicker2002EWASplatting}, since perspective projection preserves linearity, 3D triangles remain triangles in 2D screen space.}.
To determine the influence of a triangle on a pixel $\mathbf{p}$ and make the splatting process differentiable,
the \emph{signed distance field} $\phi$ of the 2D triangle in image space is defined as:
\begin{equation}
\phi(\mathbf{p}) = \max_{i\in \{1,2,3\}} L_i(\mathbf{p}),
\quad
L_i(\mathbf{p}) = \mathbf{n}_i \cdot \mathbf{p} + d_i,
\end{equation}
where $\mathbf{n}_i$ are the unit normals of the triangle edges pointing outside the triangle, and $d_i$ are offsets such that the triangle is given by the zero-level set of the function $\phi$.
The signed distance field $\phi$ thus takes positive values outside the triangle, negative values inside, and equals zero on its boundary.
The window function $I$ is then defined as:
\begin{equation}
\label{eq:indicator1}
\begin{aligned}
I(\mathbf{p}) &= \left(\text{ReLU}\left( \frac{\phi(\mathbf{p})}{\phi(\mathbf{s})} \right)\right)^{\sigma} \\
\end{aligned}
\end{equation}
with $\mathbf{s}{\in}\mathbb{R}^2$ be the \emph{incenter} of the projected triangle (\ie, the point inside the triangle with minimum signed distance).
Note how the indicator evaluates to $1$ at the triangle incenter, $0$ at the boundary and $0$ outside the triangle.
$\sigma$ is a smoothness parameter that controls the transition between the incenter and boundary of the triangle.
More specifically, as $\sigma {\to} 0$, the representation converges to a solid triangle, while larger values of $\sigma$ yield a smooth window function that gradually increases from zero at the boundary to one at the center.
This triangle parameterization has two main limitations: triangles remain isolated without vertex sharing, and treating $\sigma$ and $o$ as independent free parameters prevents them from becoming fully opaque after training.

\begin{figure}[t]
\centering
\includegraphics[width=0.99\linewidth]{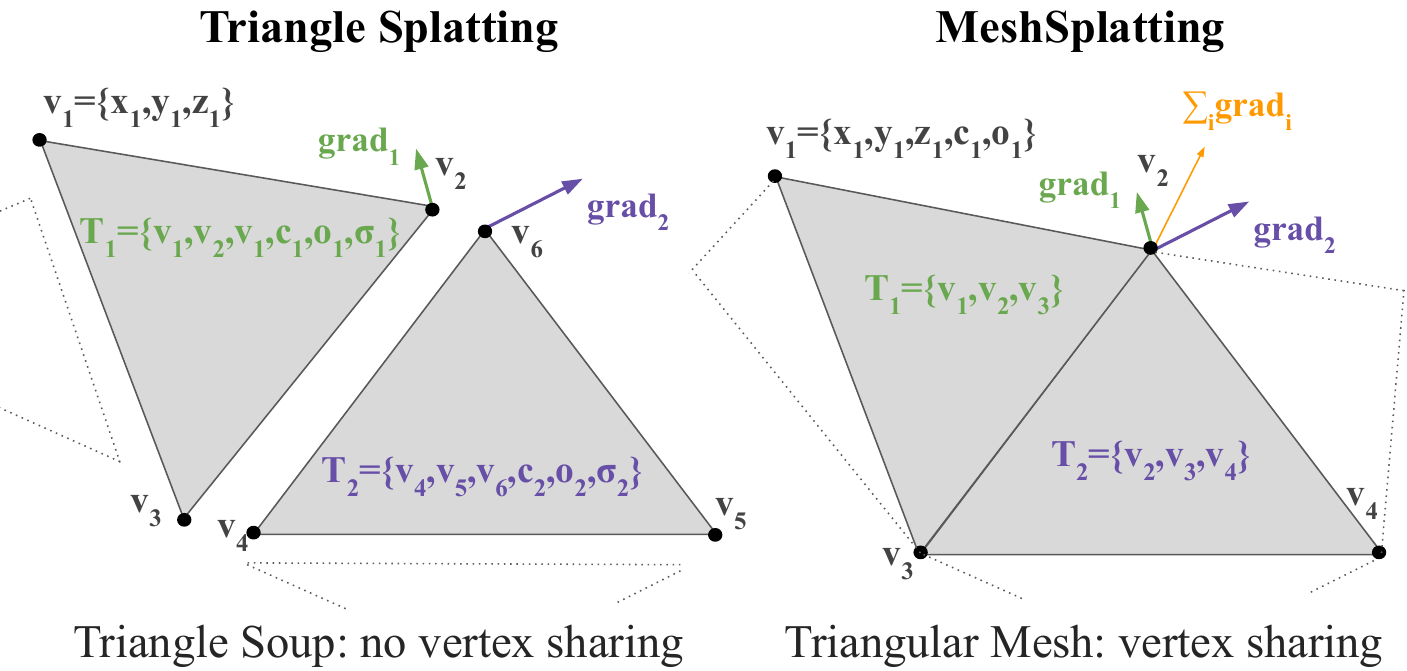}
\caption{\textbf{Mesh parametrization.}
(left) In a triangle soup, each triangle $\mathbf{T}_m$ is defined independently by three vertices $\mathbf{v}_i$, $\mathbf{v}_j$, $\mathbf{v}_k$, a color $\mathbf{c_m}$, a smoothness parameter~$\sigma_m$, and an opacity~$o_m$, without sharing vertices with neighboring triangles.
(right) \methodname parameterize a triangle $\mathbf{T}_m$ through a shared vertex set, where each vertex $\mathbf{v}_i$ stores $x_i, y_i, z_i$, $c_i$, and $o_i$. Each triangle is defined by the three indices in the vertex set that compose it. During the backward pass, gradients from all adjacent triangles are accumulated at shared vertices. The smoothness parameter~$\sigma$ is shared across all triangles.
}
\label{fig:methodology}
\end{figure}

\subsection{Vertex-sharing triangle
representation}
\label{sec:vertex_based}
In \methodname, we define our mesh vertices as
\begin{equation}
\mathcal{V} = \{\mathbf{v}_i \in \mathbb{R}^3 \mid i = 1,\dots,N\},
\end{equation}
with $N$ denoting their cardinality. 
Similarly to~\citet{Son2024DMesh++-arxiv} each vertex is parameterized as $\mathbf{v}_i = (x_i, y_i, z_i, c_i, o_i),$
where $(x_i, y_i, z_i) \in \mathbb{R}^3$ denotes its 3D position, 
$\mathbf{c}_i \in \mathbb{R}^3$ the vertex color and $o_i \in [0,1]$ the vertex opacity.\footnote{After training, the opacity parameter is \textit{discarded}, and all triangles are treated as fully opaque for compatibility with standard game engines.}
A triangle is defined by a triplet of vertices $\mathbf{T}_m = \{v_i,v_j,v_k\}$, its opacity is set to $o_{\mathbf{T}_m} = \min(o_i, o_j, o_k),$ and its color at a point inside the triangle is obtained by interpolating vertex colors with \textit{barycentric coordinates}. 
During differentiation, vertex positions, colors, and opacities therefore receive the accumulated gradients from all triangles connected to it, as shown in \Cref{fig:methodology} (right).

\begin{figure*}[t]
\vspace*{-0.5em}
\centering
\setlength{\mytmplen}{0.30\linewidth} % fit 4 images + label
\resizebox{\linewidth}{!}{ 
\begin{tabular}{c@{\hskip 0.01in}c@{\hskip 0.01in}c@{\hskip 0.01in}c@{\hskip 0.01in}c}
    % --- Stage headers ---
    & \multicolumn{2}{c}{\normalsize \textbf{Stage 1. Triangle soup optimization}} 
    & \multicolumn{2}{c}{\normalsize \textbf{Stage 2. Mesh creation \& refinement}} \\
   % --- Row 1 ---
   \rotatebox{90}{\parbox{3.5cm}{\centering \small \textbf{RGB}}}
& \zoomin{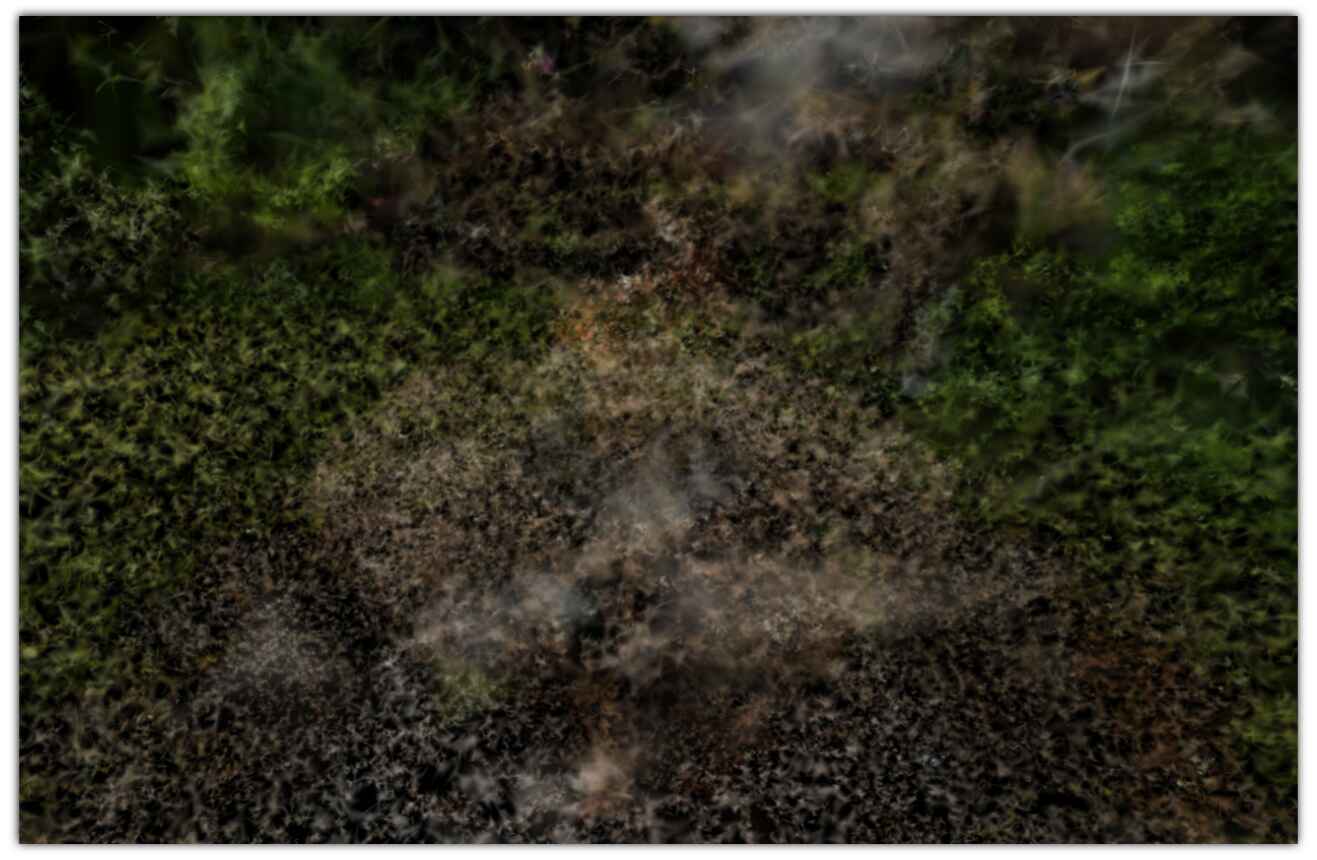}{0.55\mytmplen}{0.52\mytmplen}{0.824\mytmplen}{0.178\mytmplen}{1.70cm}{\mytmplen}{4.5}{red}
& \zoomin{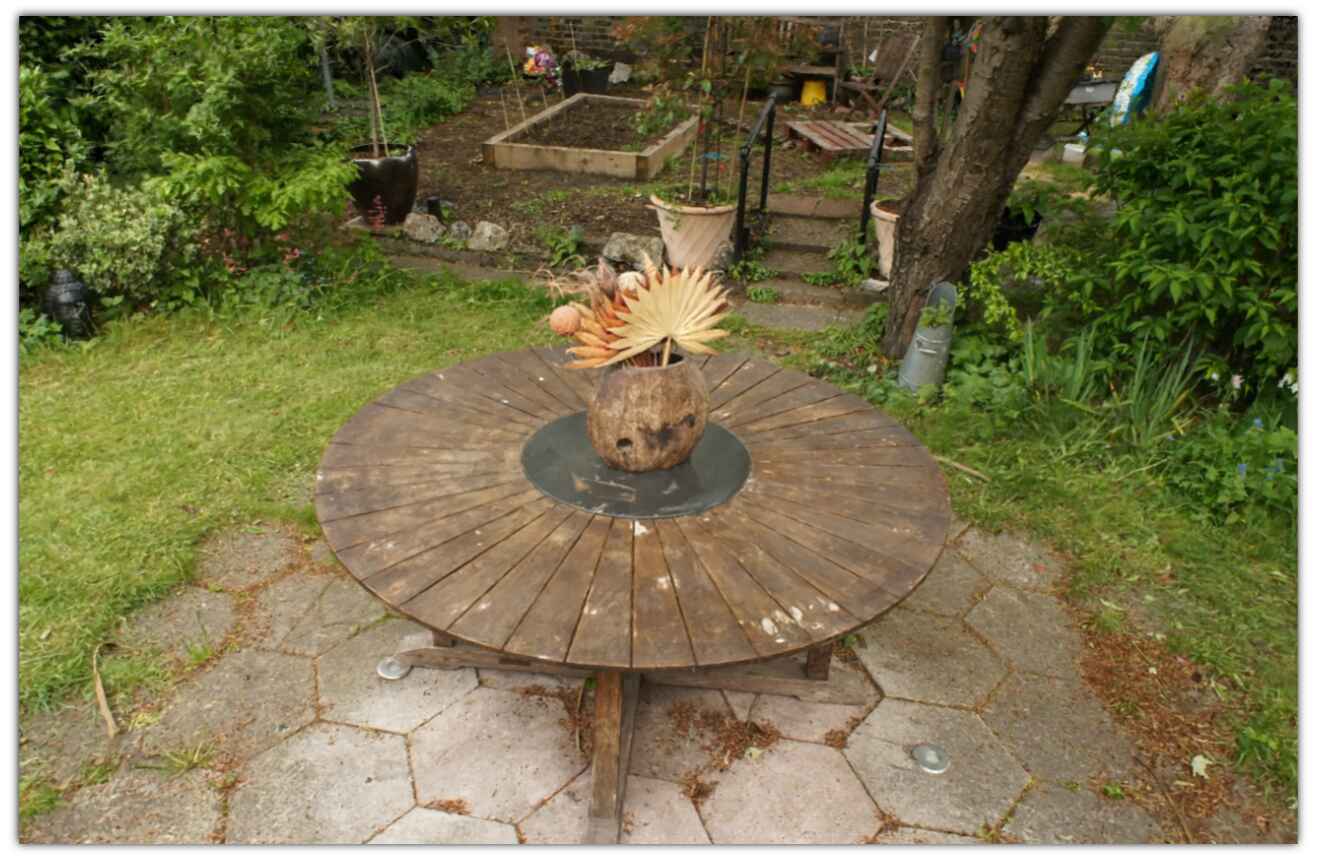}{0.55\mytmplen}{0.52\mytmplen}{0.824\mytmplen}{0.178\mytmplen}{1.70cm}{\mytmplen}{4.5}{red}
& \zoomin{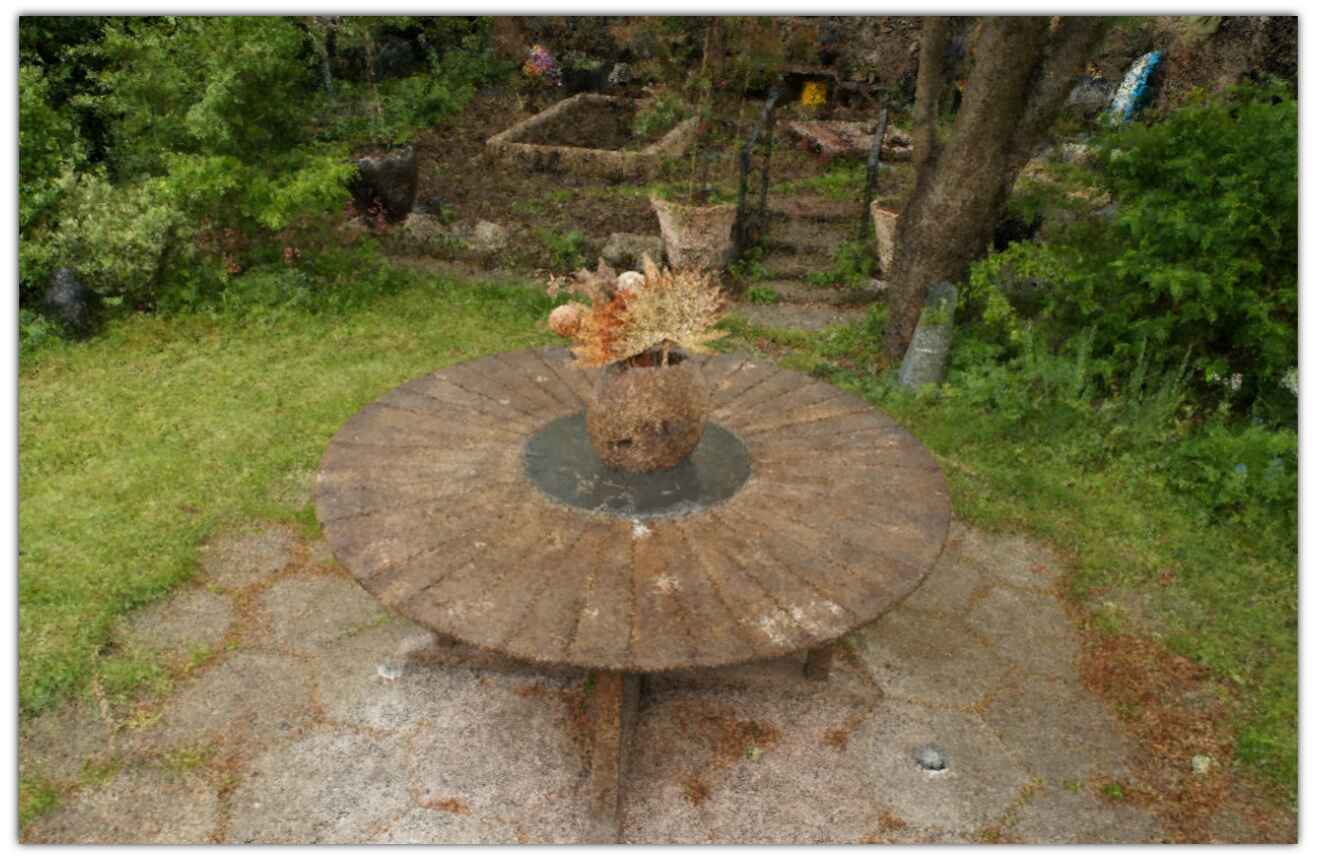}{0.55\mytmplen}{0.52\mytmplen}{0.824\mytmplen}{0.178\mytmplen}{1.70cm}{\mytmplen}{4.5}{red}
& \zoomin{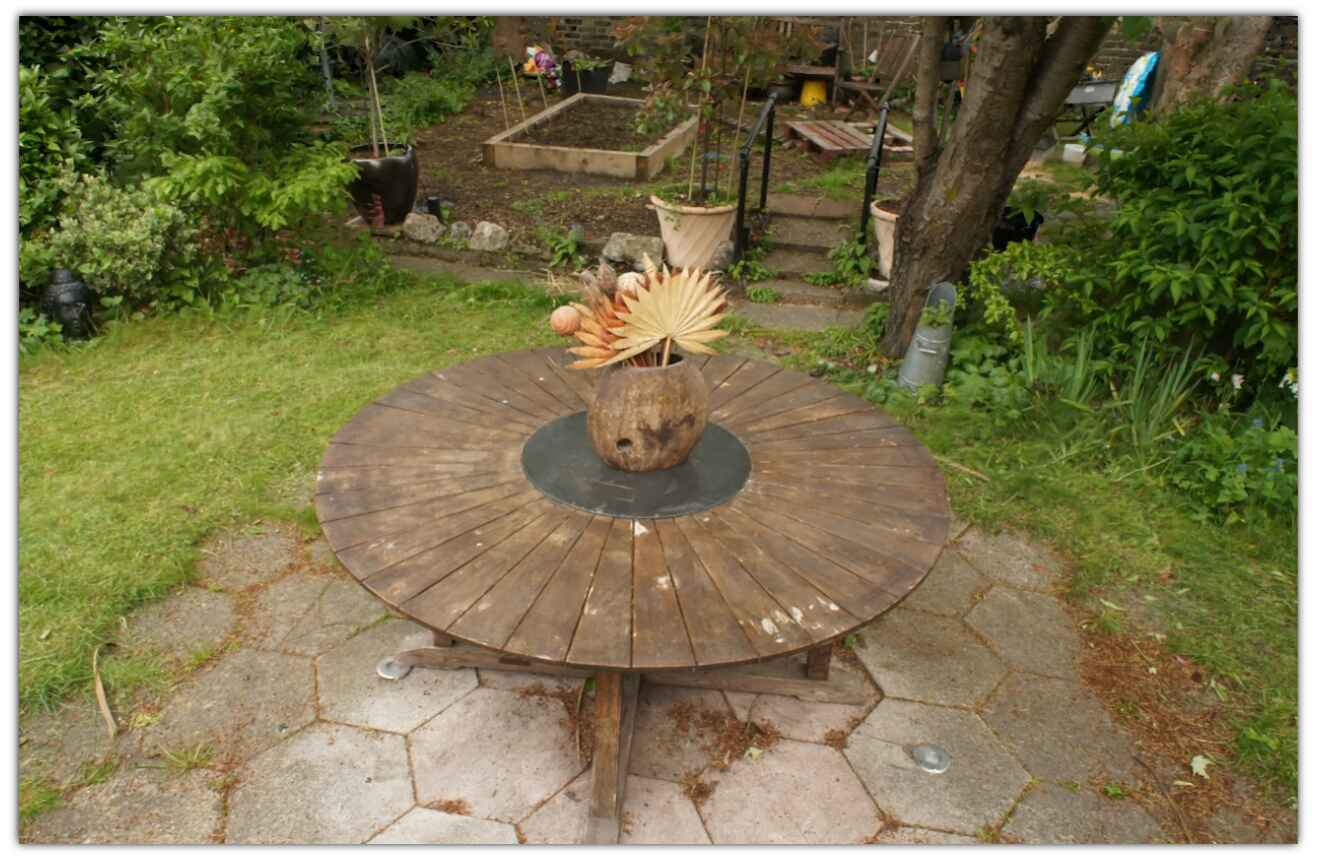}{0.55\mytmplen}{0.52\mytmplen}{0.824\mytmplen}{0.178\mytmplen}{1.70cm}{\mytmplen}{4.5}{red} \\
   % --- Row 2 ---
\rotatebox{90}{\parbox{3.5cm}{\centering \small \textbf{Normal map}}}
& \zoomin{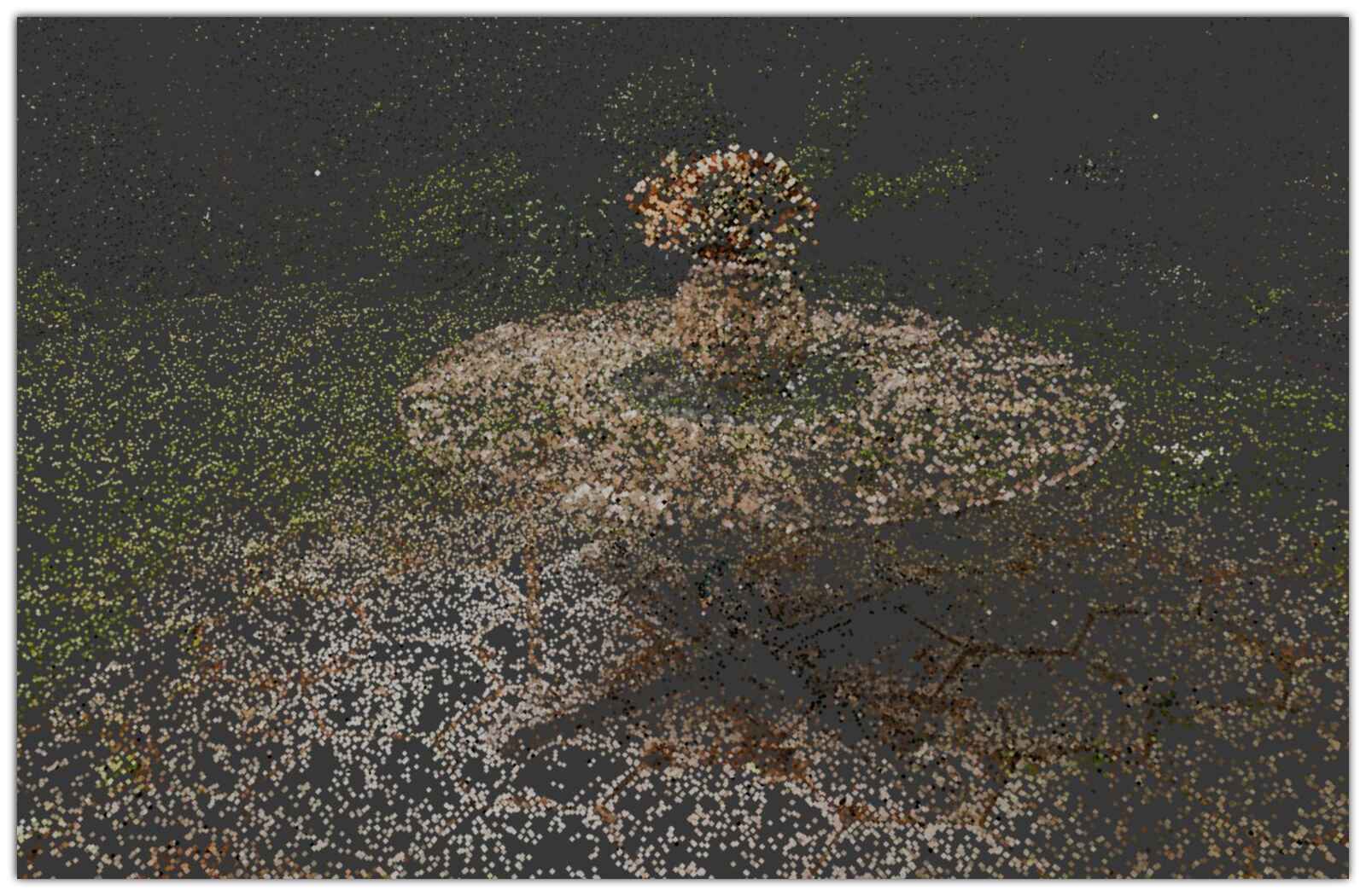}{0.37\mytmplen}{0.35\mytmplen}{0.824\mytmplen}{0.178\mytmplen}{1.70cm}{\mytmplen}{4.5}{red}
& \zoomin{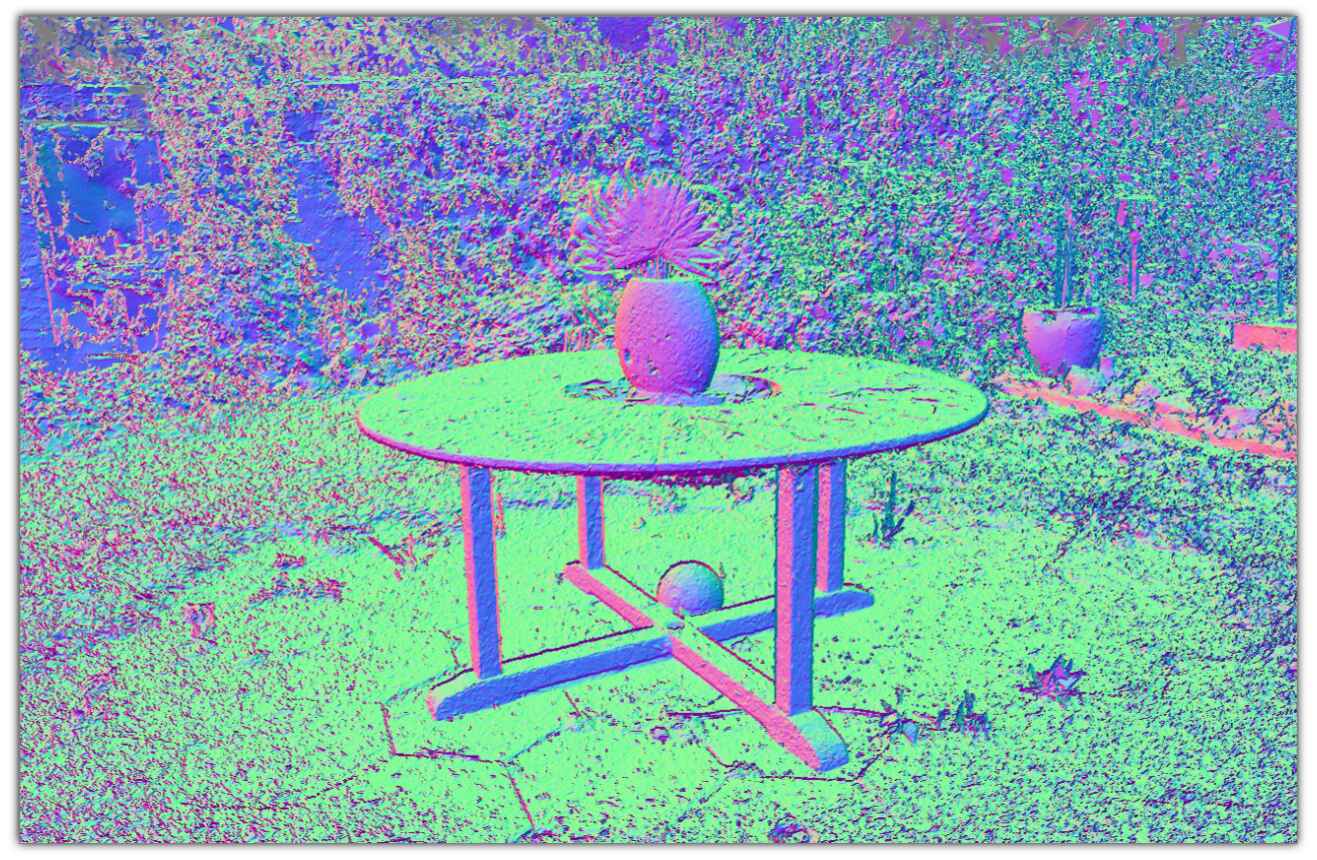}{0.37\mytmplen}{0.35\mytmplen}{0.824\mytmplen}{0.178\mytmplen}{1.70cm}{\mytmplen}{4.5}{red}
& \zoomin{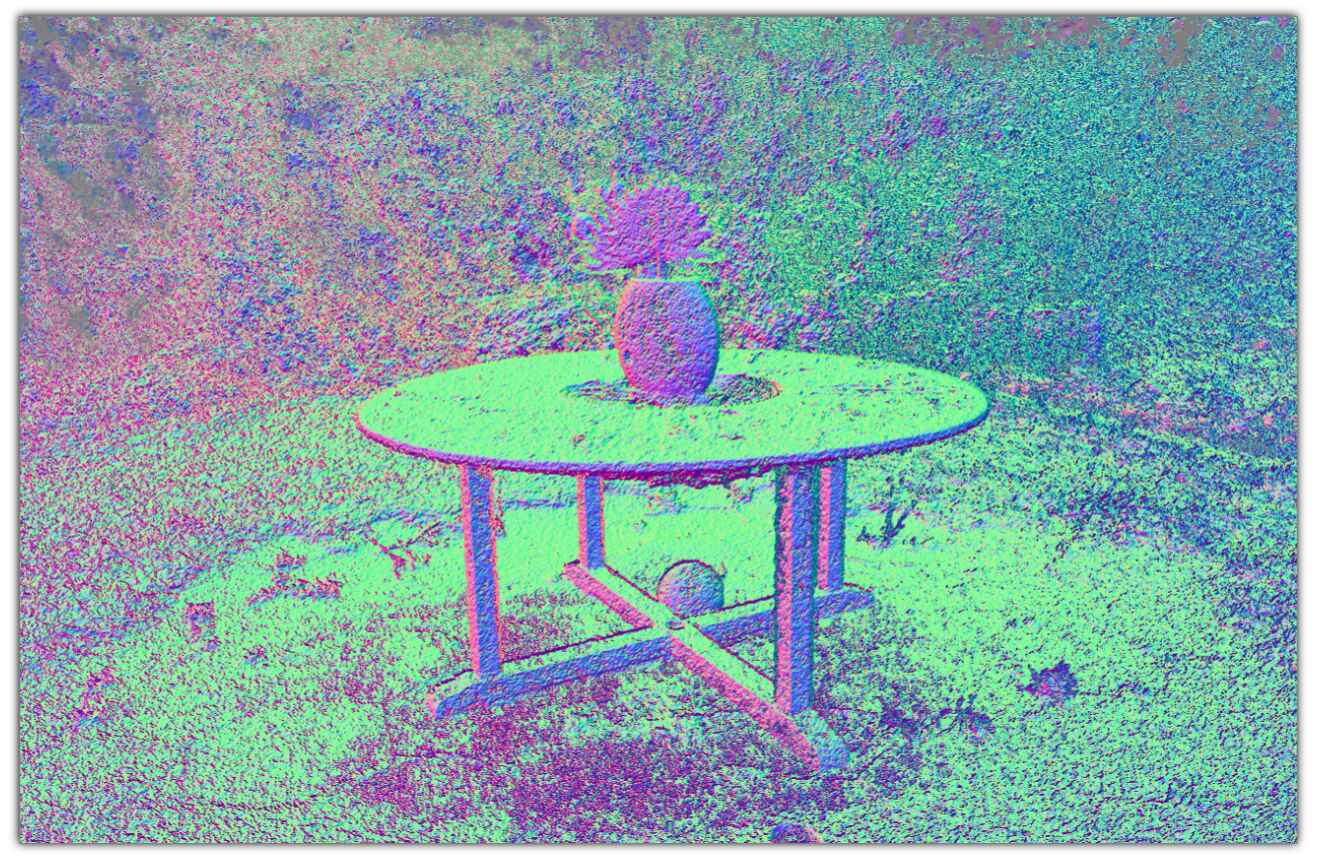}{0.37\mytmplen}{0.35\mytmplen}{0.824\mytmplen}{0.178\mytmplen}{1.70cm}{\mytmplen}{4.5}{red}
& \zoomin{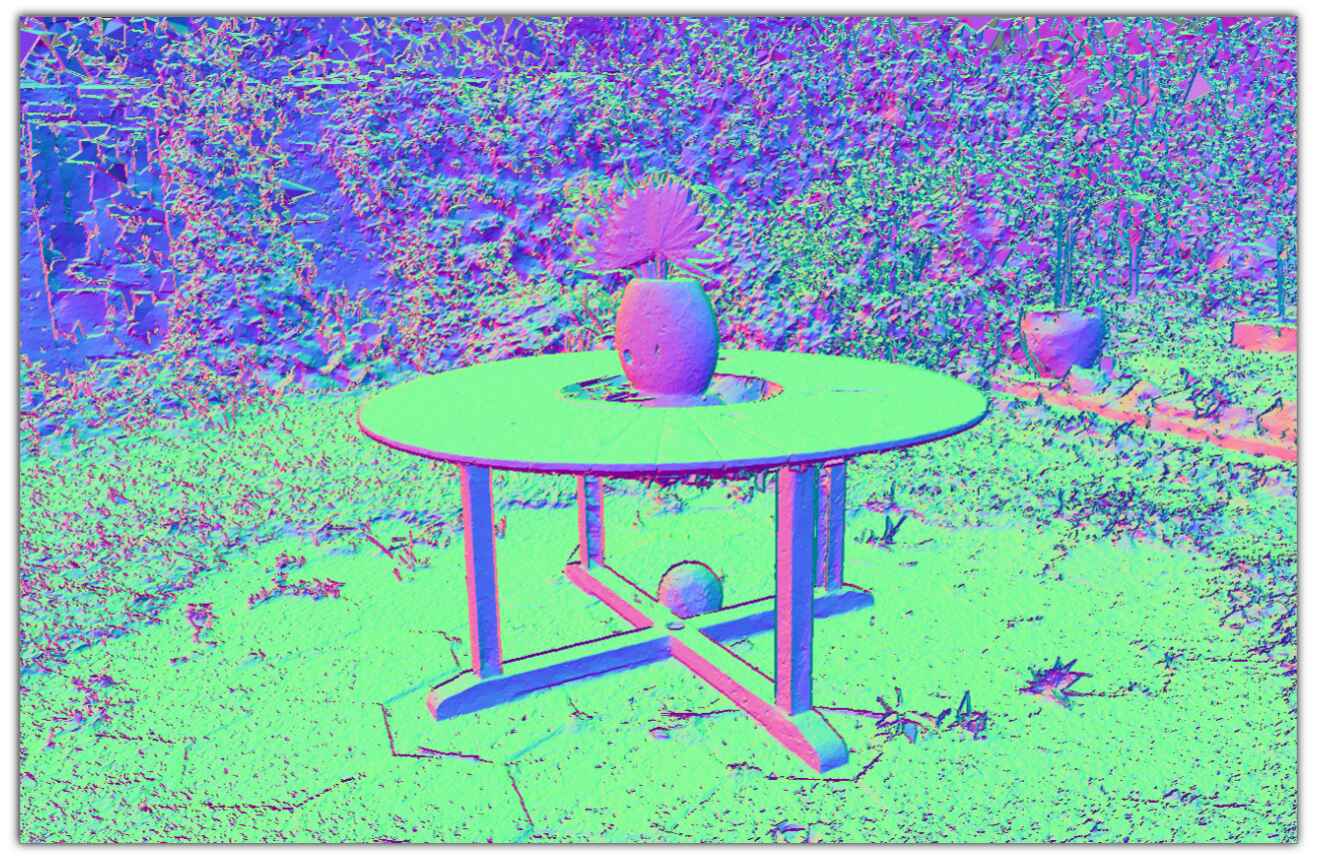}{0.37\mytmplen}{0.35\mytmplen}{0.824\mytmplen}{0.178\mytmplen}{1.70cm}{\mytmplen}{4.5}{red} \\
    % --- Labels below ---
& \makebox[\mytmplen]{\normalsize a. SfM point cloud initialization}
& \makebox[\mytmplen]{\normalsize b. transparent \& disconnected triangles}
& \makebox[\mytmplen]{\normalsize a. restricted Delaunay triangulation}
& \makebox[\mytmplen]{\normalsize b. opaque \& connected triangles} \\
\end{tabular}
}
\caption{
\textbf{From triangle soups to meshes.} 
(1a) We initialize semi-transparent triangles and scale them based on local density.
(1b) We optimize a semi-transparent triangle soup without shared vertices, leading to disconnected triangles.
(2a) Applying restricted Delaunay triangulation restores global connectivity but introduces geometric artifacts and a loss of visual quality, as vertex colors no longer accurately align with the underlying geometry.
(2b) The final fine-tuning stage refines the connected mesh, producing smooth surfaces, accurate geometry, and restoring the visual fidelity lost during triangulation. Using only opaque triangles, our method achieves high visual quality compared to the semi-transparent and isolated triangle soup.
}
\label{fig:stages}
\end{figure*}

\subsection{From soups to meshes}
\label{sec:from_soup_to_mesh}
Our optimization executes in two stages that \textit{gradually} convert an unstructured into a structured representation, as illustrated in \Cref{fig:stages}.
This design exploits the fact that, early in training, unstructured representations are \textit{easier to optimize}, as they impose less constraints on the representation.

\paragraph{Stage~1. Triangle soup optimization }
We start by taking as input a set of posed images and corresponding camera parameters obtained from structure-from-motion (SfM)~\cite{Schonberger2016Structure}, which also provides a sparse point cloud.
For each 3D point, we initialize an equilateral triangle centered at that point, with its size proportional to the average distance to its three nearest neighbors, and random orientation.
All triangles are initially defined as semi-transparent~(we selected $o_i{=}0.28$ via hyper-parameter tuning).

\noindent We begin the optimization with this \textit{unstructured} triangle soup, i.e. without any connectivity or manifold constraints between triangles. 
Each triangle is optimized independently and can move freely under the influence of image-space gradients.
This unconstrained formulation behaves similarly to point-based splatting, enabling rapid coverage of the visible scene and fast adaptation to local geometry and appearance.
Note that, while similar, this is not the same as executing \citet{Held2025Triangle-arxiv}, as we optimize \textit{interpolated} vertex quantities within a triangle, rather than keeping them uniform.

\paragraph{Stage~2. Mesh creation \& refinement}
To transform our triangle soup into a connected mesh, we execute a restricted Delaunay triangulation on the triangle soup~\cite{Cheng2013Delaunay}.
This operation first compute a standard Delaunay tetrahedralization, and then identifies tetrahedral faces whose dual Voronoi edges intersect the surface of the input triangle soup.
The restricted Delaunay triangulation generates a mesh that \textit{approximates the surface}, while at the same time, \textit{maintaining Delaunay properties} (high quality meshing) locally restricted to it.
Note that, in contrast to other reconstruction~\cite{Khan2020Poisson, Maruani2023VoroMesh, Maruani2024PoNQ}, restricted Delaunay triangulation does not introduce new vertices or modify their position.
Instead, it \textit{reuses} the optimized vertices directly, preserving both spatial accuracy and learned appearance.

\noindent Given the mesh connectivity from our restricted Delaunay triangulation, we then continue optimization so to fine-tune both vertex locations, as well as their appearance.
As vertices are shared among adjacent triangles, gradients from neighboring faces are accumulated at shared vertices, ensuring that each vertex is updated consistently according to all incident triangles; see \Cref{fig:methodology}.
We do not need to introduce additional facets or vertices, as the unstructured triangle soup optimization has already produced a sufficiently dense set of triangles to capture all spatial regions accurately.
After this fine-tuning stage, we obtain a fully opaque \textit{mesh} capable to render the scene at \textit{high photometric quality} in conventional rendering engines. In the final iterations of training, we enable supersampling, allowing even small triangles to receive gradients and be properly optimized.

\subsection{Optimizing meshes with opaque triangles}
\label{sec:towards_opaque}
Optimizing meshes composed only of \textit{opaque} triangles introduces new challenges compared to classical NeRF/3DGS optimization.
To enable gradient propagation through occlusions and allow the representation to be optimized effectively, the representation must remain semi-transparent in the early phases of training.
There are two degrees of freedom that we can control in this regard: the per-vertex opacity parameter $o$, and the smoothness parameter $\sigma$.

\paragraph{Opacity parameter scheduling}
We optimize opacity freely during the initial $5k$ iterations. 
Subsequently, we re-parameterize opacity so that the optimizer is encouraged to make triangles more opaque.
We achieve this by re-parameterizing opacity values as 
\begin{equation}
o'(o) = O_t + (1 - O_t) \cdot \mathrm{sigm}(o),
\end{equation}
where $\mathrm{sigm}(.)$ is the sigmoid function, and scheduling $O_t$ over time.
Note that if $O_t=0$ the sigmoid parameterizes opacities smoothly between 0 and 1.
Conversely, if $O_t=1$ all the opacities are mapped to a value of $1$~(fully opaque triangles).
By linearly increasing $O_t$ from $0$ to $1$ during optimization, we can control this behavior smoothly.

\begin{figure}[t]
\centering
\includegraphics[width=0.99\linewidth]{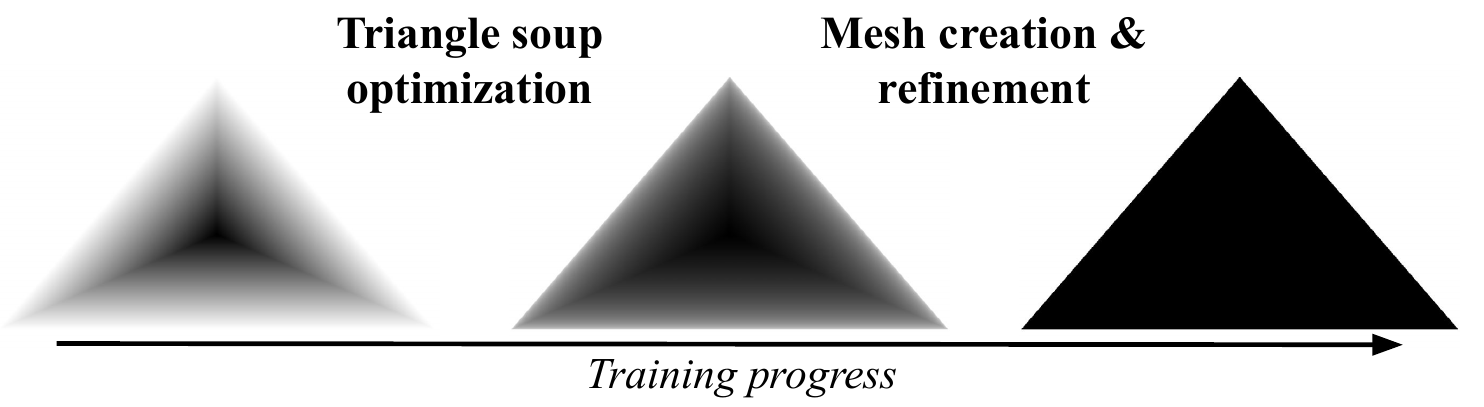}
\caption{\small
\textbf{Window parameter scheduling.} 
To ensure stable gradient flow during training, we begin with smooth triangles ($\sigma{=}1.0$, left) and linearly decrease $\sigma$ throughout training, resulting in sharper triangles by the end.
We visualize $\sigma$ for a prototypical triangle at the \textit{beginning} and \textit{end} of each optimization stage.
}
\label{fig:window}
\end{figure}

\paragraph{Window parameter scheduling}
We initialize the window parameter $\sigma{=}1.0$, corresponding to a linear transition from the incenter to the triangle boundary, and treat it as a \textit{single parameter}, shared across all triangles.
Throughout the training stages detailed in~\Cref{sec:from_soup_to_mesh}, $\sigma$ is linearly annealed from $1.0\, {\to}\, 0.0001$, ensuring strong gradient flow in the early stages while converging to opaque triangles at the end of the optimization process; see \Cref{fig:window}.
Note that this is in direct contrast to Triangle Splatting, as \citet{Held2025Triangle-arxiv} lets every triangle optimize its own window parameter.

\subsection{Optimization details}
\label{sec:optimization}
We now detail our densification/pruning strategies, training losses, and observations about our rendering process.

\paragraph{Densification}
As the initial triangle soup may not be sufficiently dense, we adapt the ideas from 3DGS-MCMC~\cite{Kheradmand20243DGaussian} to spawn additional triangles.
At each densification step, candidate triangles are selected by sampling from a probability distribution constructed directly from their opacity $o$ using Bernoulli sampling. 
Following \citet{Held2025Triangle-arxiv}, new triangles are generated through \emph{midpoint subdivision}: the midpoints of the three edges of a selected triangle are connected, splitting it into four smaller triangles. The new midpoints are added to the vertex set $\mathcal{V}$ and assigned the average color and opacity of their two adjacent vertices.
We already exploit the connectivity in the early stages, as it drastically reduces the number of newly created vertices, resulting in only $6$ new vertices after a split when connected, compared to $12$ in a triangle-soup setting.

\paragraph{Pruning}
At iteration $5k$ (Stage 1, just before opacity scheduling starts), we prune all triangles with opacity $o{<}0.2$, eliminating roughly $70\%$ of primitives.
During the rest of Stage~1, we monitor the \textit{maximum} volume rendering blending weight $w{=}T \cdot o$ across views, and prune whenever $w{<}O_t$, hence eliminating \textit{occluded} triangles as the representation becomes more opaque.
While pruning is disabled during Stage~2, we perform a final pruning pass over all training views at the end of training to remove triangles that were never rendered.

\begin{table*}[t!]
\vspace*{-1em}
	\small
\resizebox{\linewidth}{!}{
  \tabcolsep=0.15cm
\begin{tabular}{l|cccc|cccc|cccc}
	 & \multicolumn{4}{c}{Characteristics} & \multicolumn{4}{c}{Mip-NeRF360 dataset} & \multicolumn{4}{c}{Tanks \& Temples} \\
	& Mesh & Color & Connect & Ready
	& $PSNR^\uparrow$ & $LPIPS^\downarrow$     & $SSIM^\uparrow$  & $|V|^\downarrow$ & $PSNR^\uparrow$  & $LPIPS^\downarrow$     & $SSIM^\uparrow$  & $|V|^\downarrow$ \\
	\midrule 
    3DGS\cite{Kerbl20233DGaussian} & -- & -- & -- & \xmark & 27.21 & 0.214 & 0.815 & -- & 23.14 & 0.183 & 0.841 & -- \\ 
    Triangle Splatting~\cite{Held2025Triangle-arxiv} & - & - & - & \xmark & 27.16 & 0.191 & 0.814 & - & 23.14 & 0.143 & 0.857 & - \\

\midrule
  2DGS~\cite{Huang20242DGaussian} & \xmark & \xmark & \cmark & \cmark & 15.36 & 0.474  & 0.498 & \best 2M & 14.23 &  0.485  & 0.569 & 16M \\
  GOF~\cite{Yu2024Gaussian} & \xmark & \xmark & \cmark & \cmark & 20.78 & 0.465 & 0.573 & 33M  & \best 21.69 & 0.326 & 0.690 & 12M \\
  RaDe-GS~\cite{Zhang2024RaDeGS-arxiv} & \xmark & \xmark & \cmark & \cmark & 23.56 & 0.361 & 0.668 & 31M & 20.51 & 0.344 & 0.659 &10M \\
  MiLo~\cite{Guedon2025MILo-arxiv} & \cmark & \xmark & \cmark & \cmark & 24.09 & 0.323 & 0.688 & 7M &  21.46 & 0.348 & 0.706 & 4M \\
  Triangle Splatting~\cite{Held2025Triangle-arxiv} $\dagger$ & \cmark & \cmark & \xmark & \cmark & 21.05 & 0.462 & 0.558 & 3M & 17.27 & 0.402 & 0.600 & 6M \\
\midrule
    \textbf{\methodname} & \cmark & \cmark & \cmark & \cmark & \best 24.78  & \best 0.310 & \best 0.728 & 3M & 20.52 & \best 0.287 & \best 0.745 & \best 2M \\
\end{tabular}
	}
\caption{\textbf{\ul{Mesh-based} novel view synthesis on the Mip-NeRF360 dataset.} 
\methodname significantly outperforms all concurrent methods both in visual quality and in compactness, requiring far fewer vertices to achieve superior results.
\emph{Mesh} indicates whether a method directly produces a mesh (vs.\ requiring post-processing). 
\emph{Color} denotes whether the mesh is already colored or requires some form of post-processing (e.g., coloring by fine-tuning). 
\emph{Connect} specifies whether the final mesh consists of a connected component. 
\emph{Ready} means the output is directly usable in standard game engines without custom rendering shaders.
$\dagger$ with only opaque triangles.
}
 \label{tab:comparisons_main_table}
\end{table*}

\paragraph{Training losses}
We optimize the 3D vertex positions $\mathbf{v}_i$, opacity $o_i$, and spherical harmonic color coefficients $\mathbf{c_i}$ of all vertices.
Our training loss combines the photometric~$\mathcal{L}_1$ and~$\mathcal{L}_{\text{D-SSIM}}$ terms from 3DGS~\cite{Kerbl20233DGaussian}, the opacity loss~$\mathcal{L}_o$ from~\citet{Kheradmand20243DGaussian}, the \textit{depth alignment loss}~$\mathcal{L}_z$ (detailed below), the normal loss~$\mathcal{L}_n$ from~\cite{Huang20242DGaussian}, and the depth loss~$\mathcal{L}_d$ from~\citet{Kerbl2024AHierarchical}:
\begin{equation}
    \mathcal{L} = \mathcal{L}_\text{3DGS} + \beta_o \mathcal{L}_o + \beta_z \mathcal{L}_z + \beta_n \mathcal{L}_n + \beta_d \mathcal{L}_d\, .
\end{equation}
We follow~\citet{Kerbl2024AHierarchical}, and employ Depth Anything v2~\cite{Yang2024DepthV2} to align the predicted depths using their scale-and-shift procedure.
The normal loss~$\mathcal{L}_n$ can be supervised either by an external normal estimation network~\cite{Hu2024Metric3D}, or the self-supervised normal regularization from 2DGS~\cite{Huang20242DGaussian}, or both.
We employ both supervision sources in all experiments, except for the mesh quality evaluation (\Cref{subsec:dtu}), where we demonstrate that \methodname also performs effectively even in a \textit{self-supervised} setting.

\paragraph{Depth alignment loss}
To promote the creation of manifolds, we align triangles to the observed depth map using a vertex-to-surface depth loss.
To achieve this, for each rendered vertex $v_i$ with predicted depth $z_i$ and screen coordinates $(x_i, y_i)$, we sample the predicted depth $z_i^{*}$ from the rendered depth map, and (robustly via L1 losses) penalize the depth difference:
$\mathcal{L}_z = 
\frac{1}{N} \sum_{i=1}^{N} \left| z_i - z_i^{*} \right|.$
Unlike losses like normal consistency~\cite{Boss2025SF3D,Choi2024LTM} or Laplacian~\cite{Boss2025SF3D, Wang2018Pixel2Mesh-arxiv,Choi2024LTM} that rely on local mesh connectivity, this formulation acts on each vertex independently.

\paragraph{Rendering equation}
The final color of each image pixel~$\mathbf{p}$ is computed by accumulating contributions from all overlapping triangles in depth order:

$C(\mathbf{p}) = \sum_{n=1}^N \mathbf{c}_{T_n} o_{T_n} I(\mathbf{p}) \left( \prod_{i=1}^{n-1} \left(1 - o_{T_i} I(\mathbf{p})\right) \right).$
Analogously to \citet{Chen2023MobileNeRF}, at the end of training, this simplifies to $C(\mathbf{p}) {=} \mathbf{c}_{T_n} I(\mathbf{p})$, so that only a \textit{single} evaluation per pixel is required, significantly accelerating the rendering process (i.e. \textit{over-drawing is zero}).

\section{Experiments}
\label{sec:experiments}
We compare our method to concurrent approaches on Mip-NeRF360~\cite{Barron2022MipNeRF360} and Tanks and Temples (T\&T)~\cite{Knapitsch2017Tanks}.
We evaluate the visual quality using standard metrics: SSIM, PSNR, and LPIPS.
Our the total number of used vertices and training times are reported on an NVIDIA A100 (40GB).
Following \citet{Guedon2025MILo-arxiv}, we focus our evaluation on the task of \textit{Mesh-Based Novel View Synthesis}, whose evaluation protocol attempts to measure how well reconstructed meshes reproduce complete scenes.
Existing datasets such as DTU, MipNeRF360, and Tanks\&Temples are limited: MipNeRF360 lacks ground-truth geometry, DTU only contains simple objects, and T\&T provides sparse annotations with only foreground regions.
As a solution, we measure novel view synthesis quality through the visual consistency between mesh renderings and reference images, capturing \textbf{(i)} geometric alignment and surface artifacts, \textbf{(ii)} mesh completeness, and \textbf{(iii)} background reconstruction, even when ground-truth geometry is unavailable. 
Finally, to quantitatively assess the mesh quality of our method, we compute the Chamfer Distance on~DTU.

\paragraph{Baselines} 
We compare our method against Triangle Splatting~\cite{Held2025Triangle-arxiv}, and meshes derived from MiLo~\cite{Guedon2025MILo-arxiv}, 2DGS~\cite{Huang20242DGaussian}, Gaussian Opacity Fields (GOF) \cite{Yu2024Gaussian}, and RaDe-GS \cite{Zhang2024RaDeGS-arxiv}.
For MiLo, surface mesh extraction is integrated into the optimization itself, so no additional post-processing is required.
In contrast, 2DGS, GOF, and RaDe-GS rely on mesh extraction as a post-processing operation.
Furthermore, \textit{all} these methods require an additional post-processing stage to color the mesh, achieved by training a neural color field for $5k$ iterations; see MiLo for details~\cite{Guedon2025MILo-arxiv}.
For Triangle Splatting, we use the \textit{opaque}$^\dagger$ triangles version, as it produces \textit{game engine} outputs without additional post-processing needed.
For reference, we also compare against 3DGS~\cite{Kerbl20233DGaussian} to highlight the contrast in rendering quality between volumetric and mesh-based novel view synthesis.
Qualitatively, we compare against current state-of-the-art MiLo~\cite{Guedon2025MILo-arxiv} and Triangle Splatting~\cite{Held2025Triangle-arxiv}, which represents the closest line of work to ours.

\paragraph{Implementation details}
We set the spherical harmonics to degree 3, which yields 51 parameters per vertex (48 from the SH coefficients and 3 from the vertex position) and 3 parameters per triangle (the vertex indices). In comparison, a single 3D Gaussian requires 59 parameters.

\begin{figure*}[t]
\vspace*{-1em}
\centering
\setlength{\mytmplen}{0.20\linewidth} % fit 5 images + label
\resizebox{\linewidth}{!}{ 
\begin{tabular}{c@{\hskip 0.01in}c@{\hskip 0.01in}c@{\hskip 0.01in}c@{\hskip 0.01in}c}
    
    & \makebox[\mytmplen]{\scriptsize Ground Truth} &
      \makebox[\mytmplen]{\scriptsize \textbf{\methodname}} &
      \makebox[\mytmplen]{\scriptsize MiLo} &
      \makebox[\mytmplen]{\scriptsize Triangle Splatting$\dagger$} 
   \\

    \rotatebox{90}{\parbox{2.2cm}{\centering \scriptsize Bicycle}} &
    \zoomin{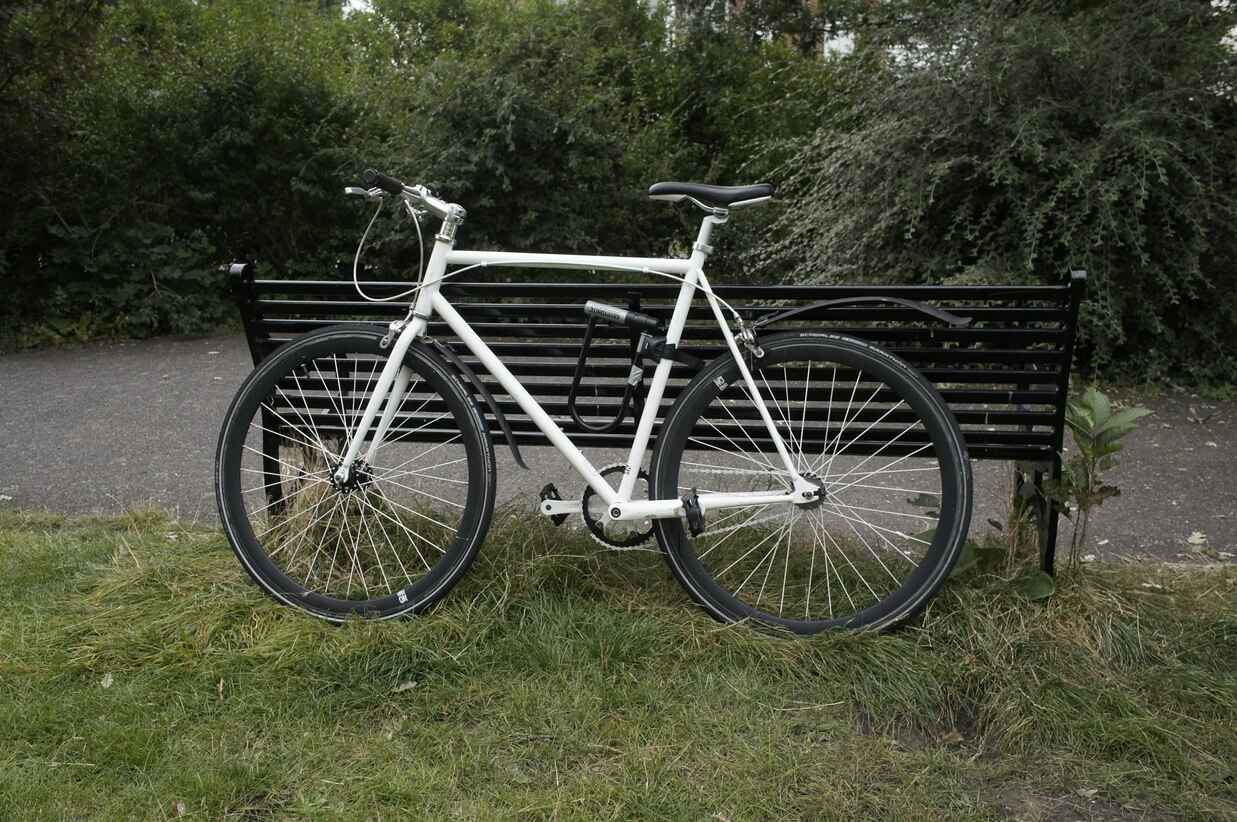}{0.27\mytmplen}{0.23\mytmplen}{0.855\mytmplen}{0.15\mytmplen}{1.0cm}{\mytmplen}{2.5}{red}
    & \zoomin{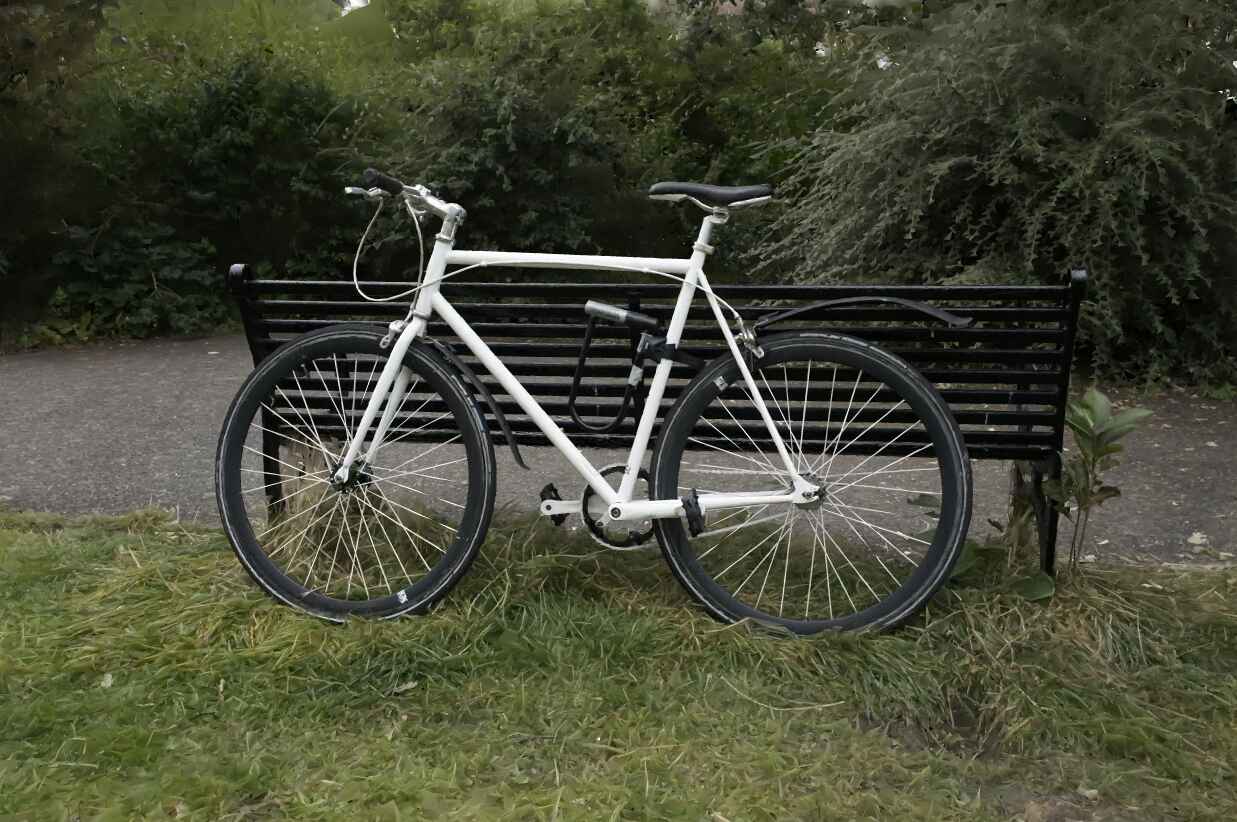}{0.27\mytmplen}{0.23\mytmplen}{0.855\mytmplen}{0.15\mytmplen}{1.0cm}{\mytmplen}{2.5}{red}
    & \zoomin{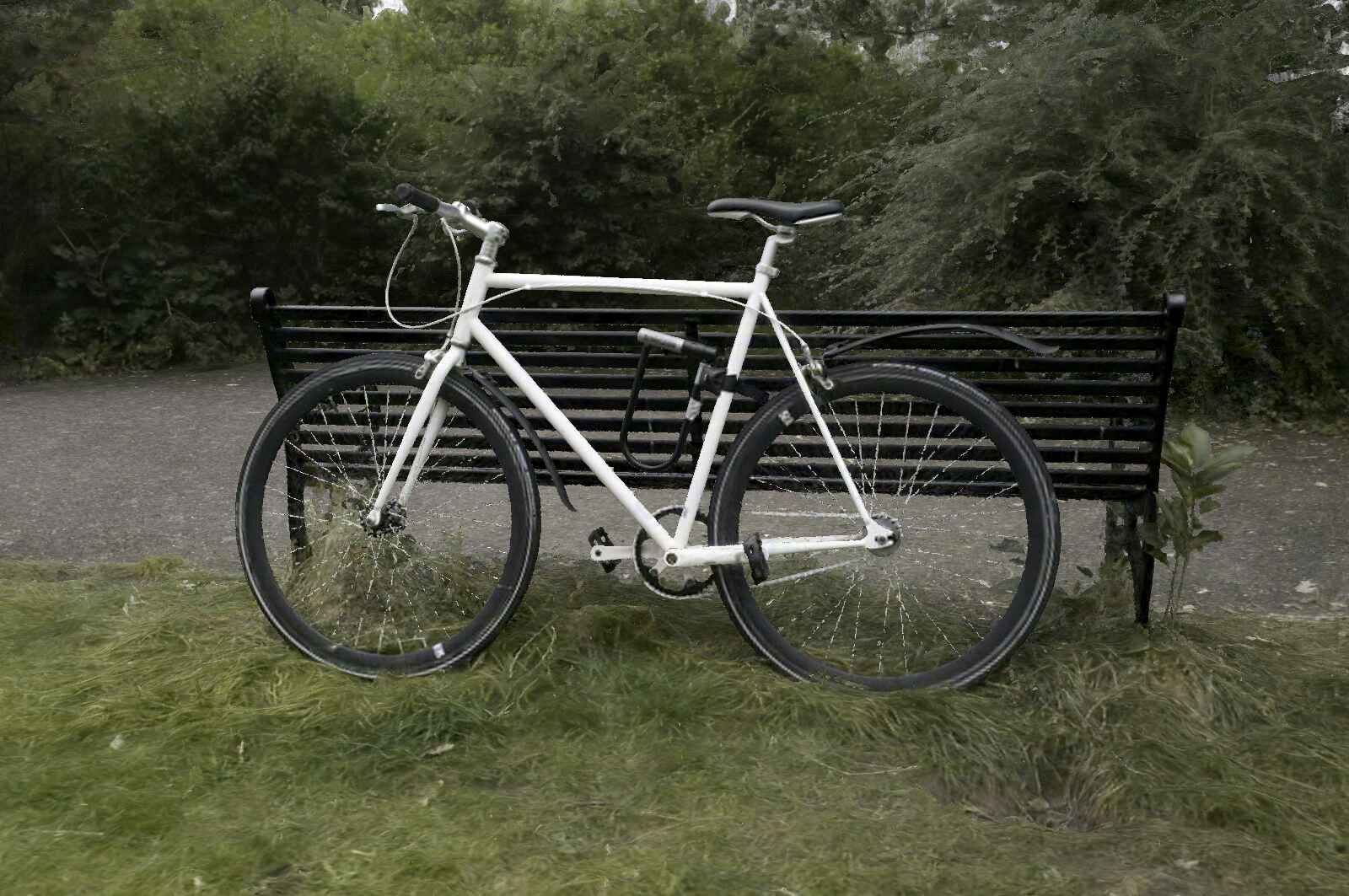}{0.27\mytmplen}{0.23\mytmplen}{0.855\mytmplen}{0.15\mytmplen}{1.0cm}{\mytmplen}{2.5}{red}
    & \zoomin{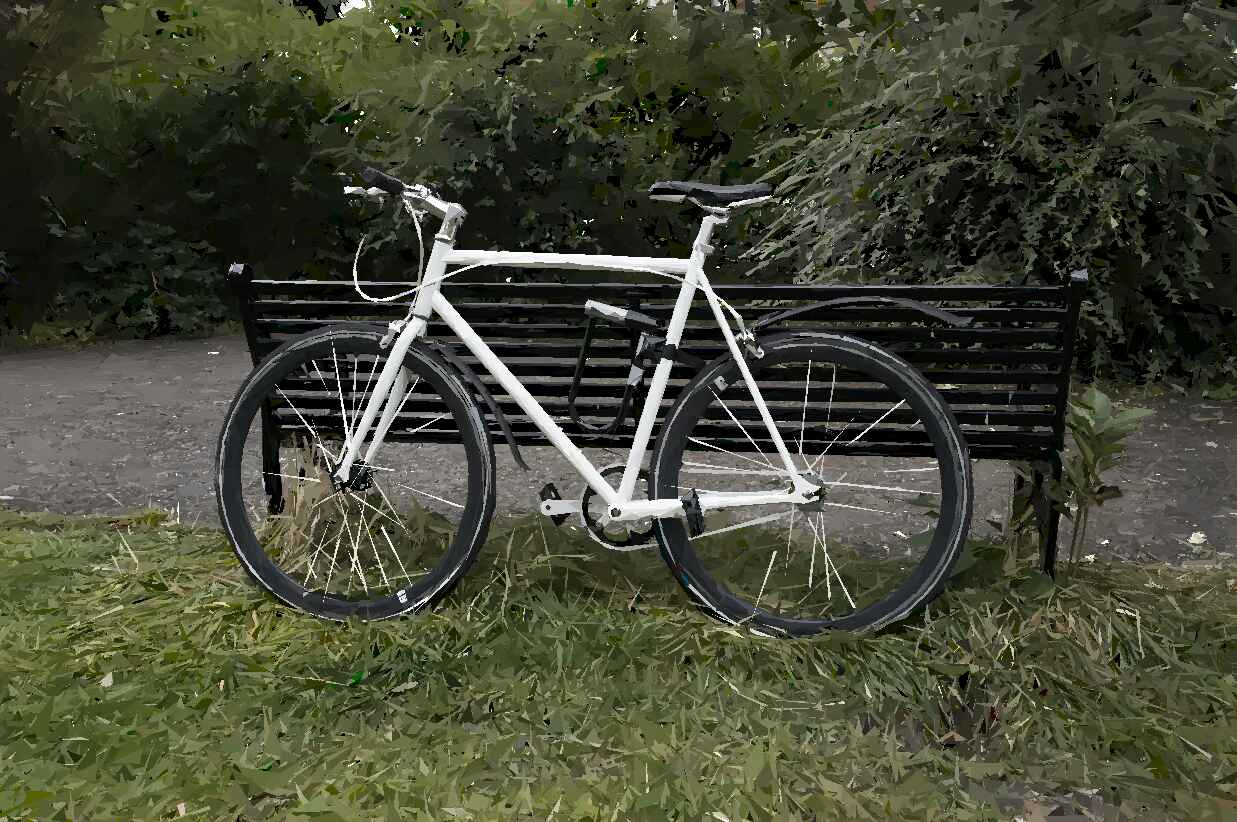}{0.27\mytmplen}{0.23\mytmplen}{0.855\mytmplen}{0.15\mytmplen}{1.0cm}{\mytmplen}{2.5}{red}
    \\

    \rotatebox{90}{\parbox{2.2cm}{\centering \scriptsize Truck}} &
    \zoomin{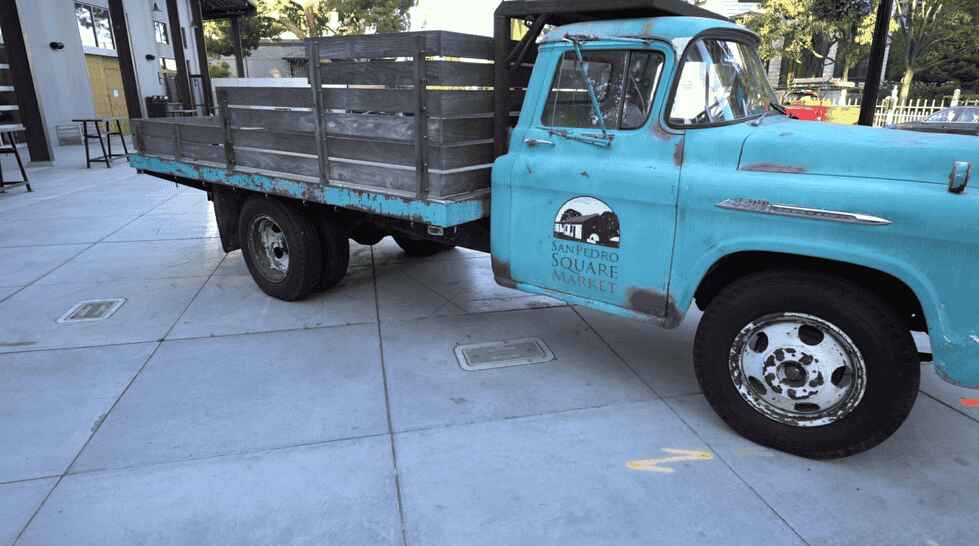}{0.12\mytmplen}{0.41\mytmplen}{0.855\mytmplen}{0.15\mytmplen}{1.0cm}{\mytmplen}{2.5}{red}
    & \zoomin{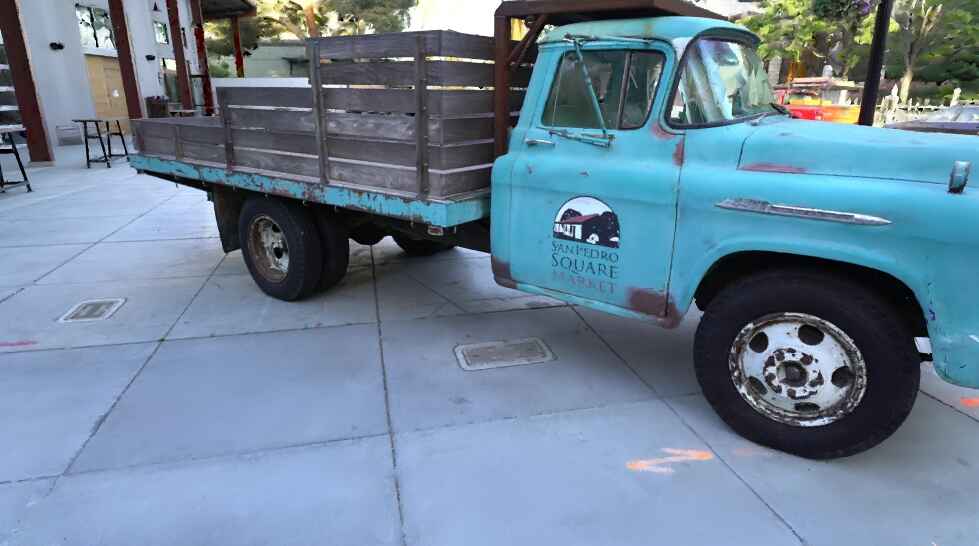}{0.12\mytmplen}{0.41\mytmplen}{0.855\mytmplen}{0.15\mytmplen}{1.0cm}{\mytmplen}{2.5}{red}
   &  \zoomin{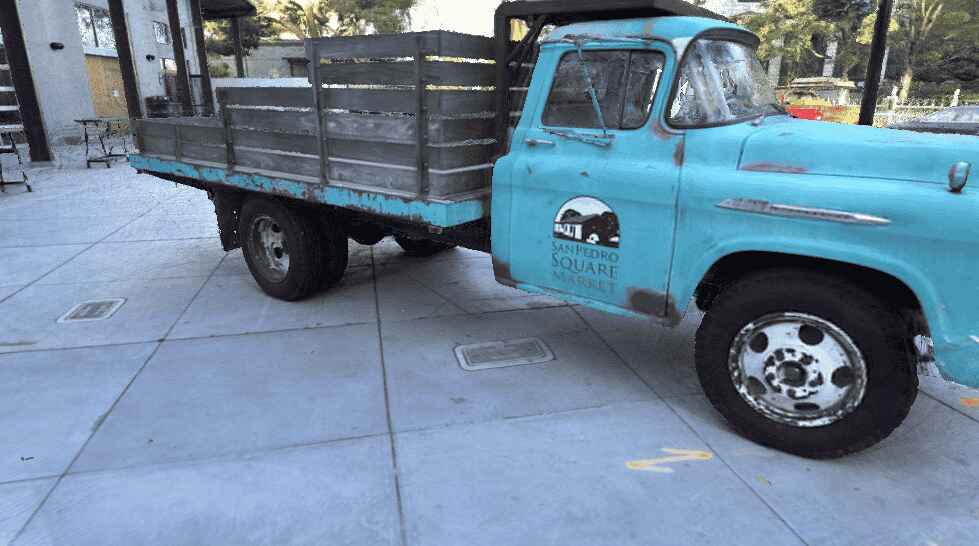}{0.12\mytmplen}{0.41\mytmplen}{0.855\mytmplen}{0.15\mytmplen}{1.0cm}{\mytmplen}{2.5}{red}
   &  \zoomin{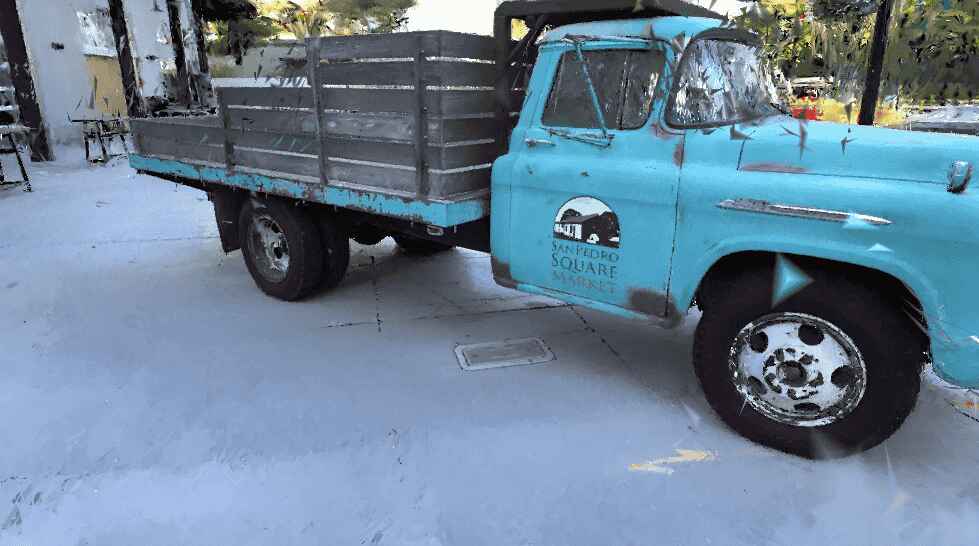}{0.12\mytmplen}{0.41\mytmplen}{0.855\mytmplen}{0.15\mytmplen}{1.0cm}{\mytmplen}{2.5}{red}
       \\

          \rotatebox{90}{\parbox{2.2cm}{\centering \scriptsize Train}} &
    \zoomin{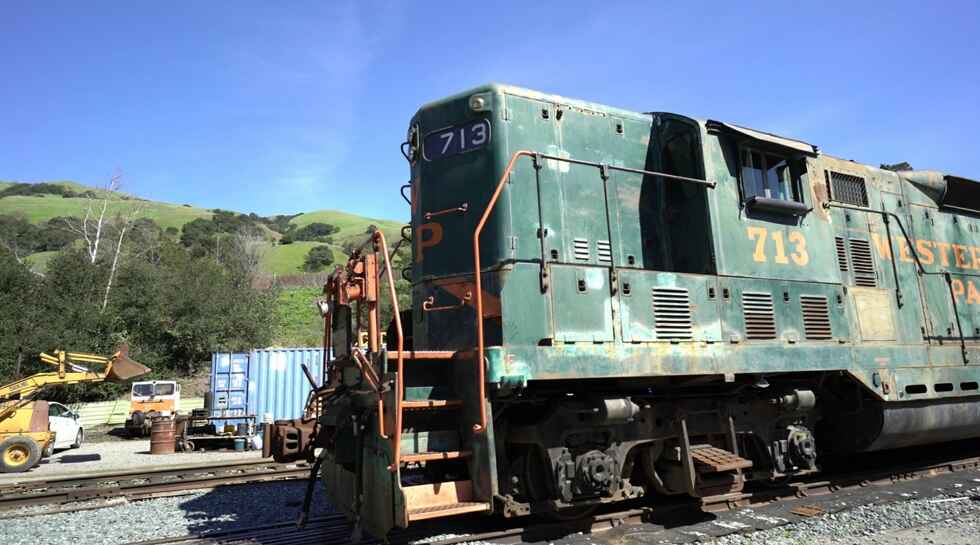}{0.57\mytmplen}{0.37\mytmplen}{0.147\mytmplen}{0.15\mytmplen}{1.0cm}{\mytmplen}{2.5}{red}
    & \zoomin{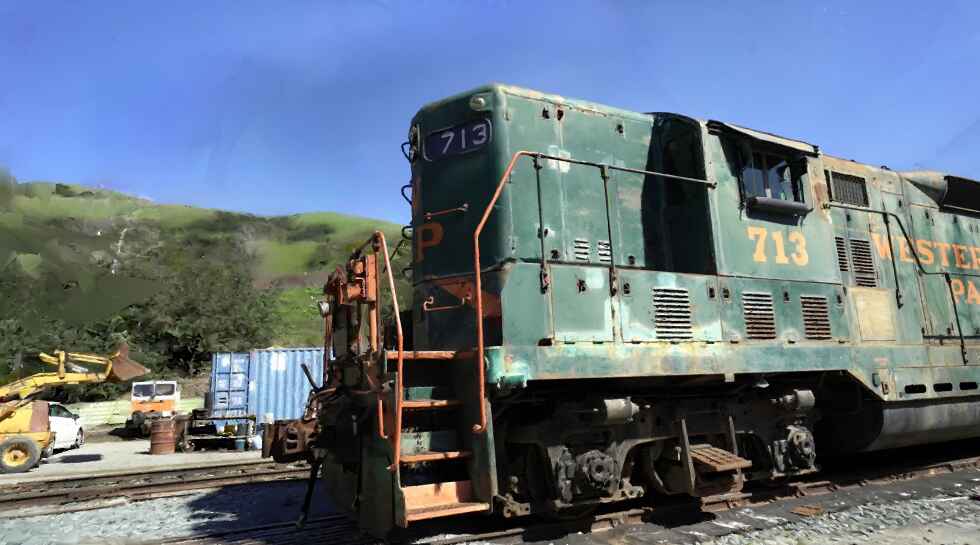}{0.57\mytmplen}{0.37\mytmplen}{0.147\mytmplen}{0.15\mytmplen}{1.0cm}{\mytmplen}{2.5}{red}
    & \zoomin{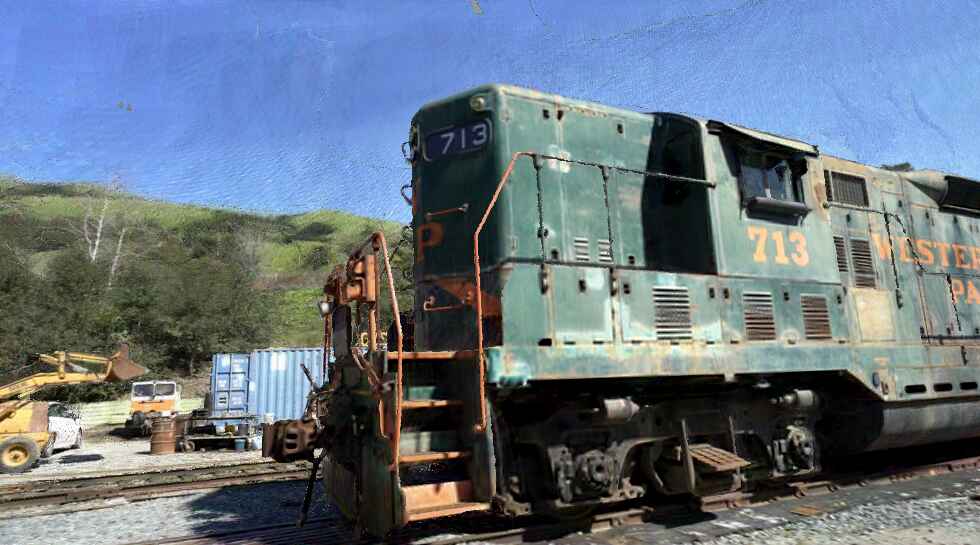}{0.57\mytmplen}{0.37\mytmplen}{0.147\mytmplen}{0.15\mytmplen}{1.0cm}{\mytmplen}{2.5}{red}
    & \zoomin{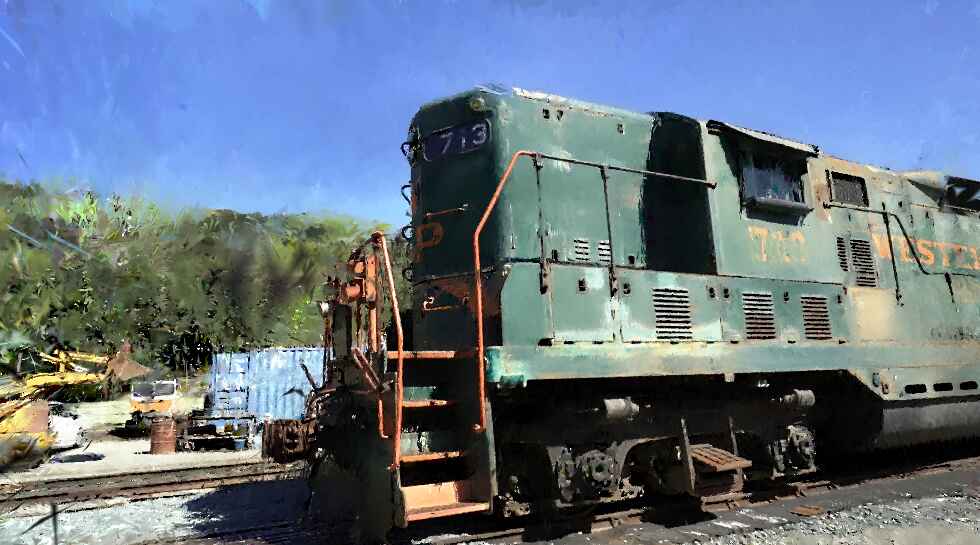}{0.57\mytmplen}{0.37\mytmplen}{0.147\mytmplen}{0.15\mytmplen}{1.0cm}{\mytmplen}{2.5}{red}

\end{tabular}
}

\caption{\small \textbf{Qualitative results.} 
Comparison of our method with ground truth, the current state-of-the-art MiLo~\cite{Guedon2025MILo-arxiv} and opaque$\dagger$ Triangle Splatting~\cite{Held2025Triangle-arxiv}.
Our approach produces renderings that are {\em closer to the ground truth, with sharper details and finer structures} (see the \textit{Bicycle} spokes), and with fewer artifacts (see the table in the \textit{Truck} scene).
More visualizations are available in the \textit{supplementary material}.
}
\label{fig:qualityresults}
\end{figure*}

\subsection{Mesh-based NVS -- \Cref{tab:comparisons_main_table} and \Cref{fig:qualityresults}}
\Cref{tab:comparisons_main_table} reports quantitative results on the Mip-NeRF360 and Tanks \& Temples datasets for mesh-based novel view synthesis.
Compared to 2DGS and Triangle Splatting, our method uses a similar number of vertices, yet it achieves a 4–10 dB higher PSNR and a significantly lower LPIPS.
Compared to GOF, RaDe-GS and MiLo, \methodname uses $2$ to $10$ times fewer vertices while obtaining significantly higher SSIM and lower LPIPS.
In terms of LPIPS~(the metric that best correlates with human visual perception), \methodname significantly outperforms all concurrent methods. 

On the Mip-NeRF360 dataset, our method achieves substantially higher PSNR than GOF, RaDe-GS and MiLo.
On T\&T, GOF and MiLo attain a higher PSNR but exhibit noticeably lower SSIM and higher LPIPS scores.
This suggests that although GOF and MiLo produce highly detailed meshes, their renderings contain more artifacts, which degrade perceptual quality and lead to worse LPIPS and SSIM scores.
We illustrate this qualitatively in \Cref{fig:qualityresults} and provide more examples in the \textit{supplementary material}.
\methodname reconstructs fine structures and details more accurately and produces less noisy renderings.

Additionally, note that 2DGS, GOF, and RaDe-GS require two additional post-processing steps after training: first \textit{extracting} a mesh, and then \textit{coloring} it. 
MiLo directly outputs a mesh, but still relies on a post-processing stage to \textit{texture} the mesh.
These extra steps limit the practicality of such methods, and increase their overall complexity.
In contrast, we \textit{directly produce colored opaque meshes that are immediately compatible} with any game engine, \textit{without} requiring additional steps.

\begin{table}[t]
\centering
\resizebox{0.99\linewidth}{!}{
\begin{tabular}{@{}l|cccc@{}}
Method & Train~$\downarrow$ & FPS~$\uparrow$ (HD) & FPS~$\uparrow$ (Full HD) & Memory~$\downarrow$  \\
\midrule
GOF & 74m & OOM & OOM & 1.5GB \\
RaDe-GS & 84m & OOM & OOM & 1.1GB \\
MiLo & 106m & 170 & 160 & 253MB \\
\midrule
\textbf{\methodname} & \best 48m & \best 220 & \best 190 & \best 100MB \\
\end{tabular}
}
\caption{\textbf{Speed \& memory on MipNeRF-360.} \methodname achieves {\em faster training and lower memory usage} than concurrent methods. FPS were measured on a costumer MacBook M4.
}
\label{tab:memory}
\end{table}

\subsection{Training speed \& memory -- \Cref{tab:memory}} 
We report the average training time and final mesh size on the Mip-NeRF360 dataset. \methodname trains in only 48 minutes on average, achieving a 35–55\% speedup over concurrent mesh-yielding methods. Our optimized restricted Delaunay triangulation runs in under two minutes, contributing negligibly to the total training time. MiLo performs Delaunay triangulation at \textit{every iteration}, whereas we run it only once.
This results in a training time of 106 minutes for MiLo, compared to just 48 minutes for \methodname.
The final mesh representation produced by \methodname is substantially more compact, with only 100 MB, which corresponds to a $2.5$–$15\times$ reduction compared to concurrent methods. GOD and RaDe-GS require 1.5 GB and 1.1 GB of memory, respectively, making them impractical for real-world applications.
This compactness enables \methodname to run efficiently even on consumer hardware, significantly opening its applicability to lightweight rendering and real-time use-cases.
\methodname renders ${\approx} 25\%$ faster on a consumer M4 MacBook compared to existing approaches. GOF and RaDe-GS exceed the device’s memory capacity, resulting in an out-of-memory error.

\subsection{Surface reconstruction}
\label{subsec:dtu}
For evaluation on the DTU dataset, we use only self-supervised regularization ($\beta_d$ to zero) to ensure a fair comparison with other methods and to show that \methodname also performs well in a self-supervised setting.
Although we designed \methodname for mesh-based novel-view synthesis in large and complex real scenes, rather than surface reconstruction, it achieves mesh quality comparable to concurrent methods. Across the 15 scenes, \methodname attains the lowest Chamfer distance in 5 of them. 
A detailed comparison table is provided in the \textit{supplementary material}.

\begin{figure}[t]
\centering
\vspace*{-0.5em}
\includegraphics[width=0.90\linewidth]{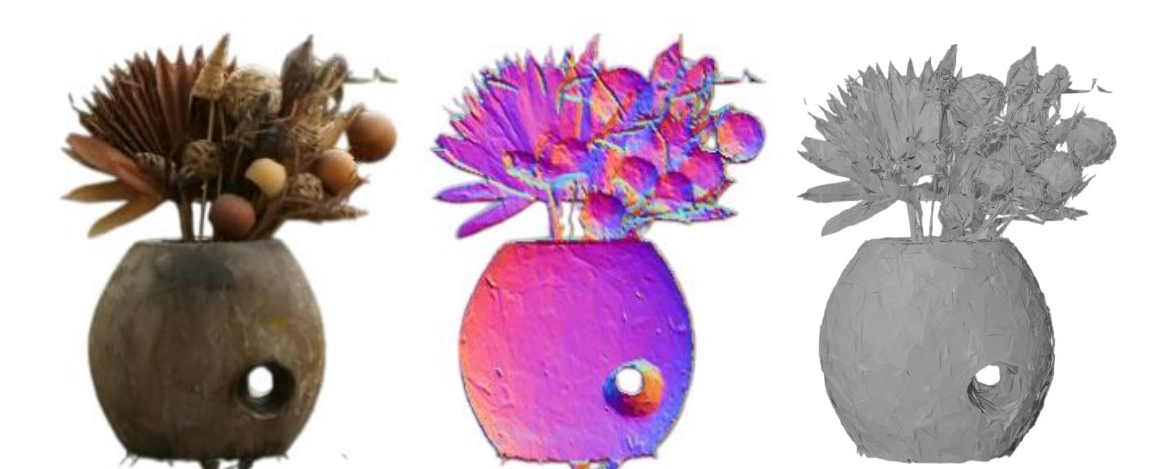}
\caption{\small
\textbf{Object segmentation.}
With \methodname, objects can be easily extracted or removed from a reconstructed scene, in this case the flower pot from the \textit{Garden} scene. From left to right: generated RGB image of the object, estimated surface normals, and resulting mesh representation.
}
\label{fig:extraction}
\end{figure}

\subsection{Applications}
We now illustrate a few application demos that are quite difficult to implement with semi-transparent representations, yet almost trivial once the scene is represented with opaque meshes like in \methodname.

\paragraph{Physical simulation}
As our representation contains no semi-transparent primitives, the triangles may directly be interpreted as hard surfaces for the purpose of physical simulation.
We demonstrate this by using an off-the-shelf non-convex mesh collider:  the one provided in the \textit{Unity game engine}.
With no post-processing, our mesh can be loaded into Unity, and used for physics interaction with dynamic objects and characters.
\Cref{fig:teaser} showcases a selection of downstream applications, with additional visualizations in the \textit{supplementary material.}

\paragraph{Object segmentation}
Current 3D Gaussian Splatting methods for object extraction or removal \cite{Cen2025Segment, Ye2024Gaussian} face a fundamental challenge: a single pixel is rendered by accumulating the contributions of many primitives.
This makes it non-trivial to decide whether a primitive belongs to a given object. 
To address this, prior work learns object associations during optimization \cite{Cen2025Segment, Ye2024Gaussian}.
In contrast, \methodname requires no additional tweak.
Since each pixel is covered by exactly one triangle, mapping objects from image space to 3D space becomes straightforward: given a 2D mask of an object, all triangles contributing to pixels within that mask are directly identified as part of the object.
By iterating over all training views, we recover the complete set of triangles belonging to the object. 
The object masks are generated using SAMv2~\cite{Ravi2025SAM2}, which enables the selection of single or multiple objects that can then be removed or extracted from a scene with minimal complexity.
\Cref{fig:extraction} shows a qualitative example of the extracted \textit{flower pot} from the \textit{Garden} scene of the Mip-NeRF360 dataset. 
Beyond object removal and extraction, this capability allows the scene to be segmented into distinct sub-meshes by disconnecting triangles belonging to different objects, enabling structured and object-aware 3D scene editing.

\begin{table}[t]
\centering
\resizebox{0.99\linewidth}{!}{%
\begin{tabular}{l|ccccc|c}
 & 0 & 1 & 2 & 3 & 4+ & Mean \\
\midrule
Restricted Delaunay & 1\% & 2\% & 5 \% & 11\% & 81\% & 6.1 \\
With pruning & 9\% & 22\% & 25\% & 19\% & 25\% & 2.5 \\
Final mesh & 2\% & 9\% & 16\% & 20\% & 53\% & 3.7 \\
\end{tabular}%
}
\caption{\small
\textbf{Connectivity.} Distribution of triangle connectivity on the \textit{Garden} scene. The final mesh mostly consists of triangles connected to three or more neighboring triangles, indicating a well-connected mesh. %(i.e. $\neq$ triangle soup).
}
\label{table:connectivity}
\end{table}

\begin{table}[t]
\centering
\resizebox{0.7\linewidth}{!}{%
\begin{tabular}{l|rrr}
& PSNR~$\uparrow$ & LPIPS~$\downarrow$ & SSIM~$\uparrow$ \\
\midrule
\textbf{Baseline} & \textbf{24.78} & \textbf{~0.31} & \textbf{~0.728} \\
\midrule
\quad w/o SH & -2.07 & +0.06 & -0.069 \\
\quad w/o $\mathcal{L}_{d}$ & +0.05 & -0.04 & +0.006 \\
\quad w/o $\mathcal{L}_{z}$ & +0.02 & -0.01 & +0.002\\
\quad w/o $\mathcal{L}_{n}$ & +0.10 & -0.02 & +0.004 \\
\end{tabular}%
}
\caption{\small
\textbf{Ablations (Mip-NeRF360).} 
We assess the impact of each design choice by removing each one of them individually.
}
\label{table:ablation_losses}
\end{table}

\subsection{Analysis}
\paragraph{Mesh connectivity -- \Cref{table:connectivity}}
We report the distribution of triangles according to their connectivity with neighboring triangles.
After the restricted Delaunay triangulation, most triangles are already well connected, with about $92\%$ having at least three neighbors.
If pruning is applied immediately after this step, a large number of triangles are removed because the Delaunay mesh contains many very small triangles that do not contribute to rendering, significantly reducing the overall connectivity.
Instead, pruning is performed only after training. We make a final pass over all training views and remove all triangles that did not contribute to any rendered image.
On average, each triangle is connected to approximately $3.7$ triangles, and fewer than $2\%$ of triangles remain isolated.
Overall, most triangles are connected to exactly three neighboring triangles.
More analyzes are provided in the \textit{supplementary material}.

\begin{figure}[t]
\vspace*{-1em}
\centering
\setlength{\tabcolsep}{2pt}
\begin{tabular}{cc}
\includegraphics[width=0.45\linewidth]{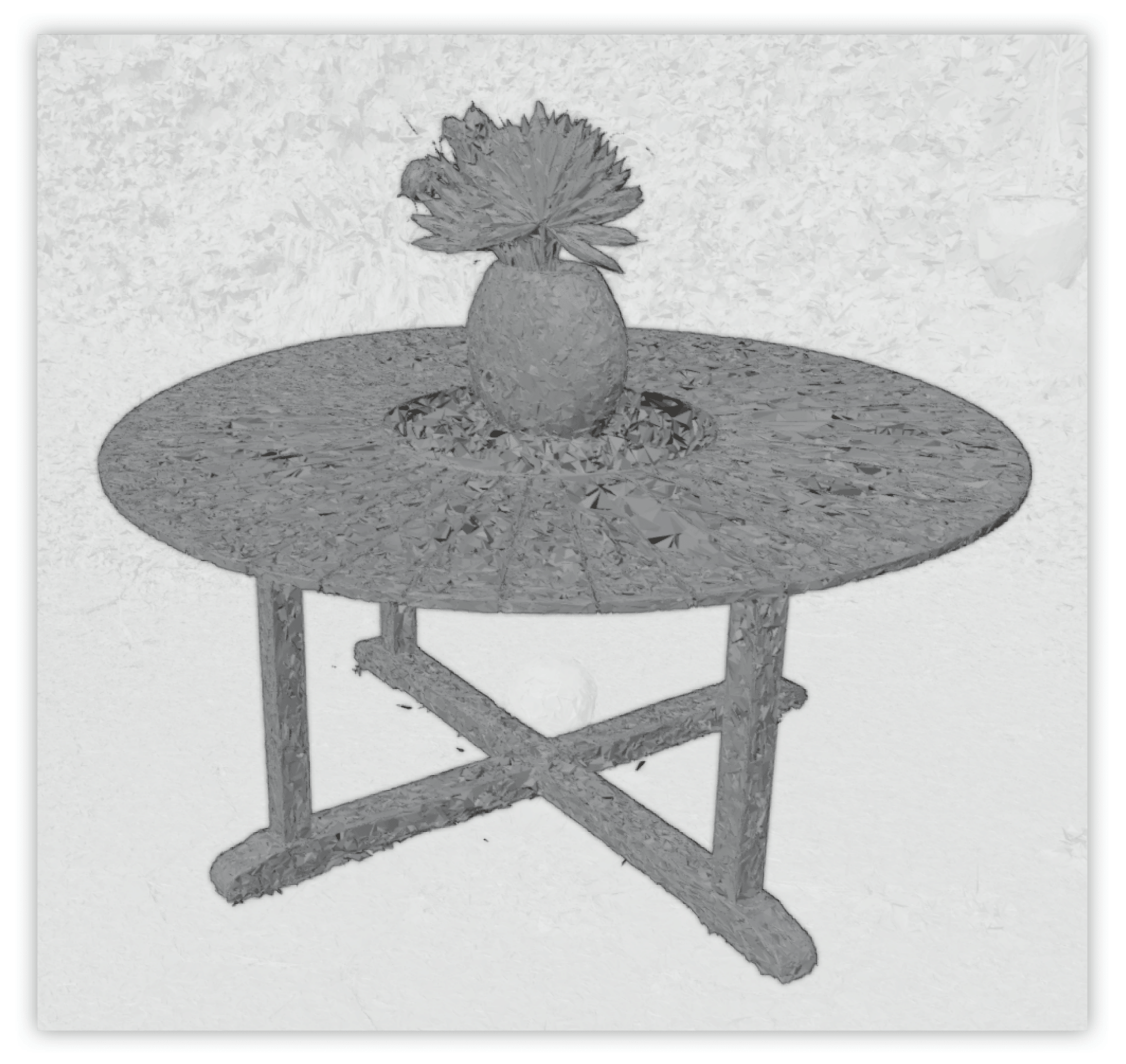} &
\includegraphics[width=0.45\linewidth]{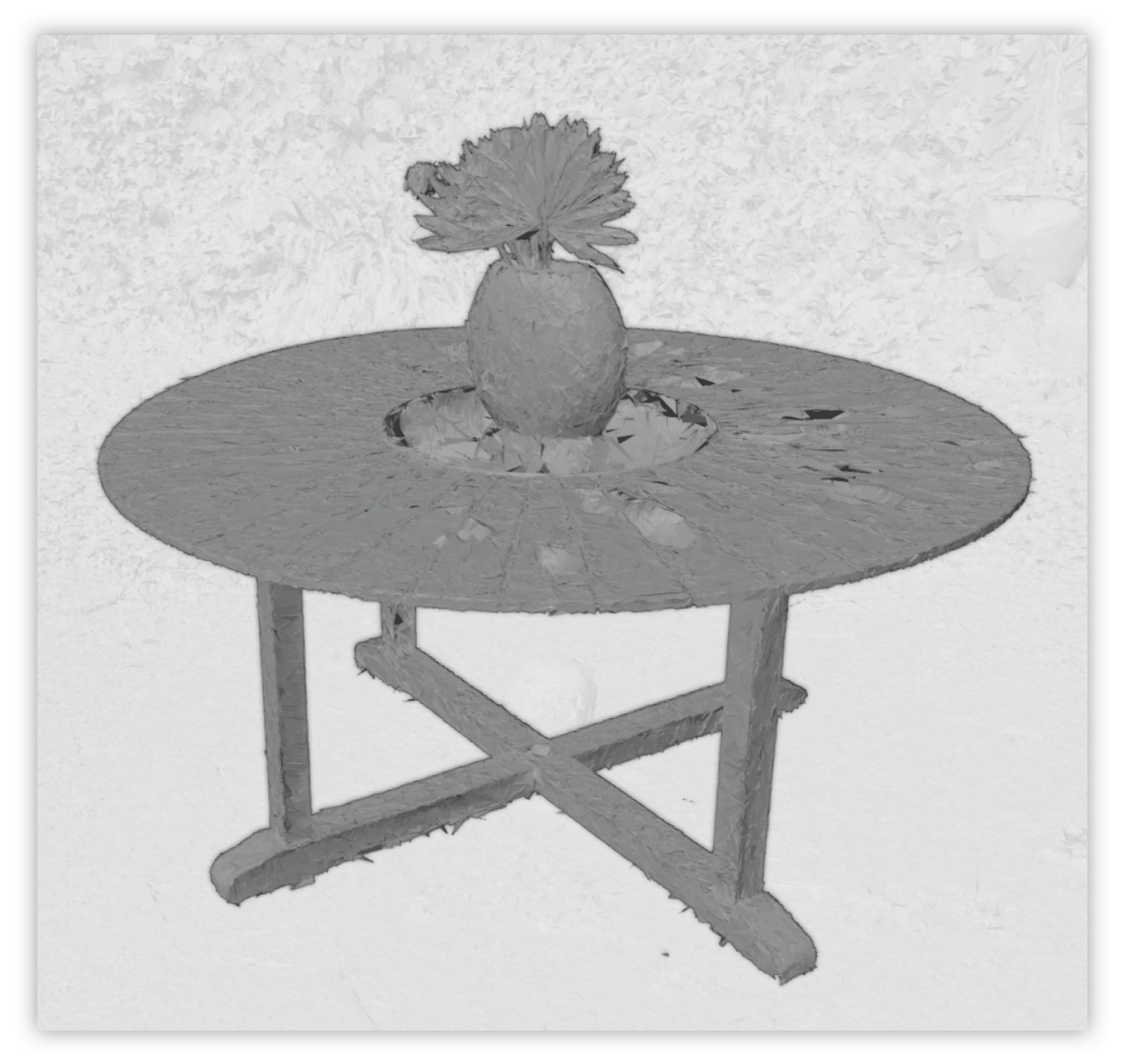}
\\[-.5em]
\scriptsize (a) no regularization &
\scriptsize (b) w/o $\mathbf{L_d}$
\\
\includegraphics[width=0.45\linewidth]{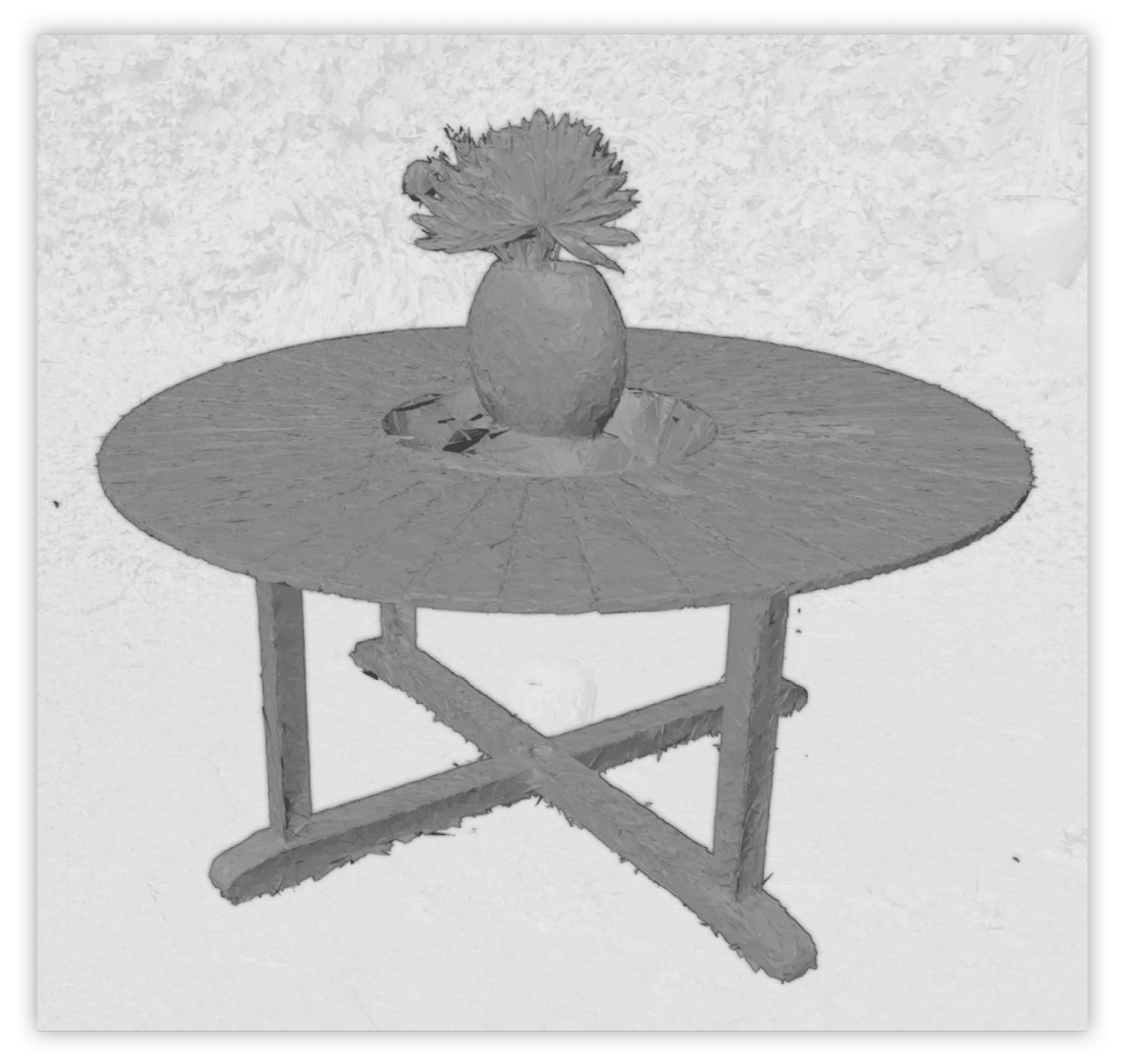} &
\includegraphics[width=0.45\linewidth]{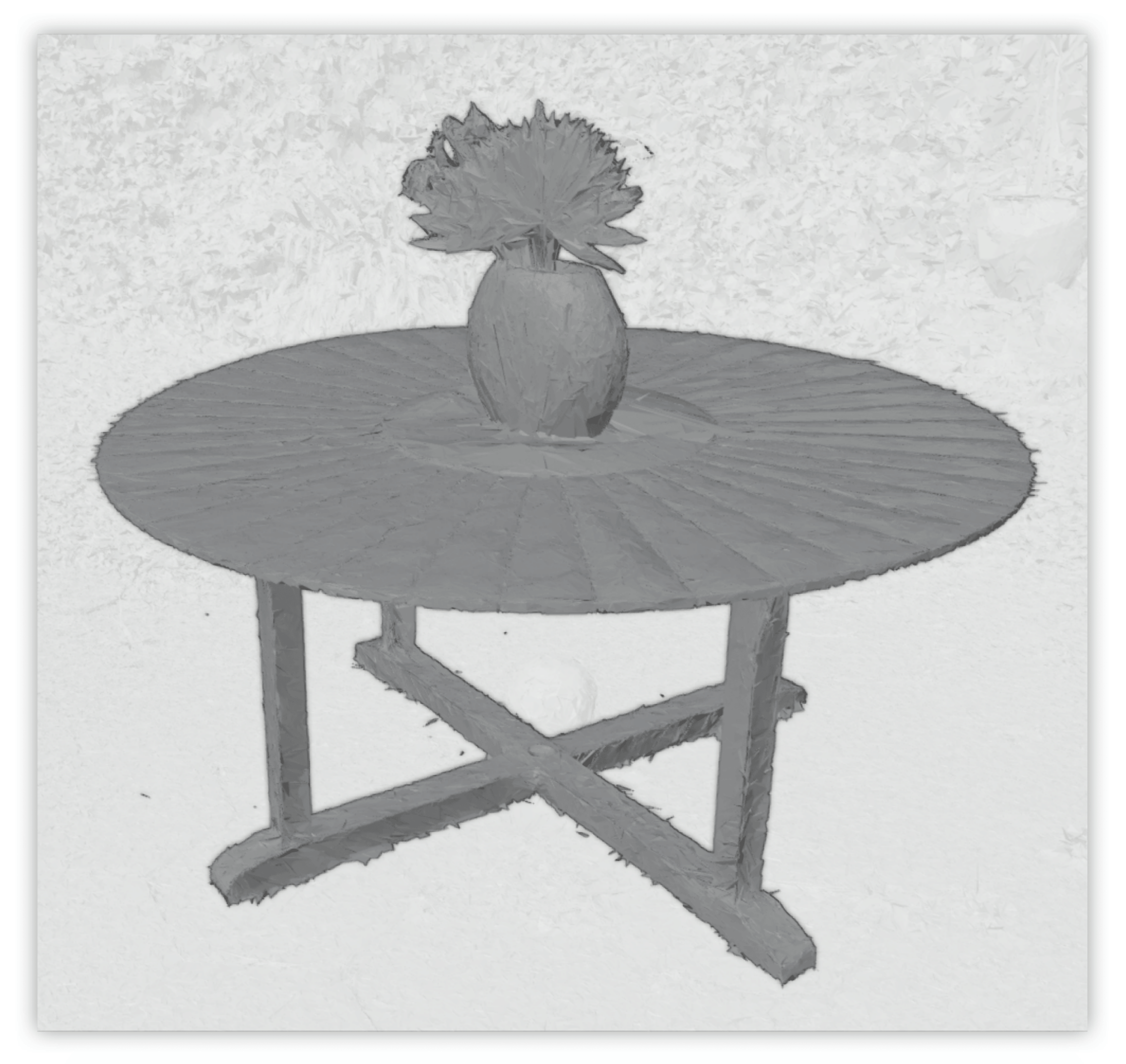}
\\[-.5em]
\scriptsize (c) ours (baseline) &
\scriptsize (d) stronger regularization \\
\end{tabular}
\caption{\small
\textbf{Regularization \textit{vs.} mesh quality.}
(a) Without any regularization, the rendered views have high visual quality, but the underlying geometry is inaccurate.
(b) The normal loss~$\mathcal{L}_n$ encourages smoother surfaces, yet without the depth regularization~$\mathcal{L}_d$, a few local regions show minor geometric inaccuracies.
(c) Our baseline model achieves both smooth and geometrically consistent surfaces.
(d) Increasing the regularization strength yields even smoother geometry, but the visual fidelity decreases as spherical harmonics fail to capture fine appearance details.
}
\label{fig:regul}
\end{figure}

\paragraph{Ablations -- \Cref{table:ablation_losses} and \Cref{fig:regul}}
While the regularization terms $\mathcal{L}_{d}$, $\mathcal{L}_{z}$, and $\mathcal{L}_{n}$ slightly reduce visual quality, they significantly improve geometric accuracy and yield smoother surfaces. The decrease in visual fidelity arises because the spherical harmonics representation cannot fully compensate for the reduced geometric flexibility, nor can it ``cheat'' by positioning triangles in non-physical configurations to better reproduce local colors or textures.
This is further confirmed by replacing spherical harmonics with simple RGB colors, which results in an average drop of about $2$ $\mathrm{PSNR}$. This highlights that expressive color representations are essential for maintaining high visual fidelity with fully opaque triangles. 
Future work could decouple geometry and appearance. By employing a more expressive appearance model, such as neural textures, the representation could achieve smoother and more accurate geometry without losing appearance information, thereby preserving high visual fidelity even under strong smoothness constraints.
More ablations are provided in the \textit{supplementary material.}
\section{Conclusions}
\label{sec:conclusion}
We introduce a differentiable framework for end-to-end optimization of mesh-based scene representations.
By reformulating the triangle parameterization to enable vertex sharing, our method produces connected meshes while maintaining high visual fidelity.
A redefined training strategy moves the optimization toward opaque triangles and connectivity, resulting in a unified representation that combines high quality appearance and accurate geometry within a compact, real-time-renderable mesh.
\methodname bridges radiance field optimization with traditional graphics pipelines, paving the way for the practical integration of neural scene representations into interactive VR applications, game engines, and simulation environments.

\paragraph{Acknowledgments}
We thank Bernhard Kerbl and George Kopanas for their helpful feedback and for proofreading the paper.
J. Held is funded by the F.R.S.-FNRS. The present research benefited from computational resources made available on Lucia, the Tier-1 supercomputer of the Walloon Region, infrastructure funded by the Walloon Region under the grant agreement n°1910247.

\clearpage
{
    \small
    \bibliographystyle{ieeenat_fullname}
    \bibliography{bib/abbreviation-short,
    bib/abbreviation-empty,
    %bib/all-removed-dois, 
    bib/all,
    bib/new_references}
}


\clearpage
\maketitlesupplementary

\subsection{Methodology.}

\begin{figure*}[th]
\vspace*{-0.5em}
\centering
\setlength{\mytmplen}{0.30\linewidth} % fit 4 images + label
\resizebox{\linewidth}{!}{ 
\begin{tabular}{c@{\hskip 0.01in}c@{\hskip 0.01in}c@{\hskip 0.01in}c@{\hskip 0.01in}c}
    % --- Stage headers ---
    & \multicolumn{2}{c}{\normalsize \textbf{Stage 1. Triangle soup optimization}} 
    & \multicolumn{2}{c}{\normalsize \textbf{Stage 2. Mesh creation \& refinement}} \\
   % --- Row 1 ---
   \rotatebox{90}{\parbox{3.5cm}{\centering \small \textbf{RGB}}}
& \zoomin{images/stages/init_garden.jpg}{0.55\mytmplen}{0.52\mytmplen}{0.824\mytmplen}{0.178\mytmplen}{1.70cm}{\mytmplen}{4.5}{red}
& \zoomin{images/stages/early_garden_13.jpg}{0.55\mytmplen}{0.52\mytmplen}{0.824\mytmplen}{0.178\mytmplen}{1.70cm}{\mytmplen}{4.5}{red}
& \zoomin{images/stages/delaunay_garden_13.jpg}{0.55\mytmplen}{0.52\mytmplen}{0.824\mytmplen}{0.178\mytmplen}{1.70cm}{\mytmplen}{4.5}{red}
& \zoomin{images/stages/final_garden_13.jpg}{0.55\mytmplen}{0.52\mytmplen}{0.824\mytmplen}{0.178\mytmplen}{1.70cm}{\mytmplen}{4.5}{red} \\

    \rotatebox{90}{\parbox{3.5cm}{\centering \small \textbf{Mesh}}}
& \zoomin{images/stages/init_point_cloud.jpg}{0.55\mytmplen}{0.42\mytmplen}{0.835\mytmplen}{0.17\mytmplen}{1.70cm}{\mytmplen}{4.5}{red}
& \zoomin{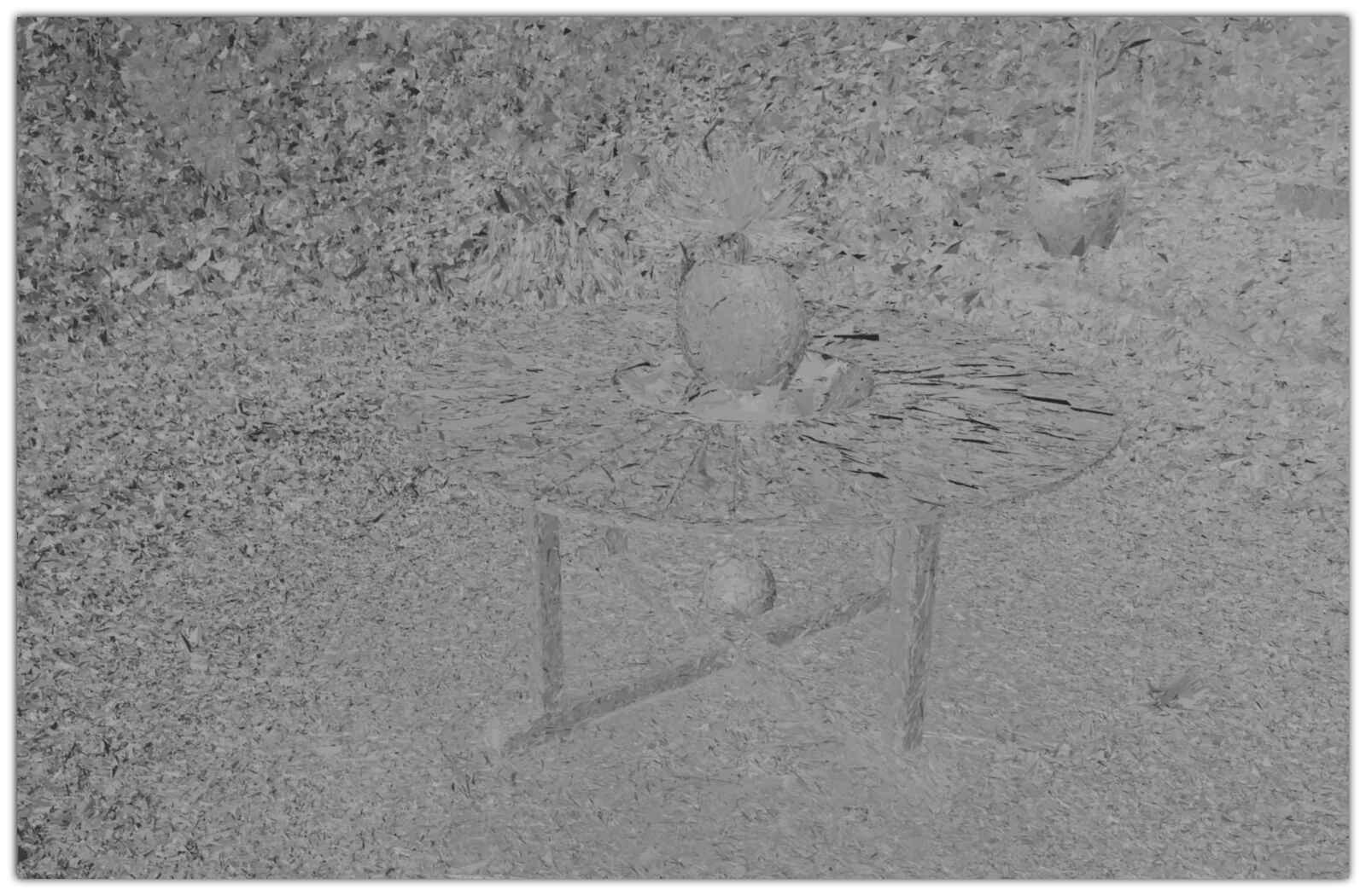}{0.55\mytmplen}{0.42\mytmplen}{0.835\mytmplen}{0.17\mytmplen}{1.70cm}{\mytmplen}{4.5}{red}
&  \zoomin{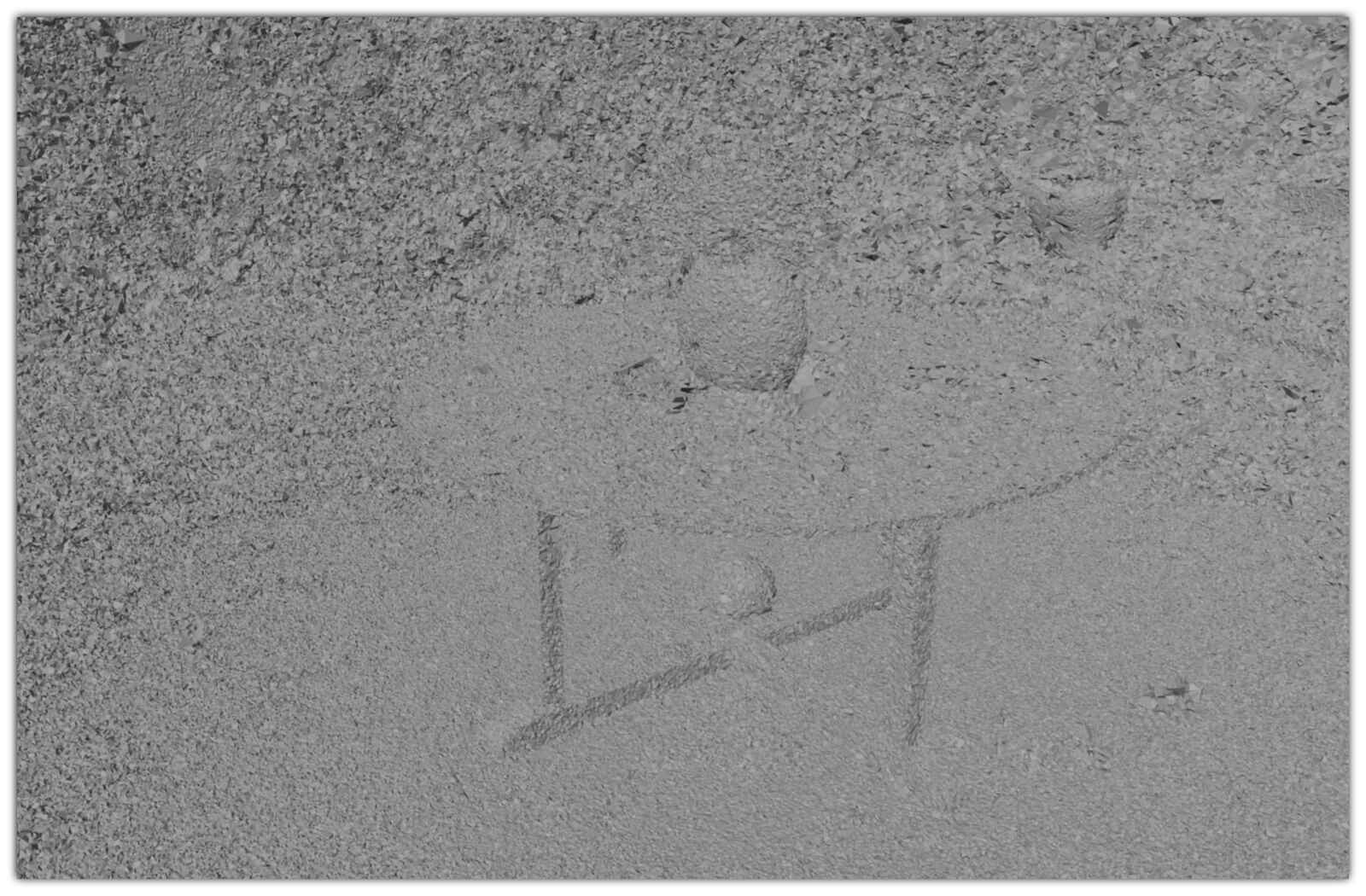}{0.55\mytmplen}{0.42\mytmplen}{0.835\mytmplen}{0.17\mytmplen}{1.70cm}{\mytmplen}{4.5}{red}
&  \zoomin{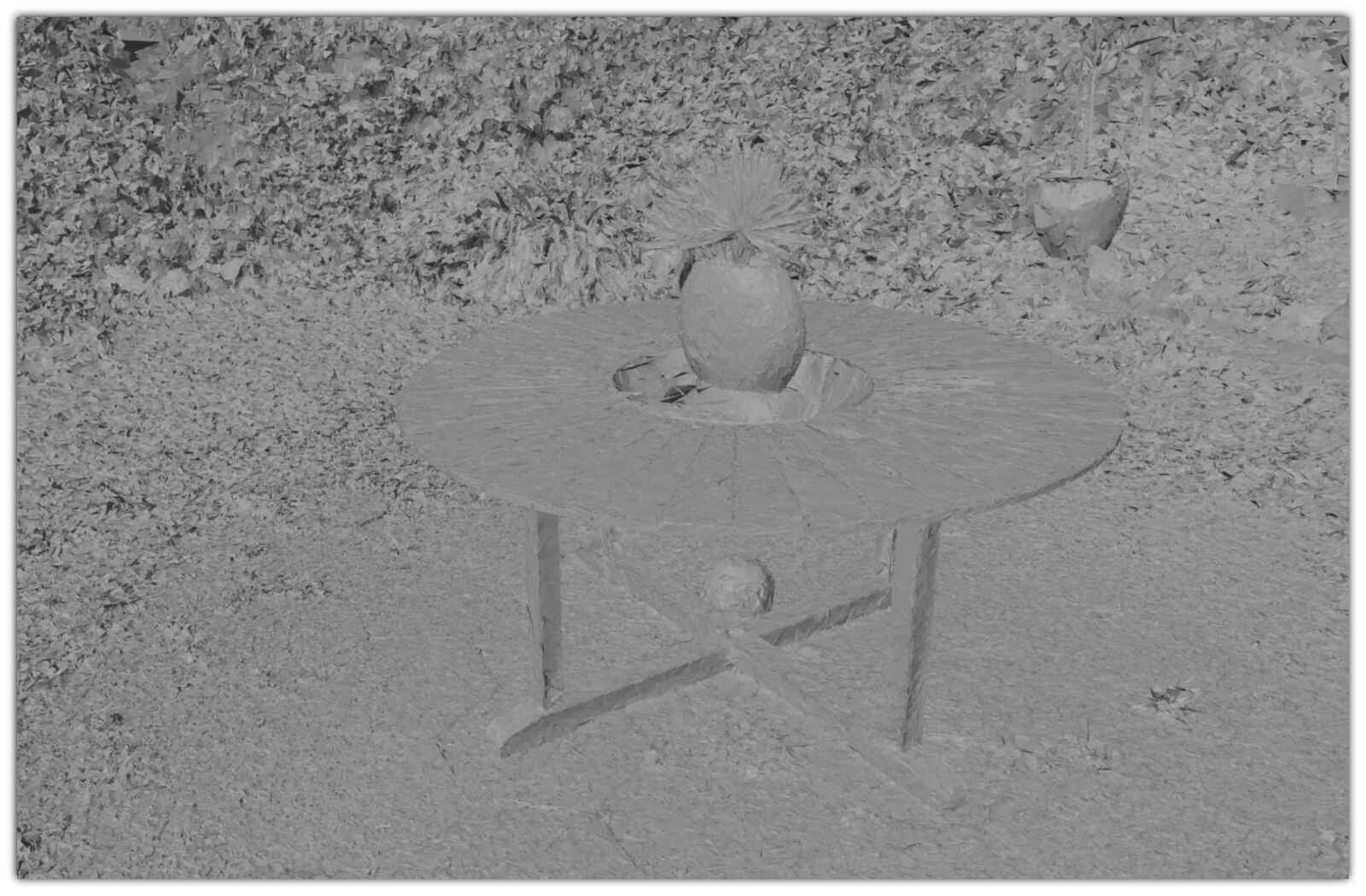}{0.55\mytmplen}{0.42\mytmplen}{0.835\mytmplen}{0.17\mytmplen}{1.70cm}{\mytmplen}{4.5}{red} \\

& \makebox[\mytmplen]{\normalsize structure-from-motion point cloud}
    & \makebox[\mytmplen]{\normalsize transparent \& disconnected triangles}
    & \makebox[\mytmplen]{\normalsize restricted Delaunay triangulation}
    & \makebox[\mytmplen]{\normalsize  opaque \& connected triangles}
   \\
\end{tabular}
}
\caption{
\textbf{From soups to triangle meshes.} 
The geometry improves significantly after refining the mesh. 
}
\label{fig:stages_supp}
\end{figure*}

\paragraph{Mesh creation \& refinement - \Cref{fig:stages_supp}}
We provide an additional visualization that illustrates both the visual and geometric improvements achieved across the different stages. \Cref{fig:stages_supp} shows how the geometry evolves by rendering the mesh without color in \textit{Blender}. Fine-tuning the output of the restricted Delaunay triangulation not only enhances visual fidelity but also leads to more accurate geometry.

\paragraph{Restricted Delaunay triangulation.}
As mentioned earlier, to construct the restricted Delaunay triangulation we first compute the standard Delaunay tetrahedralization of the input vertices, using the implementation in SciPy~\cite{virtanen2020scipy}. For each triangular face $F$ in this tetrahedralization, we identify the two tetrahedra $T^F_{1}$ and $T^F_{2}$ that are adjacent to $F$; faces on the boundary of the tetrahedralization that have only a single incident tetrahedron are discarded. We then compute the circumcenters of $T^F_{1}$ and $T^F_{2}$, which are vertices of the dual Voronoi diagram of the Delaunay tetrahedralization. The dual edge associated with $F$ is obtained by connecting the circumcenters of $T^F_{1}$ and $T^F_{2}$.

After computing the dual edges $E$ associated with the Delaunay faces, we determine which of these edges intersect the triangles in our optimized triangle soup. To perform intersection tests efficiently, we build a bounding volume hierarchy over these triangles. During traversal, when testing an edge in $E$ against a BVH node, we first check for overlap between their axis-aligned bounding boxes. If the node is internal, we continue the traversal to its children; if it is a leaf node, we apply a precise ray–triangle intersection test based on the M\"oller–Trumbore algorithm~\cite{Moller2005Fast}. Whenever an intersection is detected, we mark the corresponding Delaunay face as part of the output mesh. We implement this detection pipeline in Taichi~\cite{Hu2019Taichi}, whose automatic CPU/GPU parallelization yields a substantial speedup.

\paragraph{Hyerparameters - \Cref{tab:hyperparameters}}
The code will be released together with all hyperparameters used, ensuring that all results are fully reproducible. For completeness, we also list the most important hyperparameters here.

\begin{table}[ht]
\centering
\begin{tabular}{l|cc}
\toprule
 & Outdoor & Indoor \\
\midrule
feature learning rate & 0.0016  & 0.004 \\
opacity learning rate & 0.03 & 0.05 \\
vertex pos. learning rate & 0.0015 & 0.0015 \\
\midrule
densification start & 500  & 500 \\
densification end & 10000 & 10000 \\
densification interval & 500 & 500 \\
start pruning & 4000 & 3000 \\
pruning threshold & 0.235 & 0.235 \\
mesh creation & 11000 & 11000 \\
\midrule
$\beta_z$ & 0.00025 & 0.00025 \\
$\beta_n$ & 0.0001 & 0.0001 \\
$\beta_d$ & 0.01 & 0.0 \\
$\beta_o$ & 2e-06 & 0.0 \\

\bottomrule
\end{tabular}
\caption{\small \textbf{Hyperparameters}
List of the most important hyperparameters.
}
\label{tab:hyperparameters}
\end{table}

\begin{table}[ht]
    \centering
    \tabcolsep=0.1cm
    \resizebox{0.45\columnwidth}{!}{      
    \begin{tabular}{l|cc}
     & Truck & Train \\
    \midrule
    PSNR$\uparrow$ & 22.32 & 18.72 \\
    LPIPS$\downarrow$ & 0.237 & 0.337 \\
    SSIM$\uparrow$ & 0.799 & 0.693 \\
    \end{tabular}
    }    
    \footnotesize
    \caption{Per scene LPIPS, PSNR, and SSIM scores for the Truck and Train scenes of the T\&T dataset.}
    \label{tab:4a}
\end{table}

\begin{table}[ht]
    \centering
    \tabcolsep=0.1cm
    \resizebox{0.98\columnwidth}{!}{      
    \begin{tabular}{l|ccccc}
         & Bicycle & Flowers & Garden & Stump & Treehill \\
        \midrule
        PSNR$\uparrow$  & 23.04 & 19.34 & 24.70 & 24.78 & 20.53 \\
        LPIPS$\downarrow$ & 0.348 & 0.417 & 0.217 & 0.316 & 0.428 \\
        SSIM$\uparrow$  & 0.641 & 0.480 & 0.762 & 0.678 & 0.540 \\
    \end{tabular}
    }
    \caption{Per-scene PSNR, LPIPS, and SSIM scores for outdoor MipNeRF360 scenes.}
    \label{tab:mipnerf_outdoor}
\end{table}

\begin{table}[ht]
    \centering
    \tabcolsep=0.1cm
    \resizebox{0.8\columnwidth}{!}{      
    \begin{tabular}{l|cccc}
         & Room & Counter & Kitchen & Bonsai \\
        \midrule
        PSNR$\uparrow$  & 28.52 & 26.51 & 27.42 & 28.19 \\
        LPIPS$\downarrow$ & 0.271 & 0.279 & 0.227 & 0.294 \\
        SSIM$\uparrow$  & 0.873 & 0.846 & 0.858 & 0.876 \\
    \end{tabular}
    }
    \caption{Per-scene PSNR, LPIPS, and SSIM scores for indoor MipNeRF360 scenes.}
    \label{tab:mipnerf_indoor}
\end{table}

\subsection{Addition experimental results}
\begin{figure*}[t]
\vspace*{-1em}
\centering
\setlength{\mytmplen}{0.20\linewidth} % fit 5 images + label
\resizebox{\linewidth}{!}{ 
\begin{tabular}{c@{\hskip 0.01in}c@{\hskip 0.01in}c@{\hskip 0.01in}c@{\hskip 0.01in}c}
    
    & \makebox[\mytmplen]{\scriptsize Ground Truth} &
      \makebox[\mytmplen]{\scriptsize \textbf{\methodname}} &
      \makebox[\mytmplen]{\scriptsize MiLo} &
      \makebox[\mytmplen]{\scriptsize Triangle Splatting$\dagger$} 
      \vspace*{-.5em}
   \\

    \rotatebox{90}{\parbox{2.2cm}{\centering \scriptsize Truck}} &
    \zoomin{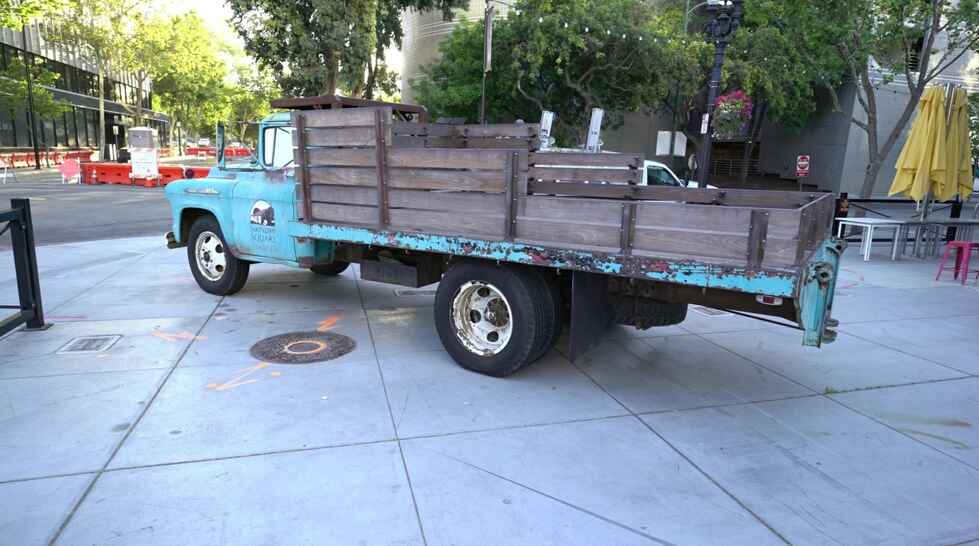}{0.9\mytmplen}{0.32\mytmplen}{0.147\mytmplen}{0.15\mytmplen}{1.0cm}{\mytmplen}{2.5}{red}
    & \zoomin{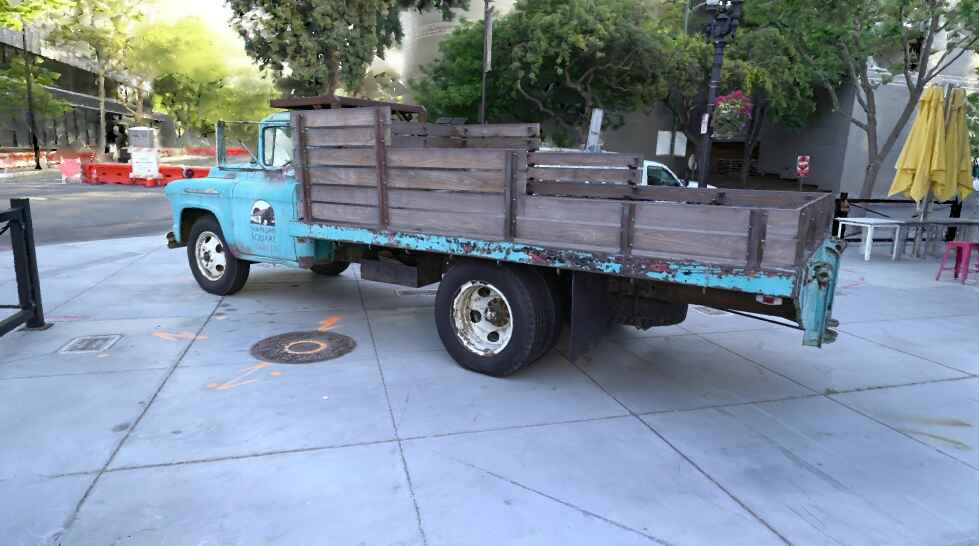}{0.9\mytmplen}{0.32\mytmplen}{0.147\mytmplen}{0.15\mytmplen}{1.0cm}{\mytmplen}{2.5}{red}
    & \zoomin{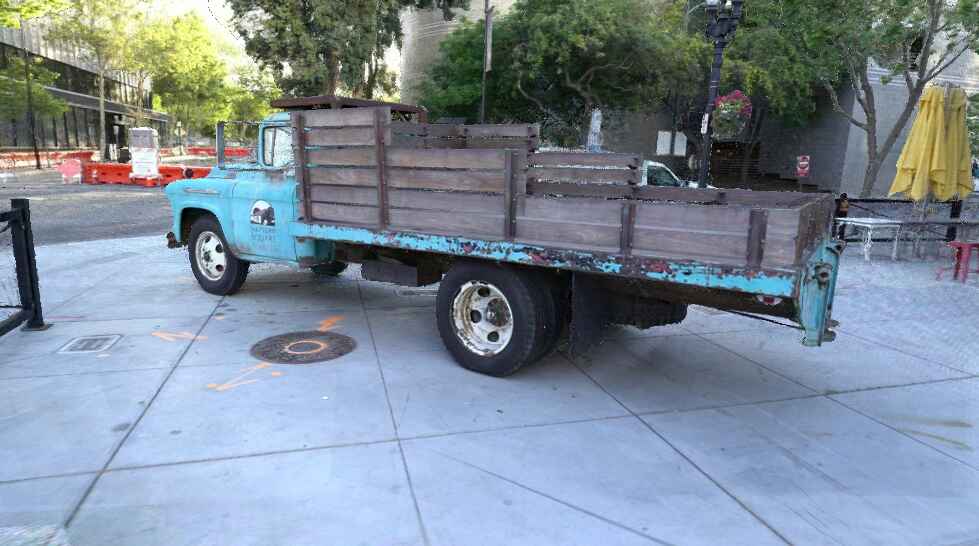}{0.9\mytmplen}{0.32\mytmplen}{0.147\mytmplen}{0.15\mytmplen}{1.0cm}{\mytmplen}{2.5}{red}
    & \zoomin{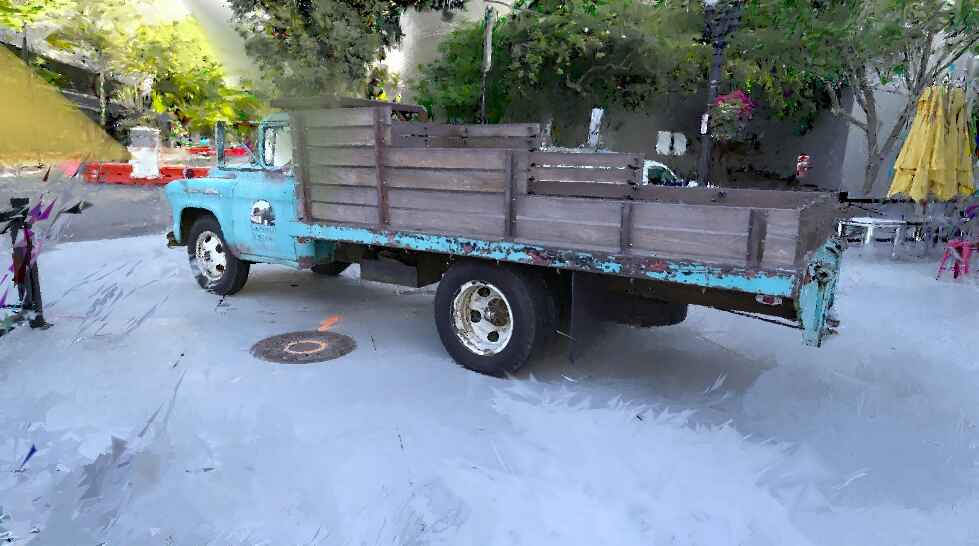}{0.9\mytmplen}{0.32\mytmplen}{0.147\mytmplen}{0.15\mytmplen}{1.0cm}{\mytmplen}{2.5}{red}
    \\

          \rotatebox{90}{\parbox{2.2cm}{\centering \scriptsize Train}} &
    \zoomin{images/qualitative_results/train_gt2.jpg}{0.57\mytmplen}{0.37\mytmplen}{0.147\mytmplen}{0.15\mytmplen}{1.0cm}{\mytmplen}{2.5}{red}
    & \zoomin{images/qualitative_results/train_ms2.jpg}{0.57\mytmplen}{0.37\mytmplen}{0.147\mytmplen}{0.15\mytmplen}{1.0cm}{\mytmplen}{2.5}{red}
    & \zoomin{images/qualitative_results/train_milo2.jpg}{0.57\mytmplen}{0.37\mytmplen}{0.147\mytmplen}{0.15\mytmplen}{1.0cm}{\mytmplen}{2.5}{red}
    & \zoomin{images/qualitative_results/train_ts.jpg}{0.57\mytmplen}{0.37\mytmplen}{0.147\mytmplen}{0.15\mytmplen}{1.0cm}{\mytmplen}{2.5}{red}
    \\

       \rotatebox{90}{\parbox{2.2cm}{\centering \scriptsize Train}} &
    \zoomin{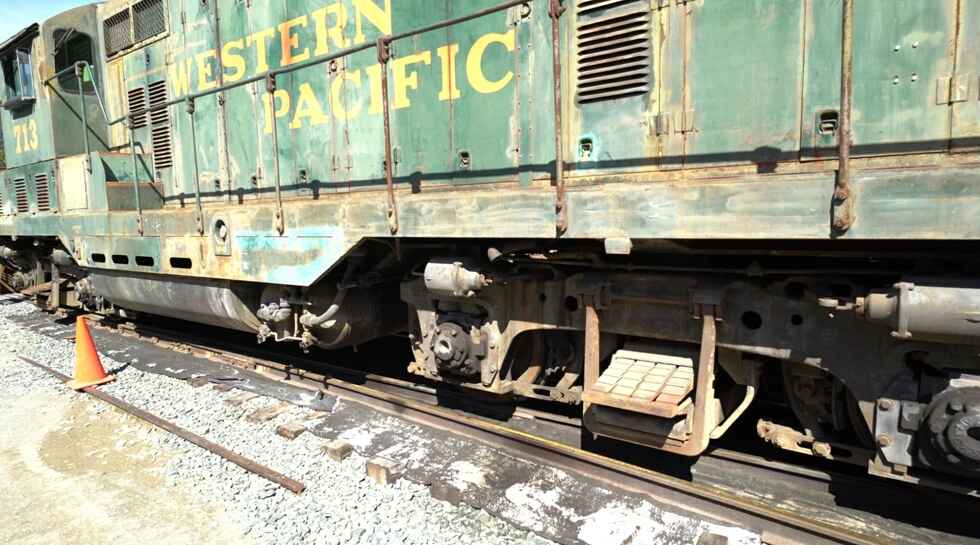}{0.57\mytmplen}{0.47\mytmplen}{0.147\mytmplen}{0.15\mytmplen}{1.0cm}{\mytmplen}{2.5}{red}
    & \zoomin{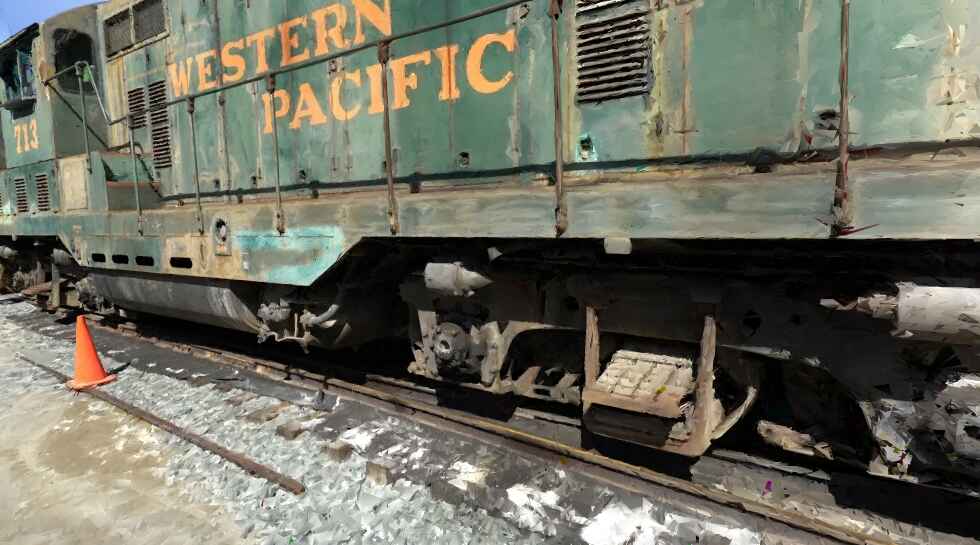}{0.57\mytmplen}{0.47\mytmplen}{0.147\mytmplen}{0.15\mytmplen}{1.0cm}{\mytmplen}{2.5}{red}
    & \zoomin{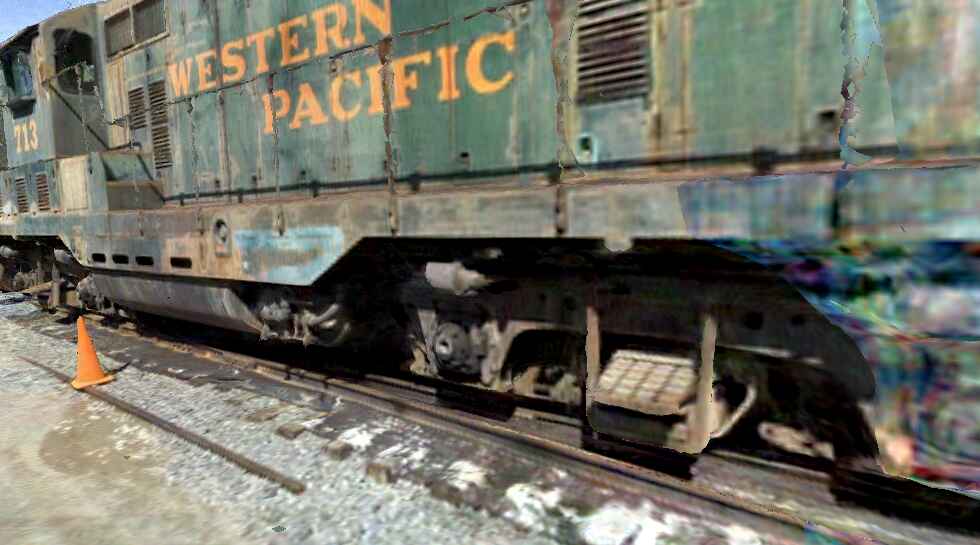}{0.57\mytmplen}{0.47\mytmplen}{0.147\mytmplen}{0.15\mytmplen}{1.0cm}{\mytmplen}{2.5}{red}
    & \zoomin{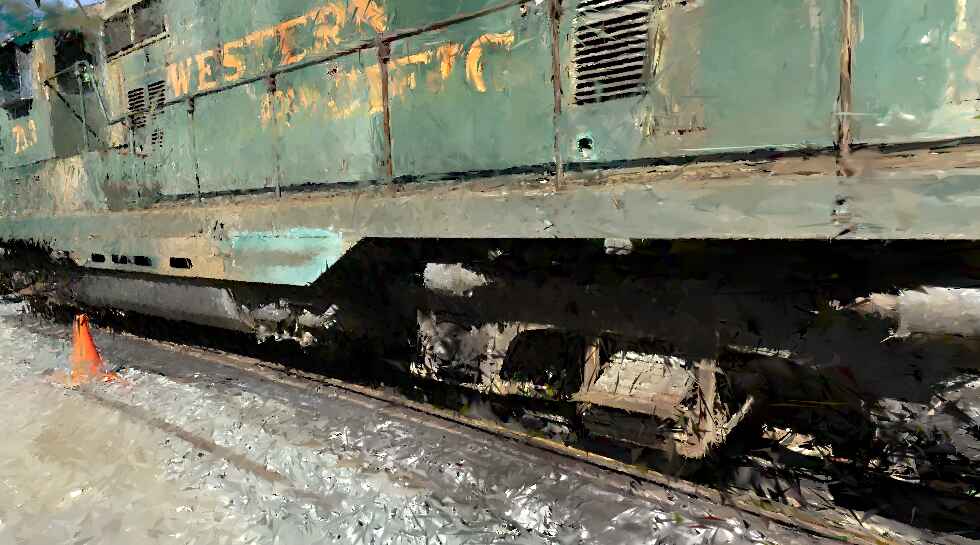}{0.57\mytmplen}{0.47\mytmplen}{0.147\mytmplen}{0.15\mytmplen}{1.0cm}{\mytmplen}{2.5}{red}

    \\

\end{tabular}
}

\caption{\small \textbf{More qualitative results on the \textit{Tanks and Temples} dataset.} 
\methodname reconstructs more details and finer structures compared to concurrent works. While MiLo achieves a higher PSNR, \methodname produces fewer artifacts and less noisy renderings, which results in a significantly lower LPIPS and a higher SSIM.
}
\label{fig:supp_qualityresults}
\end{figure*}

\begin{figure*}[t]
\vspace*{-1em}
\centering
\setlength{\mytmplen}{0.20\linewidth} % fit 5 images + label
\resizebox{\linewidth}{!}{ 
\begin{tabular}{c@{\hskip 0.01in}c@{\hskip 0.01in}c@{\hskip 0.01in}c@{\hskip 0.01in}c}
    
    & \makebox[\mytmplen]{\scriptsize Ground Truth} &
      \makebox[\mytmplen]{\scriptsize \textbf{\methodname}} &
      \makebox[\mytmplen]{\scriptsize MiLo} &
      \makebox[\mytmplen]{\scriptsize Triangle Splatting$\dagger$} 
   \\

    \rotatebox{90}{\parbox{2.2cm}{\centering \scriptsize Bicycle}} % view 12
    & \zoomin{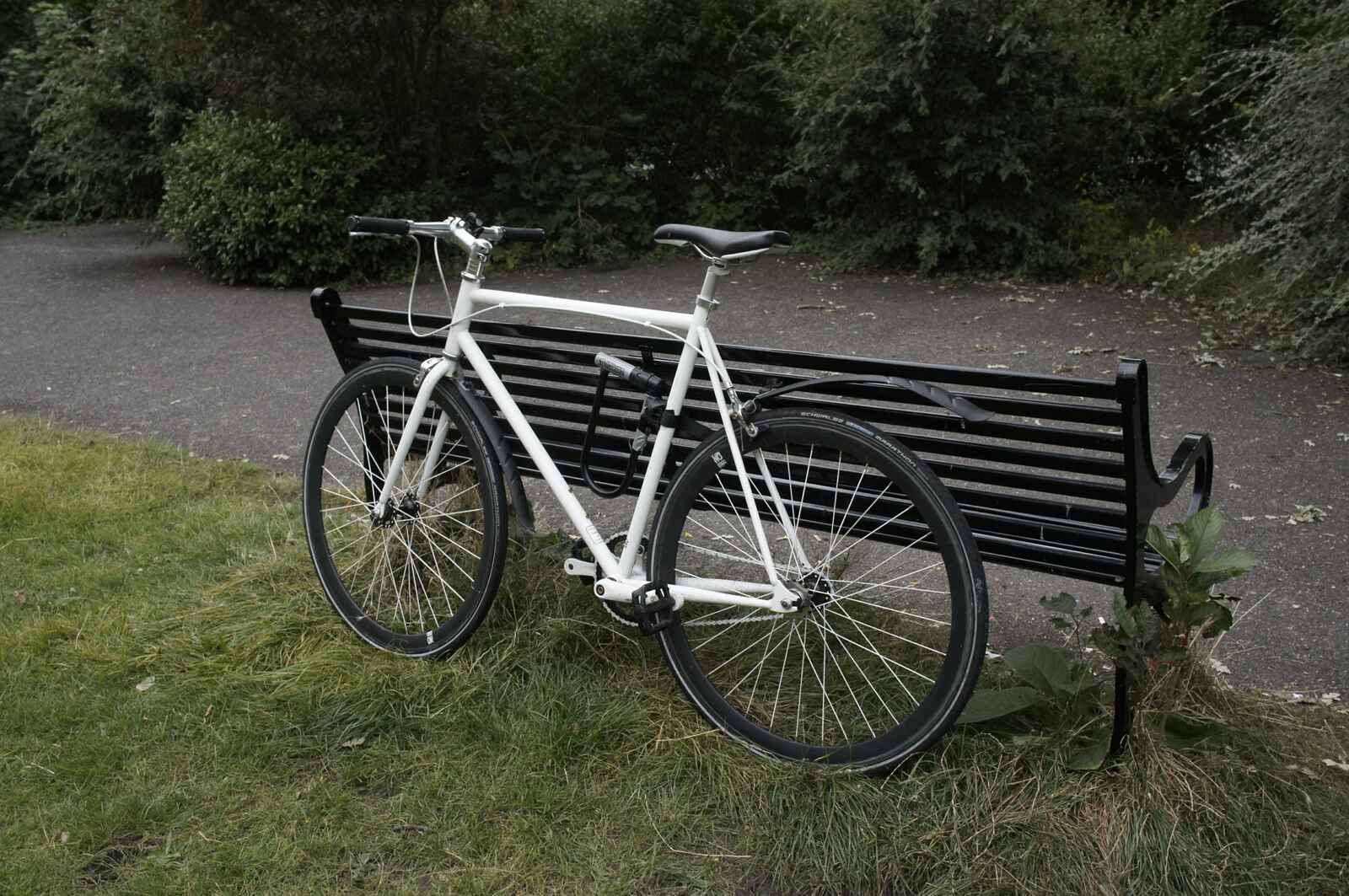}{0.65\mytmplen}{0.25\mytmplen}{0.18\mytmplen}{0.178\mytmplen}{1.2cm}{\mytmplen}{2.5}{red}
    & \zoomin{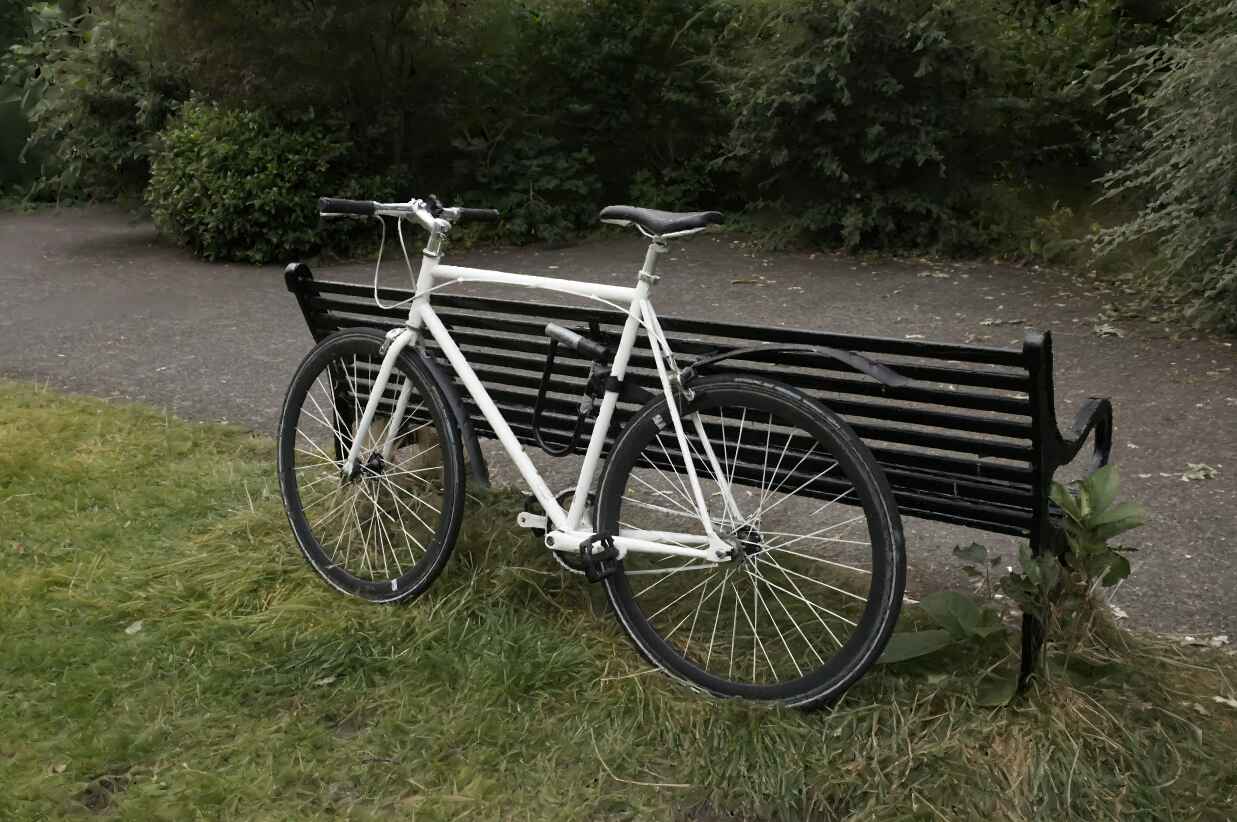}{0.65\mytmplen}{0.25\mytmplen}{0.18\mytmplen}{0.178\mytmplen}{1.2cm}{\mytmplen}{2.5}{red}
    & \zoomin{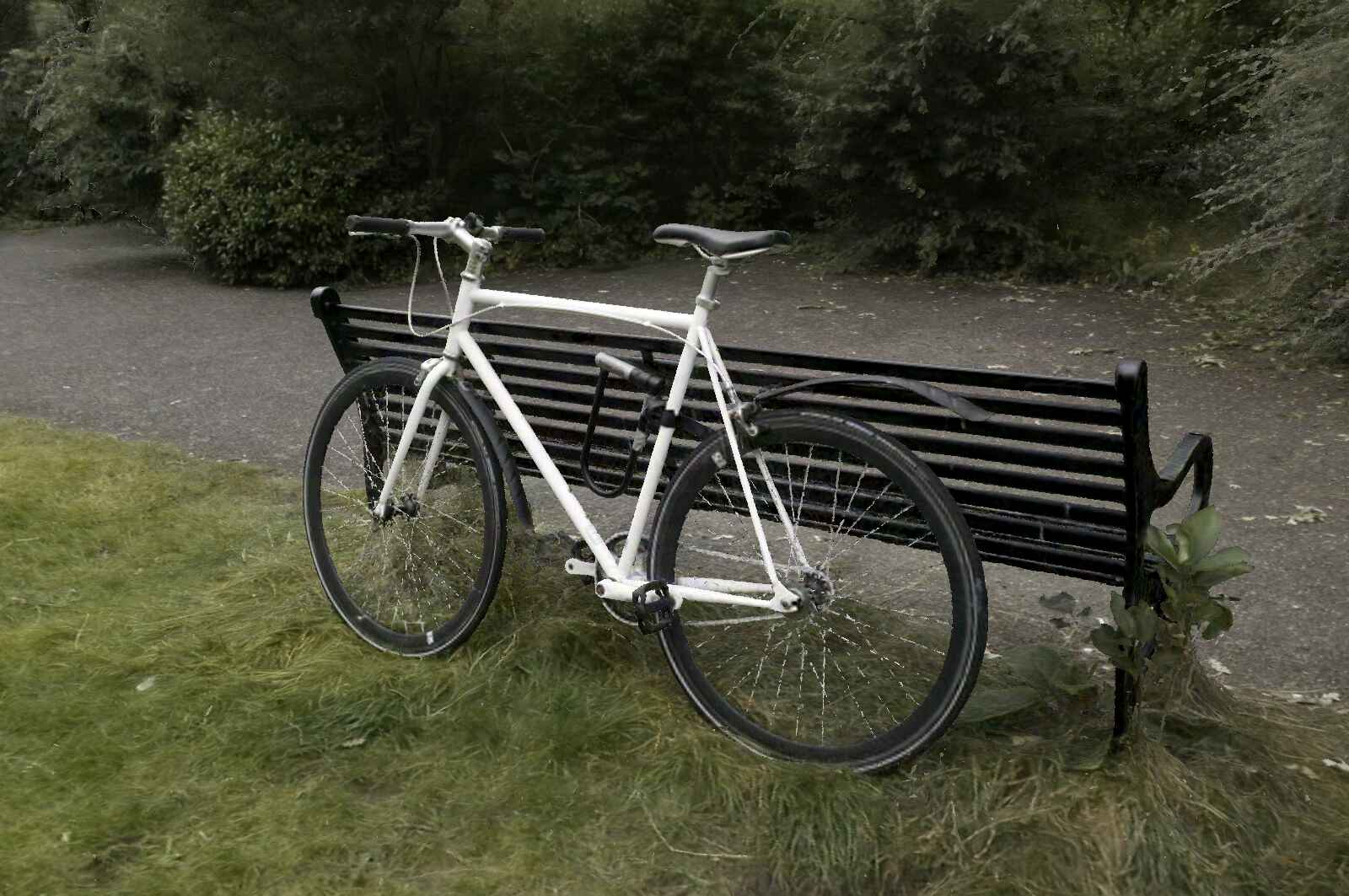}{0.65\mytmplen}{0.25\mytmplen}{0.18\mytmplen}{0.178\mytmplen}{1.2cm}{\mytmplen}{2.5}{red}
    & \zoomin{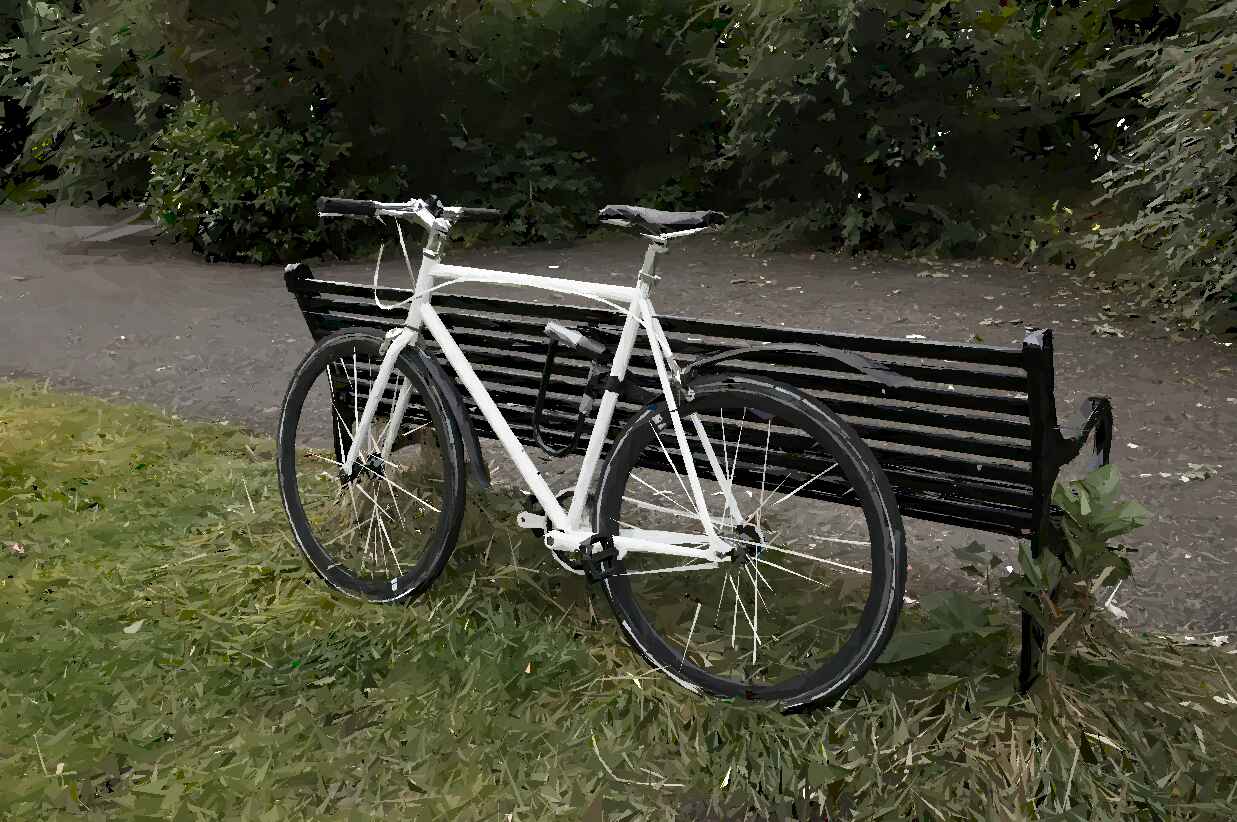}{0.65\mytmplen}{0.25\mytmplen}{0.18\mytmplen}{0.178\mytmplen}{1.2cm}{\mytmplen}{2.5}{red}
    \\

    \rotatebox{90}{\parbox{2.2cm}{\centering \scriptsize Bonsai}} % view 5
    & \zoomin{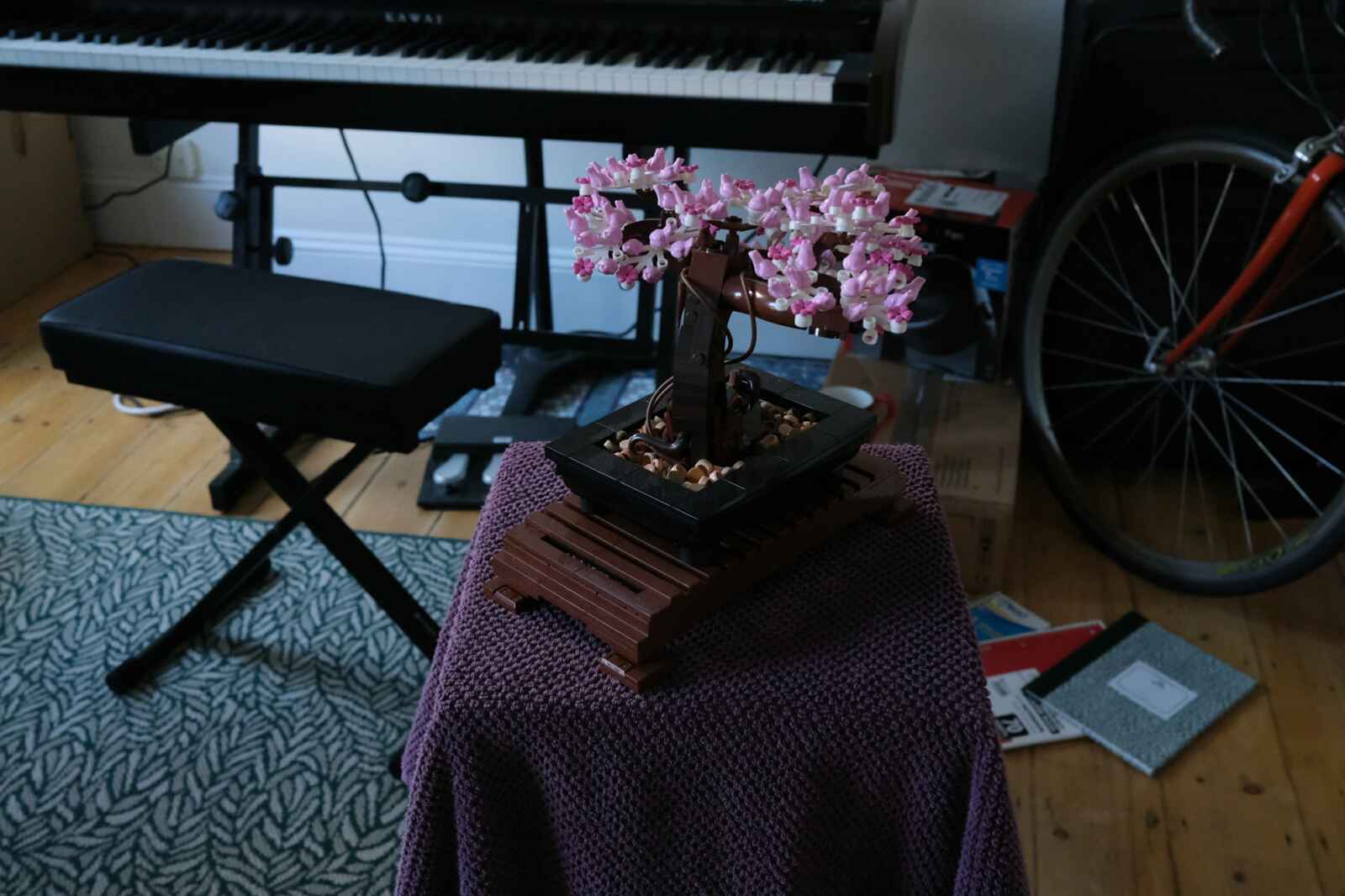}{0.9\mytmplen}{0.32\mytmplen}{0.18\mytmplen}{0.178\mytmplen}{1.2cm}{\mytmplen}{4.5}{red}
    & \zoomin{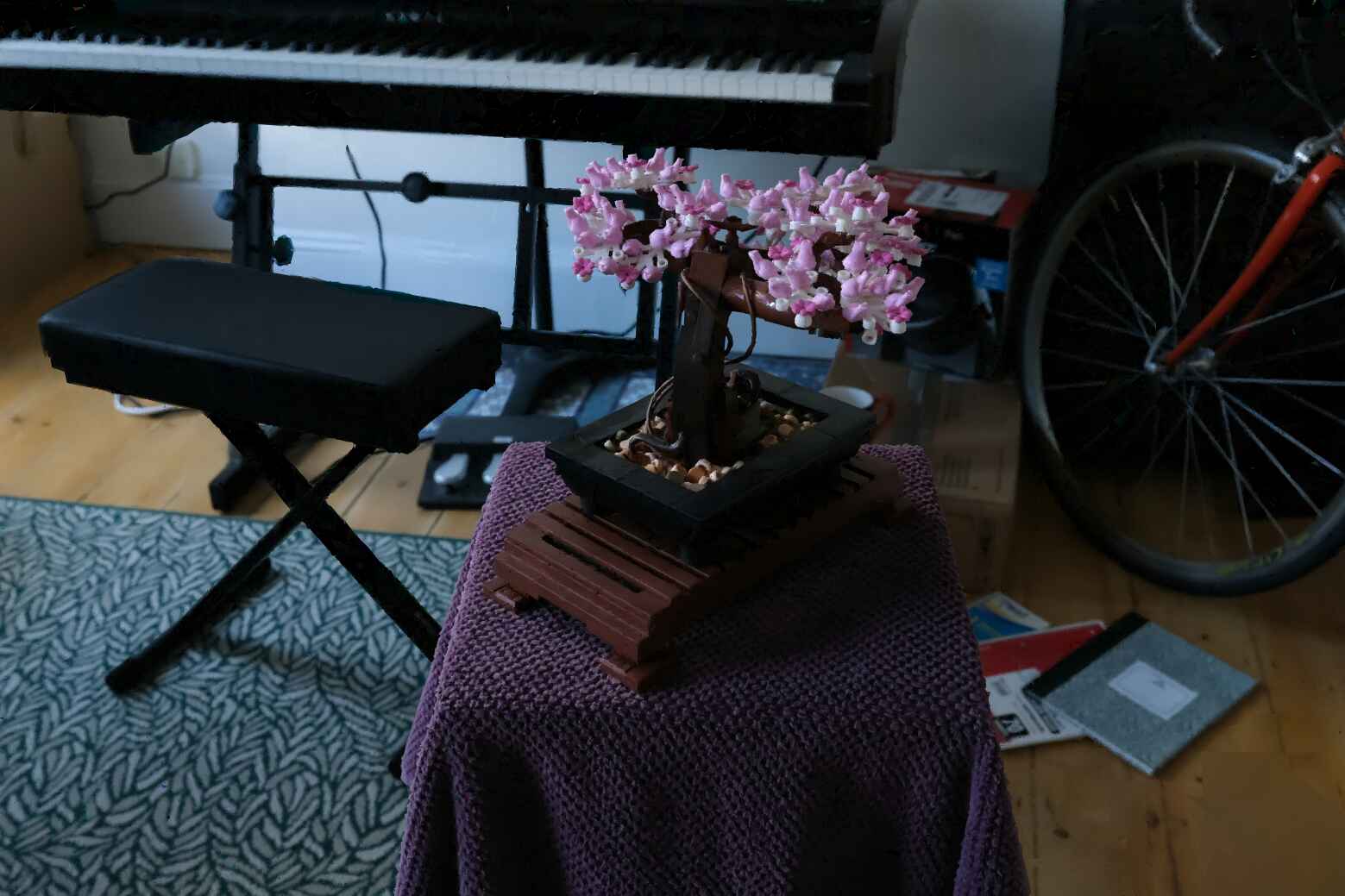}{0.9\mytmplen}{0.32\mytmplen}{0.18\mytmplen}{0.178\mytmplen}{1.2cm}{\mytmplen}{4.5}{red}
    & \zoomin{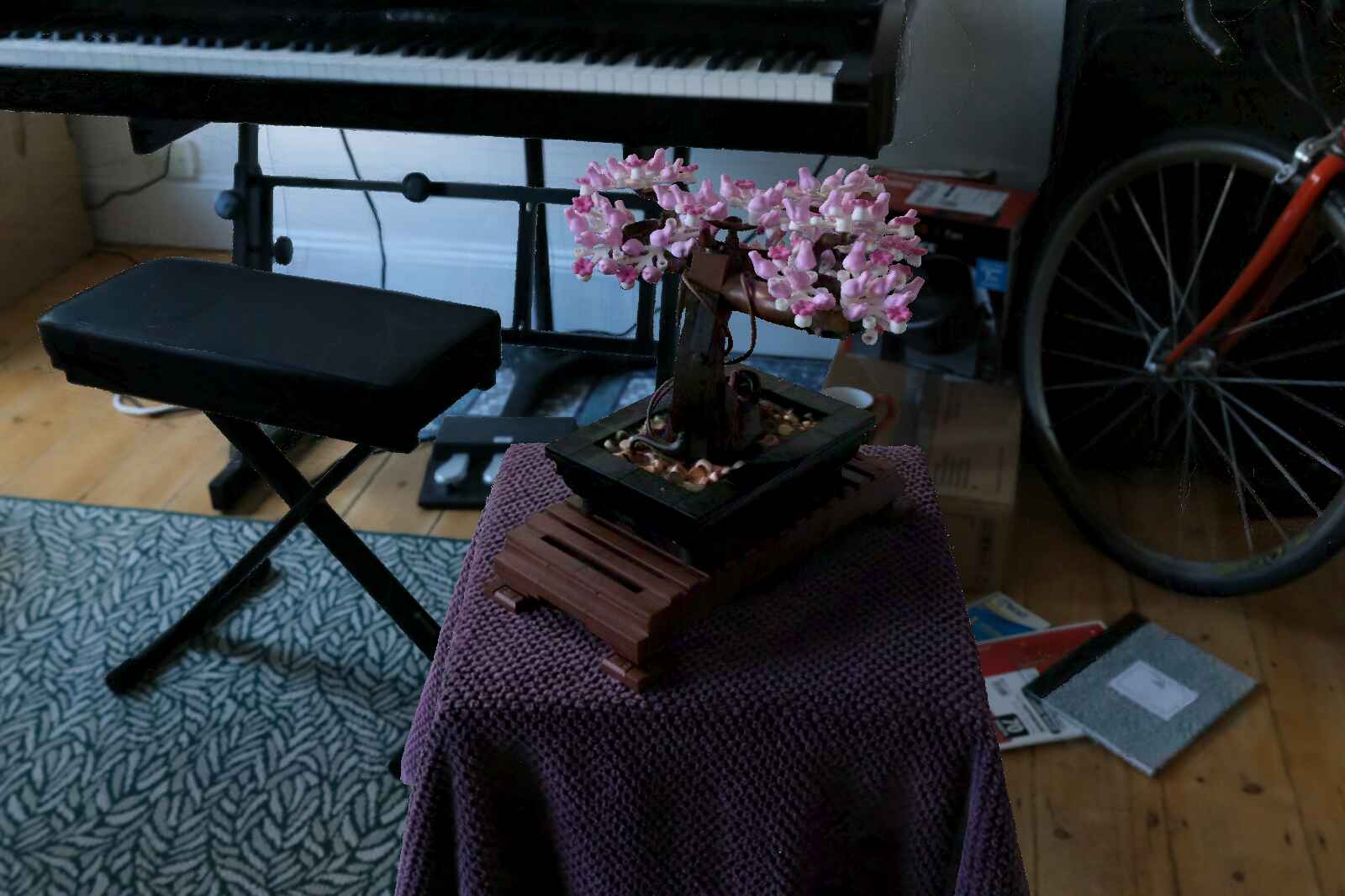}{0.9\mytmplen}{0.32\mytmplen}{0.18\mytmplen}{0.178\mytmplen}{1.2cm}{\mytmplen}{4.5}{red}
    & \zoomin{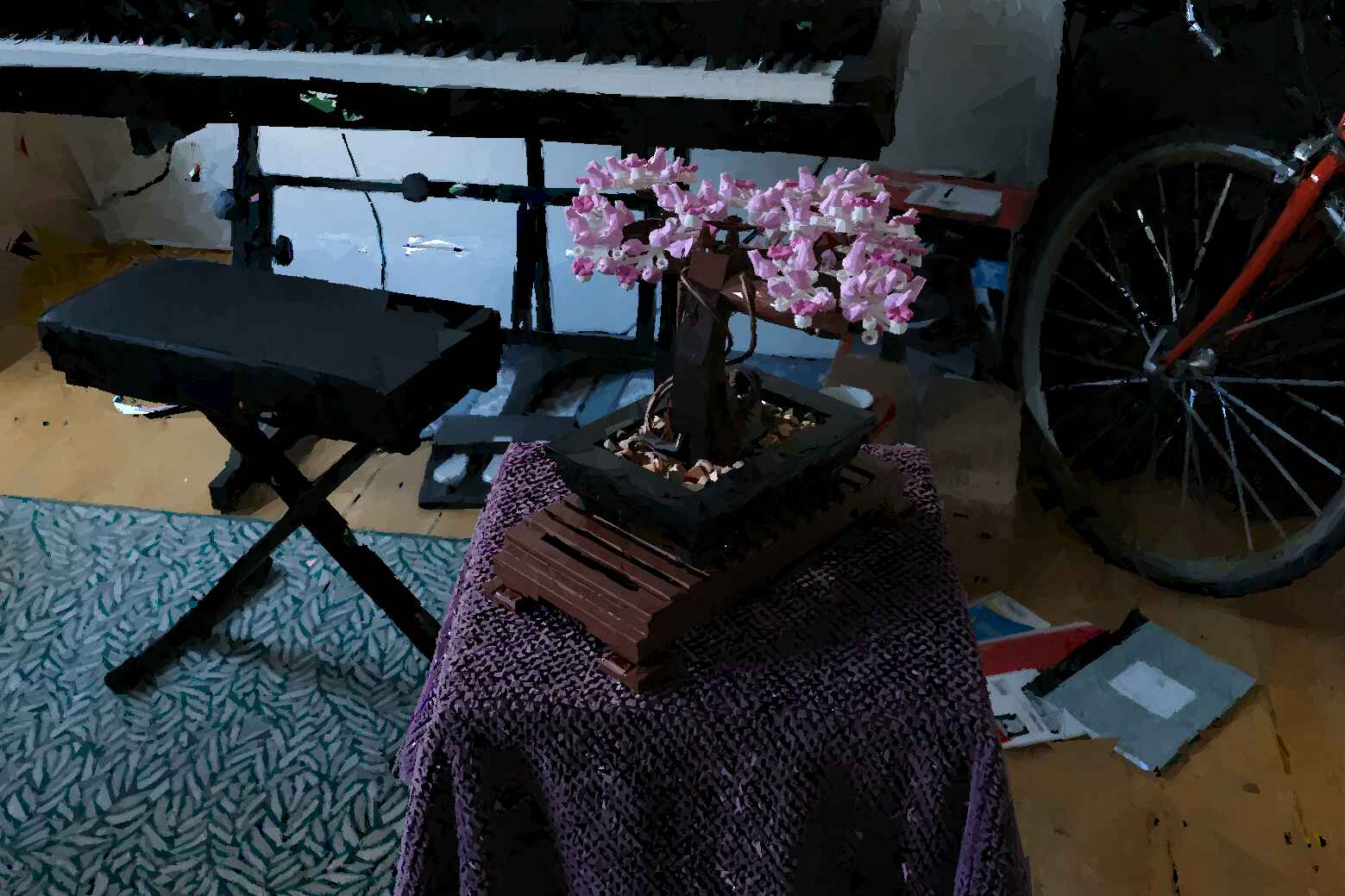}{0.9\mytmplen}{0.32\mytmplen}{0.18\mytmplen}{0.178\mytmplen}{1.2cm}{\mytmplen}{4.5}{red}
    \\

    \rotatebox{90}{\parbox{2.2cm}{\centering \scriptsize Stump}} &
    \zoomin{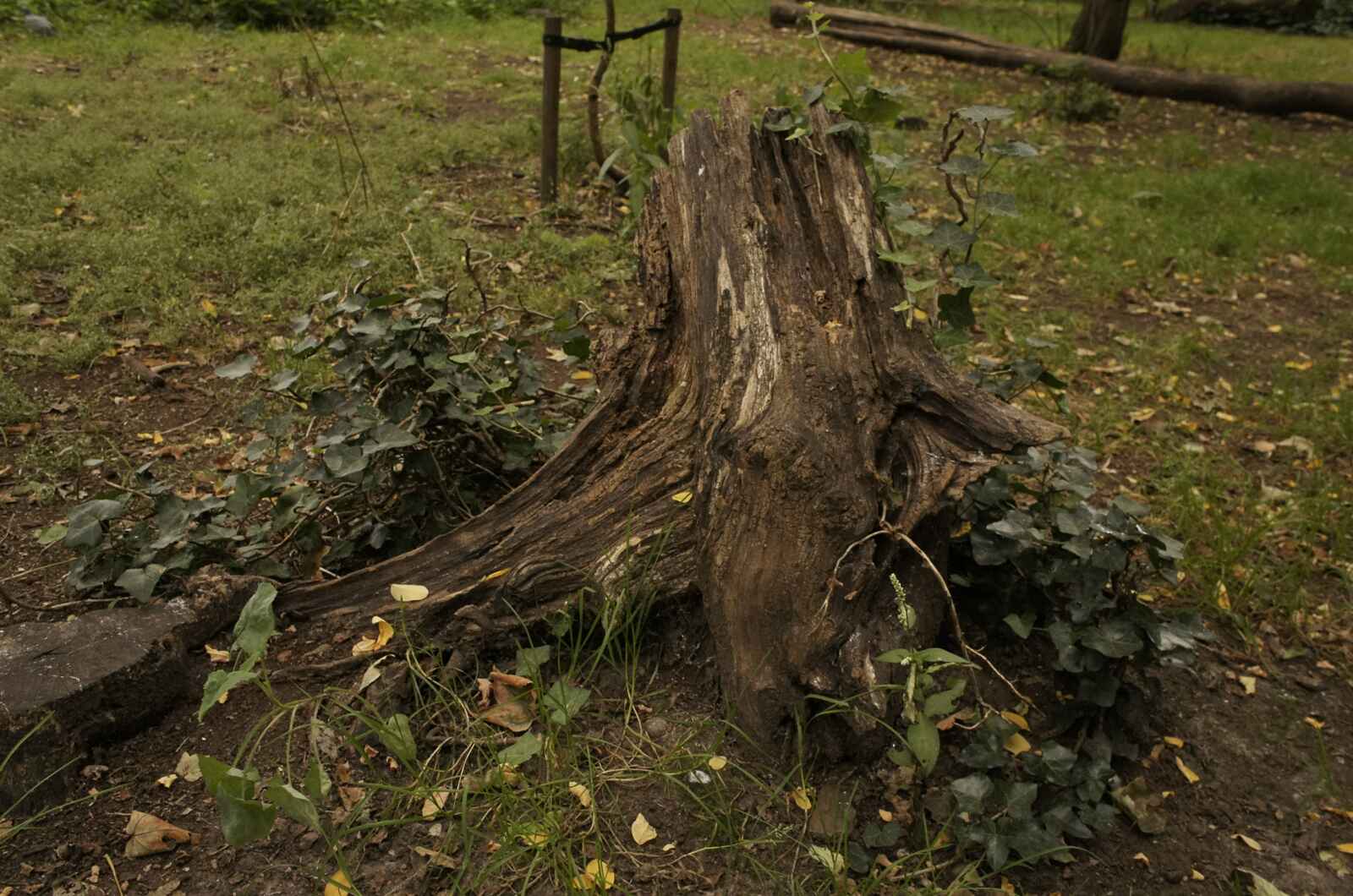}{0.4\mytmplen}{0.13\mytmplen}{0.82\mytmplen}{0.181\mytmplen}{1.2cm}{\mytmplen}{4.5}{red}
   & \zoomin{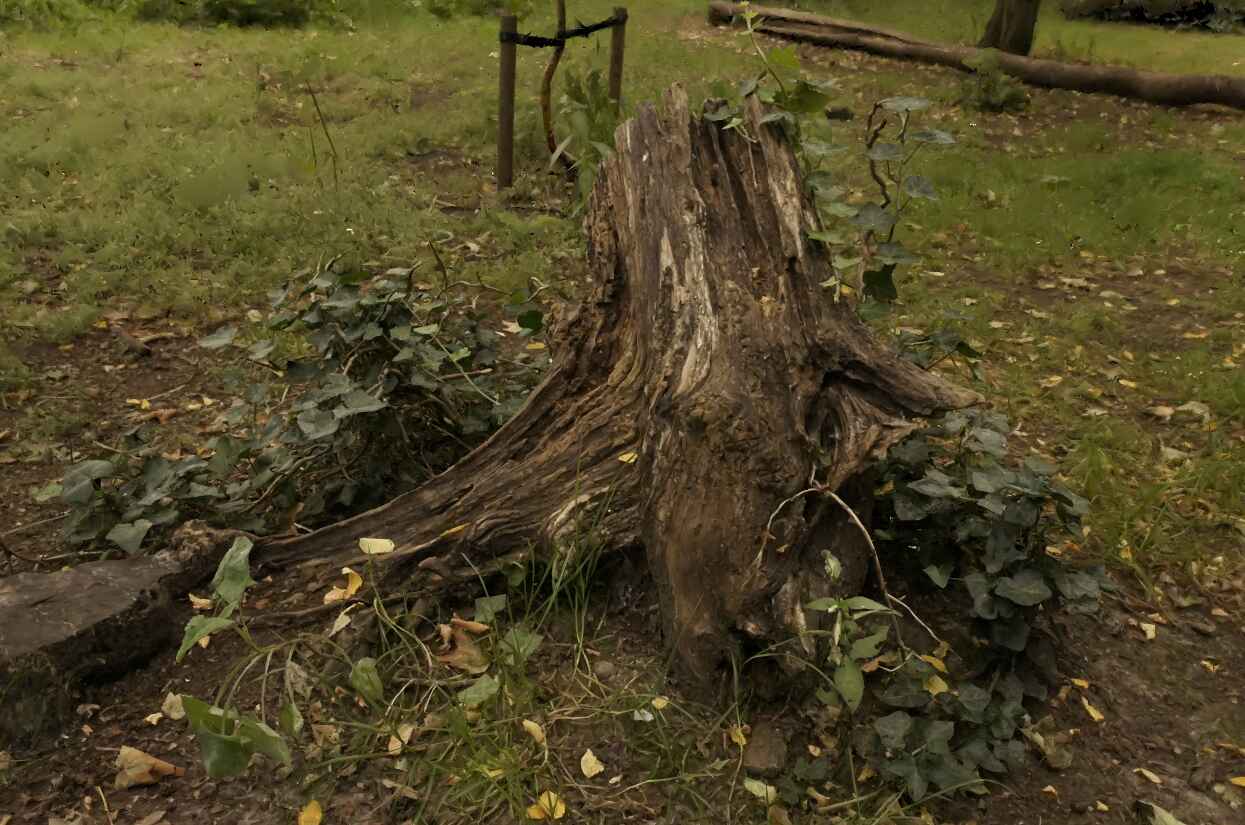}{0.4\mytmplen}{0.13\mytmplen}{0.82\mytmplen}{0.181\mytmplen}{1.2cm}{\mytmplen}{4.5}{red}
   &  \zoomin{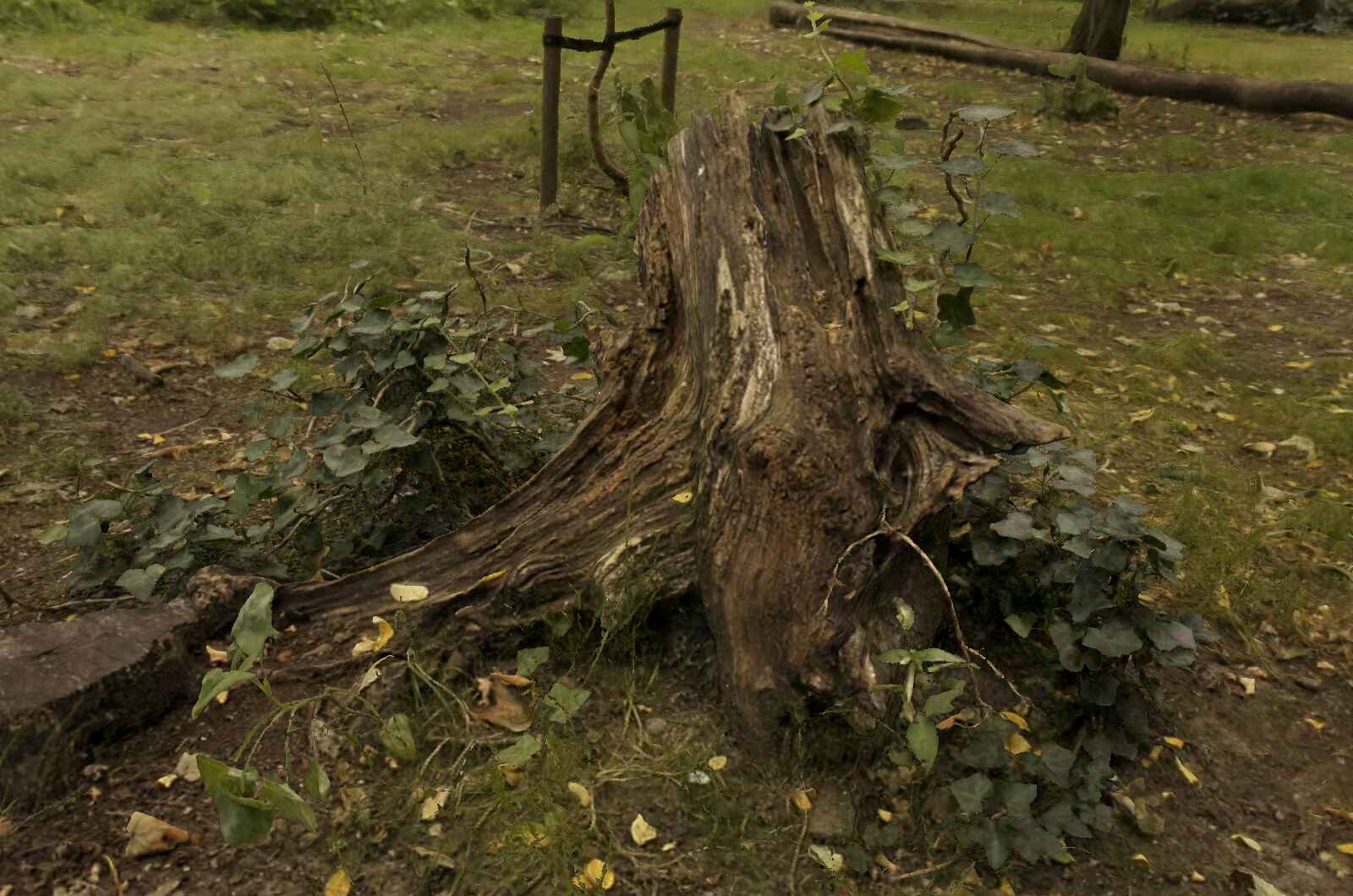}{0.4\mytmplen}{0.13\mytmplen}{0.82\mytmplen}{0.181\mytmplen}{1.2cm}{\mytmplen}{4.5}{red}
   &  \zoomin{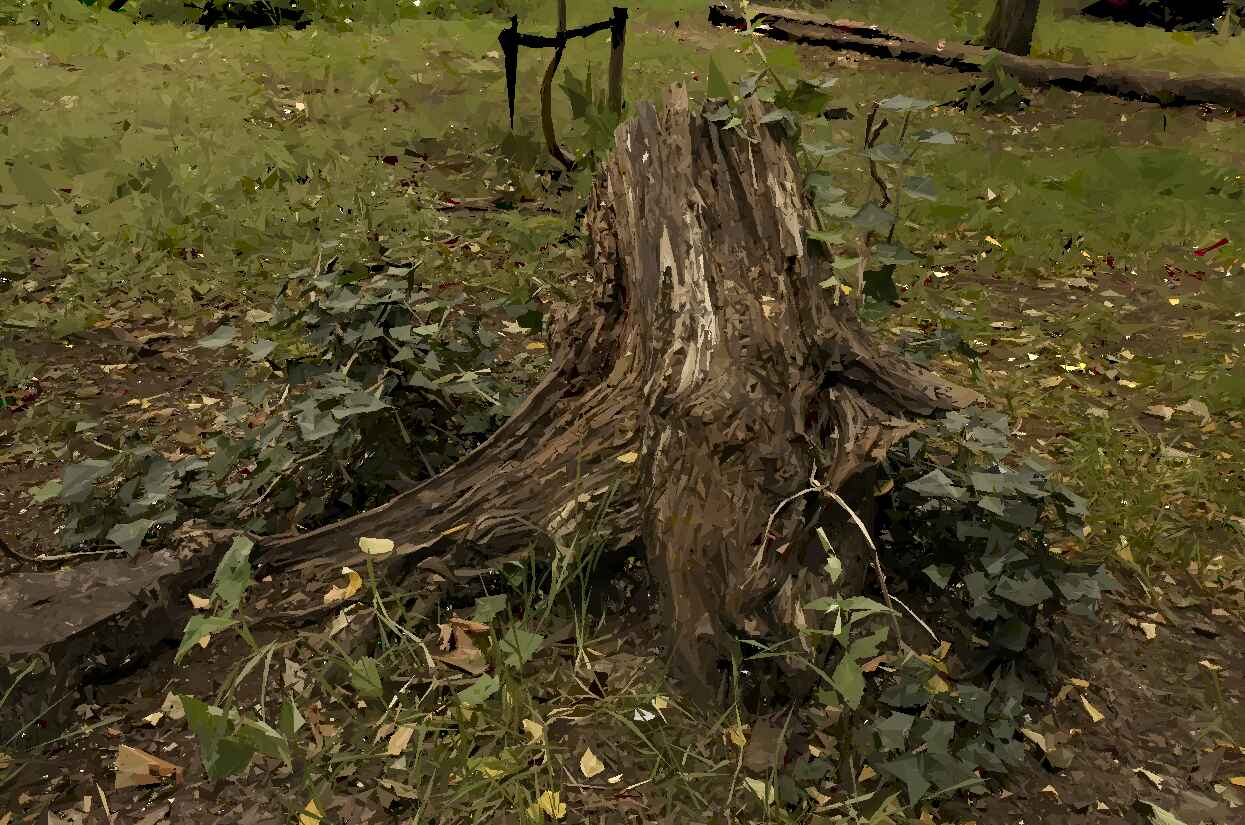}{0.4\mytmplen}{0.13\mytmplen}{0.82\mytmplen}{0.181\mytmplen}{1.2cm}{\mytmplen}{4.5}{red}
   
    \\
    \rotatebox{90}{\parbox{2.2cm}{\centering \scriptsize Garden}} & % view 14
    \zoomin{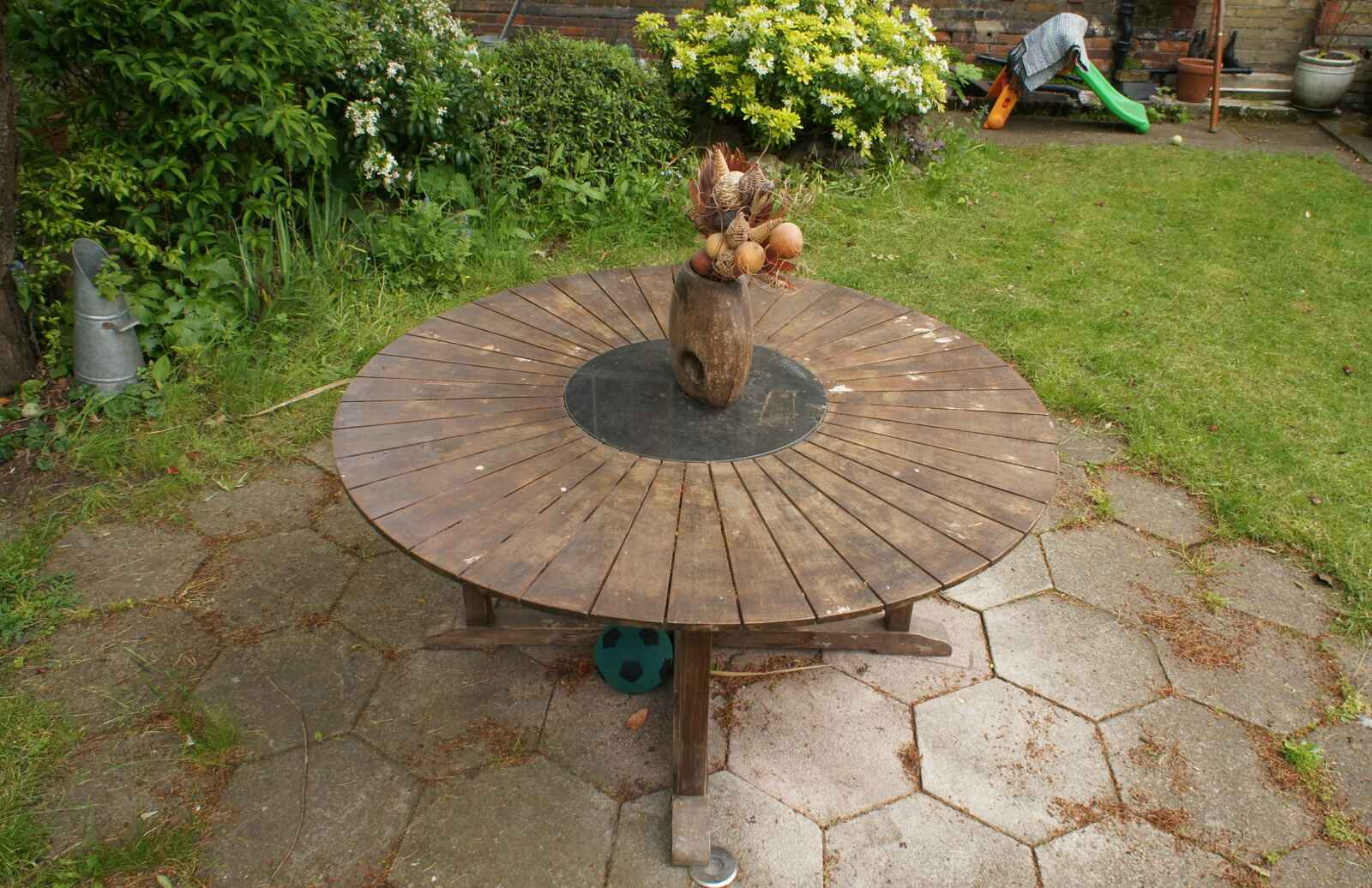}{0.56\mytmplen}{0.48\mytmplen}{0.82\mytmplen}{0.181\mytmplen}{1.2cm}{\mytmplen}{4.5}{red}
   & \zoomin{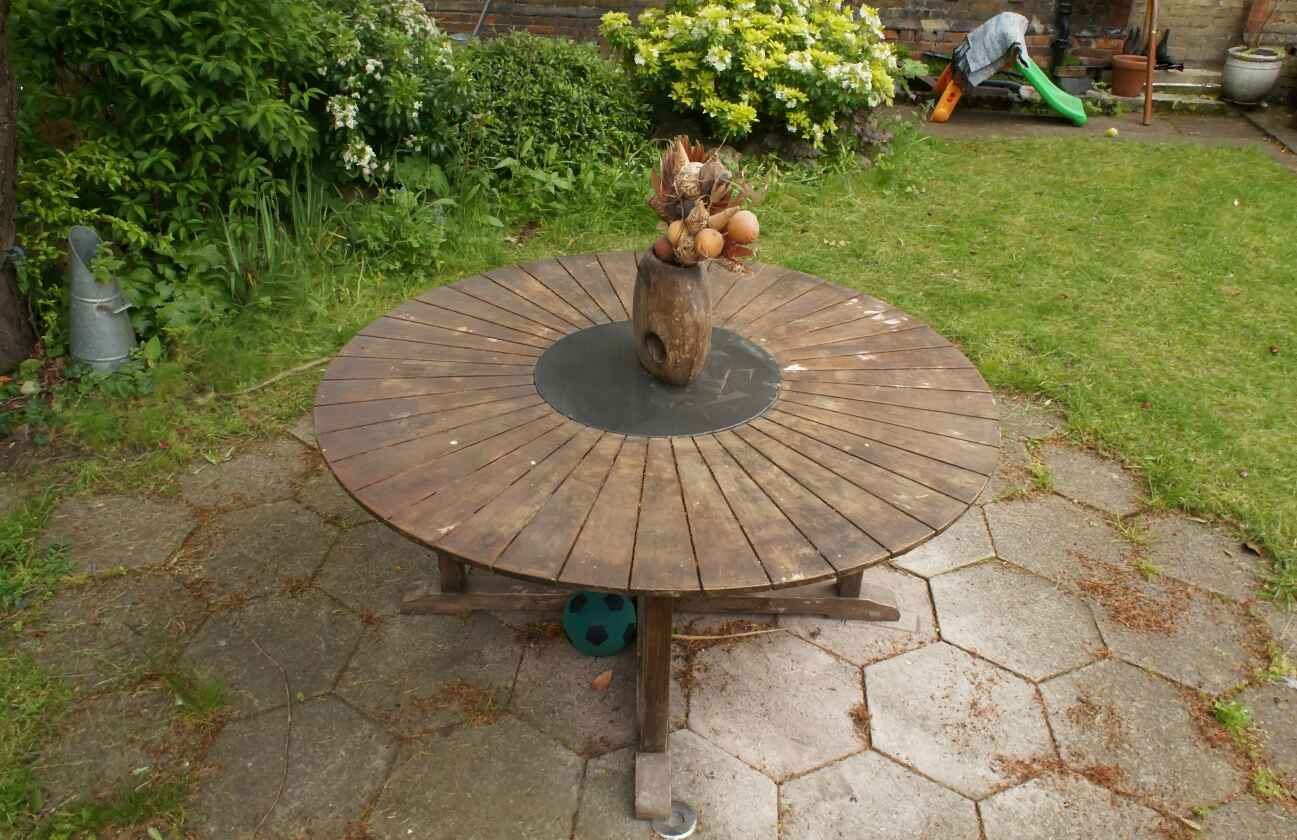}{0.56\mytmplen}{0.48\mytmplen}{0.82\mytmplen}{0.181\mytmplen}{1.2cm}{\mytmplen}{4.5}{red}
   &  \zoomin{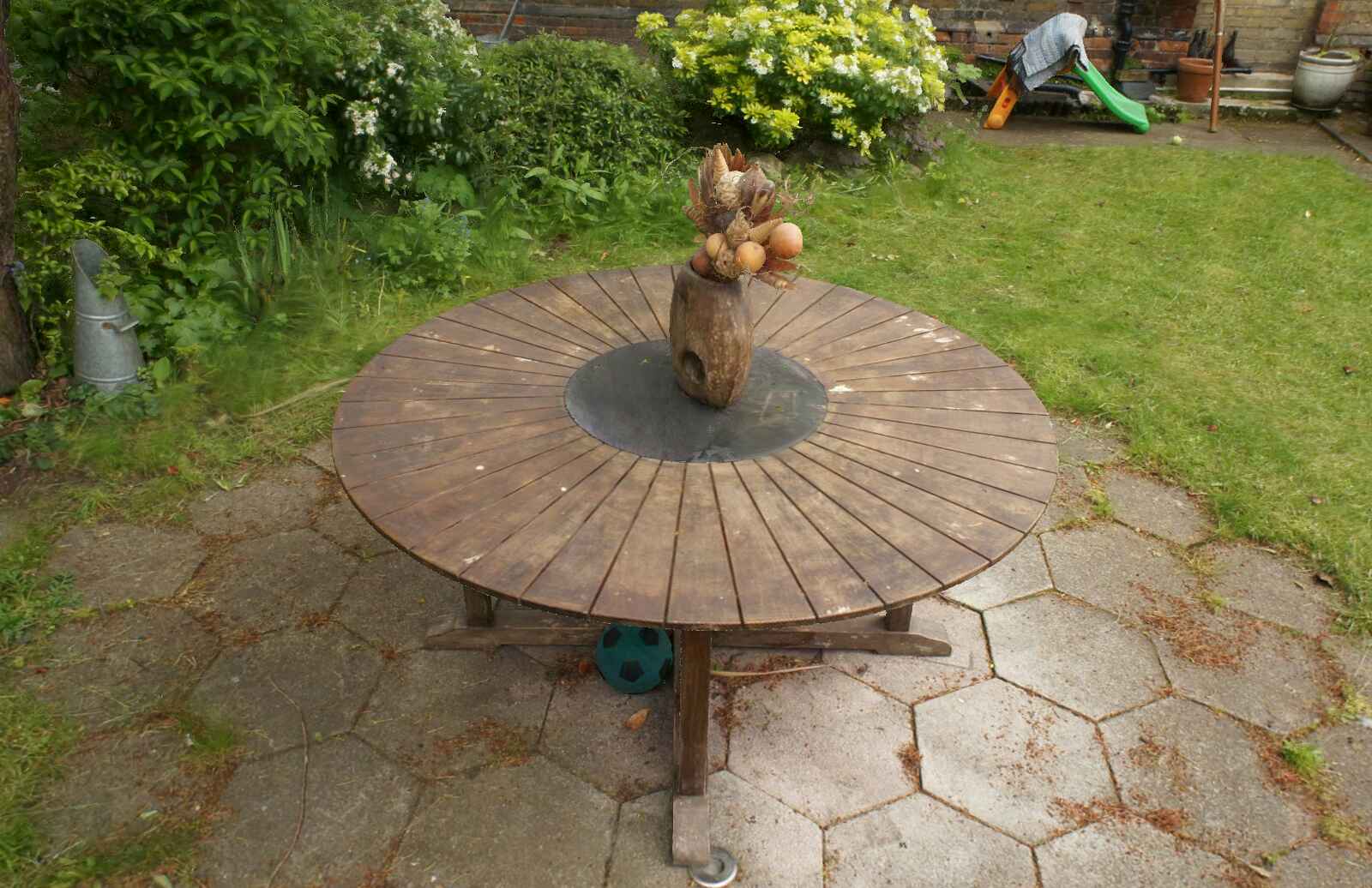}{0.56\mytmplen}{0.48\mytmplen}{0.82\mytmplen}{0.181\mytmplen}{1.2cm}{\mytmplen}{4.5}{red}
   &  \zoomin{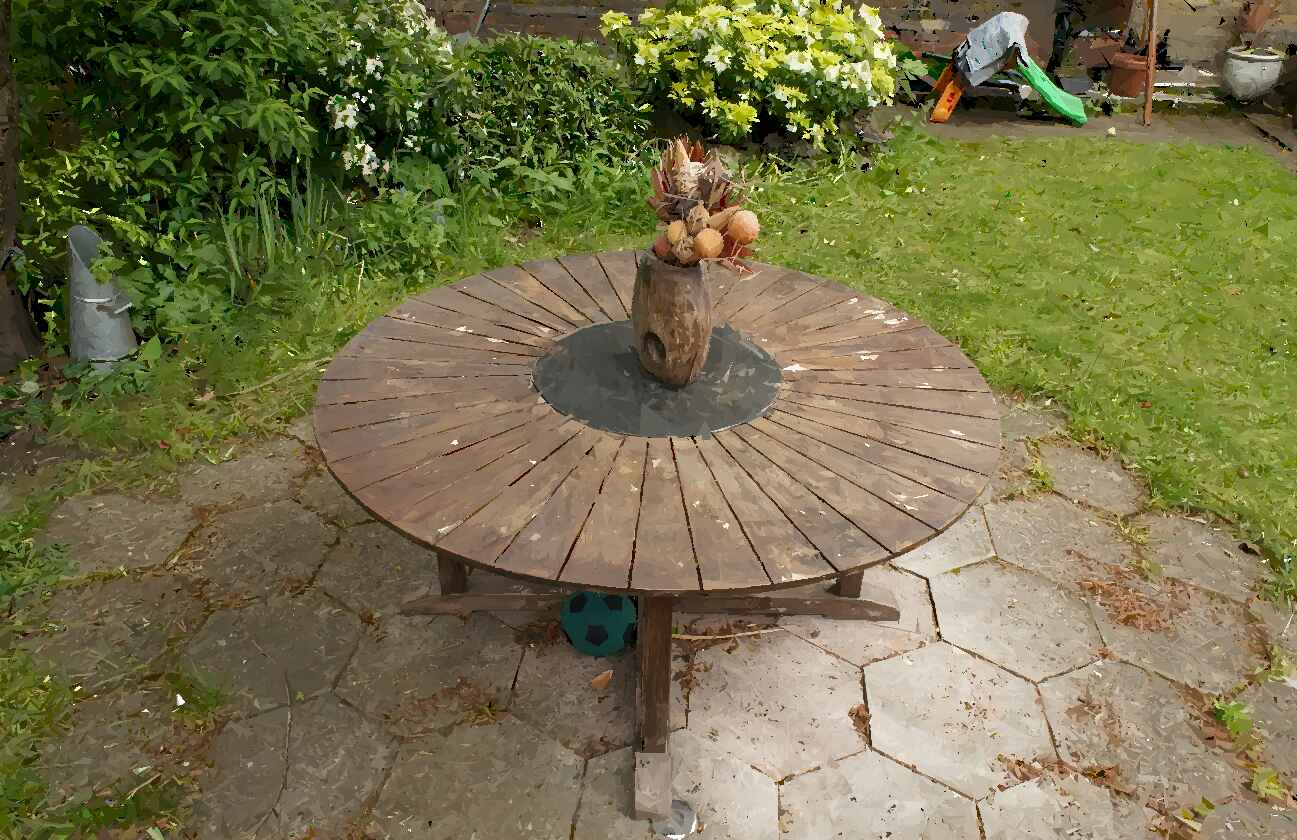}{0.56\mytmplen}{0.48\mytmplen}{0.82\mytmplen}{0.181\mytmplen}{1.2cm}{\mytmplen}{4.5}{red}
    \\

\end{tabular}
}

\caption{\small \textbf{More qualitative results on the \textit{MipNeRf-360}.} 
\methodname is able to reconstruct fine details with only \textit{opaque triangles.}
}
\label{fig:supp_qualityresults_2}
\end{figure*}

\paragraph{Detailed results - \Cref{tab:4a} \& \Cref{tab:mipnerf_outdoor} \& \Cref{tab:mipnerf_indoor} \& \Cref{fig:supp_qualityresults} \& \Cref{fig:supp_qualityresults_2}}
We provide the per-scene results of \methodname to facilitate detailed comparison and reproducibility.
In addition, we provide further qualitative results demonstrating that \methodname produces renderings with fewer artifacts and less noise compared to the current state of the art, MiLo. While MiLo generates high-quality meshes and therefore achieves a strong PSNR on the T\&T dataset, its outputs exhibit more noise, leading to a higher LPIPS and a lower SSIM compared to \methodname.

\paragraph{Surface reconstruction - \Cref{tab:dtu_sum}}
\methodname attains the best Chamfer distance on 5 of the 15 scenes. It achieves a significantly better mean score than Triangle Splatting and 3DGS. Compared to methods specifically tailored for mesh extraction, our results remain competitive and are on par with 2DGS.
Even though our focus is mesh-based novel view synthesis, these results show that \methodname also produces accurate surface meshes.

\begin{table}[ht]
    \centering
    \tabcolsep=0.1cm
    \resizebox{0.8\columnwidth}{!}{      
    \begin{tabular}{l|cccc}
         & 2M & 3M & 4M & 5M \\
        \midrule
        PSNR$\uparrow$  & 22.14 & +0.15 & +0.36 & +0.46 \\
        LPIPS$\downarrow$ & 0.39 & -0.02 & -0.05 & -0.06 \\
    \end{tabular}
    }
    \caption{\textbf{Number of vertices \textit{vs} visual quality.} \methodname scales effectively with the number of vertices, showing consistent improvements in visual quality as the vertex count increases. All improvements are shown relative to 2M.}
    \label{tab:vertices_psnr}
\end{table}

\begin{figure}[t]
\centering

\includegraphics[width=0.90\linewidth]{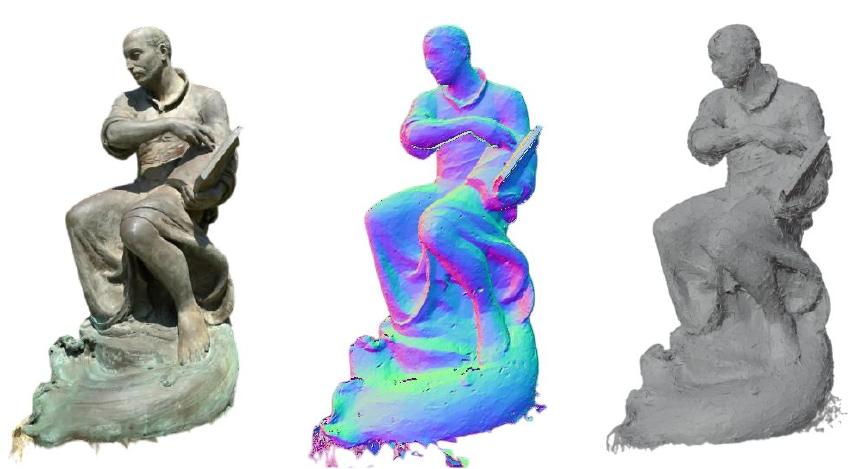}
\caption{\small
\textbf{Object extraction.}
Additional visual examples demonstrating the object extraction capabilities of \methodname.
From left to right: generated RGB image of the object, estimated surface normals, and resulting mesh representation.
}
\label{fig:extraction_supp}
\end{figure}

\paragraph{Impact of triangle count on visual quality - \Cref{tab:vertices_psnr}}
We evaluate the visual quality across different triangle counts on the outdoor scenes of MipNeRF360, as these scenes offer more room for improvement compared to indoor ones. Increasing the number of triangles consistently enhances visual quality, with clear gains in both PSNR and especially in LPIPS.
The results show that \methodname scales well with triangle count, with visual quality improving consistently as more triangles are used.

\begin{table*}[t]
\vspace*{-1em}
\centering
\resizebox{0.98\linewidth}{!}{
\begin{tabular}{l|ccccccccccccccc|c}
Method & 24 & 37 & 40 & 55 & 63 & 65 & 69 & 83 & 97 & 105 & 106 & 110 & 114 & 118 & 122 & Mean \\
\midrule
3DGS \cite{Kerbl20233DGaussian} & 2.14 & 1.53 & 2.08 & 1.68 & 3.49 & 2.21 & 1.43 & 2.07 & 2.22 & 1.75 & 1.79 & 2.55 & 1.53 & 1.52 & 1.50 & 1.96 \\
2DGS \cite{Huang20242DGaussian} & 0.48 & 0.91 & 0.39 & 0.39 & 1.01 & 0.83 & 0.81 & 1.36 & 1.27 & 0.76 & 0.70 & 1.40 & 0.40 & 0.76 & 0.52 & 0.80 \\
GOF \cite{Yu2024Gaussian} & 0.50 & 0.82 & 0.37 & \best 0.37 & 1.12 & \best 0.74 & 0.73 & 1.18 & 1.29 & 0.68 & 0.77 & 0.90 & 0.42 & \best 0.66 & 0.49 & 0.74 \\
RaDe-GS \cite{Zhang2024RaDeGS-arxiv} & 0.46 & 0.73 & \best 0.33 & 0.38 & \best 0.79 & 0.75 & 0.76 & 1.19 & 1.22 & 0.62 & 0.70 & \best 0.78 & \best 0.36 & 0.68 & \best 0.47 & \best 0.68 \\
MiLo \cite{Guedon2025MILo-arxiv} & \best 0.43 & 0.74 & 0.34 & \best 0.37 & 0.80 & \best 0.74 & \best 0.70 & 1.21 & 1.22 & 0.66 & \best 0.62 & 0.80 & 0.37 & 0.76 & 0.48 & \best 0.68 \\
Triangle Splatting \cite{Held2025Triangle-arxiv} & 0.98 & 1.07 & 1.07 & 0.51 & 1.67 & 1.44 & 1.17 & 1.32 & 1.75 & 0.98 & 0.96 & 1.11 & 0.56 & 0.93 & 0.72 & 1.06 \\
\midrule

\textbf{\methodname} & 0.77 & \best 0.72 & 0.74 & 0.60  & 0.89 & 1.00 & 0.81 & \best 1.09 & \best 1.19 & \best 0.58 &  0.68 & 0.93  & 0.63 & \best 0.66 & 0.59 & 0.79 \\
\end{tabular}
}
\caption{\small \textbf{Chamfer distance on the DTU dataset}~\cite{Jensen2014Large}. \methodname achieves performance comparable to concurrent methods.
}
\label{tab:dtu_sum}
\end{table*}

\paragraph{Object extraction - \Cref{fig:extraction_supp}}
Given a 2D mask from SAMv2, we identify all visible triangles contributing to the masked pixels and aggregate them across all training views, yielding the complete set of triangles belonging to the selected object. The entire process takes up to two minutes, depending on the scene and the number of input views. Concretely, we first store the object masks for each training image, then iterate over all views to mark triangles as part of the object whenever they render at least one pixel within the mask. The extracted triangles can then be removed or exported as a standalone submesh without retraining.

\begin{table}[t]
\centering
\resizebox{0.9\linewidth}{!}{%
\begin{tabular}{l|rrr}
& PSNR~$\uparrow$ & LPIPS~$\downarrow$ & SSIM~$\uparrow$ \\
\midrule
Baseline & 24.78 & ~0.31 & ~0.728 \\
\midrule
~~w/o hard pruning  & -0.67 & +0.02 & -0.021 \\
~~w/o stage 2 & -8.56 & +0.25 & -0.260  \\
~~w/o supersampling & -0.80 & +0.04 &  -0.040 \\
~~w/o \textit{w} pruning & -0.62 & +0.05 & 0.045  \\
~~w/o SH & -2.07 & +0.06 & -0.07 \\
~~prune w/o conn. & -0.19 & +0.01 & -0.01 \\
\midrule
~~w/o $\mathcal{L}_{d}$ & +0.05 & -0.04 & +0.006\\
~~w/o $\mathcal{L}_{z}$ & +0.02 & -0.01 & +0.002\\
~~w/o $\mathcal{L}_{n}$ & +0.10 & -0.02 & +0.004\\
\midrule
~~w/o sigma decay & -7.96 & +0.27 & -0.329 \\
~~cosine opacity schedule  & -0.20 & +0.01 & -0.01 \\
~~cosine $\sigma$ schedule & -0.76 & +0.03 & -0.028\\
\end{tabular}%
}
\caption{\small
\textbf{Detailed ablations (Mip-NeRF360).} 
We isolate the impact of each design choice by removing them individually.
}
\label{table:ablation_losses_detailed}
\end{table}

\subsection{Additional ablations - \Cref{table:ablation_losses_detailed}}

\paragraph{Hard pruning step}
An effective pruning strategy is crucial for achieving high visual quality with opaque primitives.
Without the hard pruning step at iteration $5k$, visual quality degrades noticeably. Unnecessary triangles remain in the scene and can no longer be removed.

\paragraph{W/o stage 2}
Running restricted Delaunay triangulation directly on the SfM point cloud produces a connected mesh that is not yet geometrically consistent, leaving many regions, particularly in the background, uncovered.
Training with only connected triangles restricts the degrees of freedom of both vertices and triangles, making optimization considerably harder and leading to a drop in visual quality.
 
\paragraph{W/o supersampling}
To reduce aliasing from opaque triangles, similar to~\citet{Chen2023MobileNeRF}, we render at $s\times$ the target resolution, and then downsample to the final resolution using area interpolation, which averages over input pixel regions to implement a box anti-aliasing filter.
By disabling supersampling during the final training iterations and testing, visual quality decreases. This occurs because the representation uses fully opaque triangles, and metrics such as PSNR penalize small pixel-level shifts. In contrast, when supersampling followed by average downsampling is applied, transitions between triangles become smoother, leading to more continuous transitions and higher visual quality. We applied anti-aliased rendering for both training and testing. 

\paragraph{W/o \textit{w} pruning}
When pruning is based only on $opacity$ rather than the maximum blending weight $w$, many unnecessary triangles remain in the scene. As these triangles become opaque, they introduce artifacts and reduce visual quality.

\paragraph{W/o SH} When training with only RGB colors, the visual quality drops significantly. The opaque and connected triangles struggle to accurately reproduce fine texture details. Real-world scenes exhibit complex spatial variations in color and illumination, which would require an impractically large number of triangles to model using RGB alone. Incorporating spherical harmonics enables each triangle to capture part of this variation, resulting in a noticeably improved appearance.
This observation suggests that a more expressive appearance model could further enhance visual fidelity. Future work could explore learning per-triangle textures either jointly during training or as a post-processing refinement, as done in recent mesh-based methods. Such an approach could further narrow the visual quality gap between 3D Gaussian-based representations and fully opaque triangle-based ones.

\paragraph{Pruning w/o connectivity.}
During the \textit{first stage}, we exploit triangle connectivity by applying midpoint subdivision while ensuring that all four newly created triangles remain connected. This approach reduces the number of introduced vertices by half: subdividing each triangle independently would yield 12 vertices, whereas maintaining shared connectivity requires only 6. By connecting triangles, we can allocate a larger portion of the memory budget to rendering more vertices and triangles, which leads to improved visual quality.

\begin{figure}[t]
\centering
\setlength{\tabcolsep}{2pt}
\begin{tabular}{cc}
\includegraphics[width=0.4\linewidth]{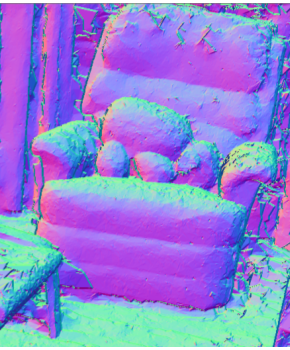} &
\includegraphics[width=0.4\linewidth]{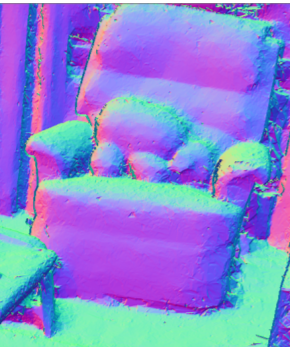} \\
\scriptsize (a) Without $\mathcal{L}_{z}$ &
\scriptsize (b) With $\mathcal{L}_{z}$ \\
\end{tabular}
\caption{\small
\textbf{Impact of the $\mathcal{L}_{z}$ regularization term.}
With the regularization we obtain smoother surfaces.
}
\label{fig:loss_L_vertex}
\end{figure}

\paragraph{W/o $\mathcal{L}_{z}$ - \Cref{fig:loss_L_vertex}}
Even with normal regularization, some triangles remain misoriented. 
By pushing the vertices closer to the surface, we improve both mesh smoothness and normal consistency. The vertex regularization term $\mathcal{L}_{\text{vertex}}$ reduces artifacts and promotes smoother surfaces, particularly in background regions where supervision is typically weaker.

\paragraph{Cosine scheduler}
Compared to a linear schedule for increasing $opacity$ and decreasing $\sigma$, we also experimented with a cosine scheduler. However, performance, particularly for $\sigma$, drops noticeably. The main reason is that the linear schedule maintains higher $\sigma$ values for longer, allowing gradients to remain stable during the final iterations, whereas the cosine scheduler reaches low $\sigma$ values earlier, causing gradients to vanish prematurely.
Finally, we also analyze the impact of initialization. Instead of starting from soft triangles (i.e., $\sigma = 1.0$) that gradually converge toward solid ones, we evaluate the case where training begins directly with fully solid triangles. In this configuration, gradients can only flow through the opacity term, causing optimization to stagnate. As a result, both performance and visual quality degrade noticeably due to severe vanishing gradients.

\subsection{Mesh connectivity.}
The resulting mesh exhibits a \textit{vertex--face ratio} of $0.48$, which is close to the theoretical $0.5$ expected for a closed manifold surface, indicating that the global mesh topology is already compact and near-manifold despite local non-manifold regions remaining.
For reference, a completely isolated mesh with no vertices-sharing, would have a ratio of $1.5$.
We further analyze vertex connectivity by measuring the number of incident triangles (vertex valence) across the reconstructed mesh.
The mean valence is $6.2$ with a median of $5$, which aligns with the expected value of~$6$ for regular triangular tessellations but with a broader distribution.
Roughly $35\%$ of vertices exhibit low valence ($\leq4$), indicating remaining boundary regions and small disconnected fragments, while about $30\%$ fall within the regular interior range ($5$–$8$).
A small fraction ($<10\%$) shows high valence ($>12$), corresponding to locally dense or merged zones.
This distribution shows that while the mesh is largely connected and compact, it remains not fully watertight.

\begin{figure}[t]
\setlength\mytmplen{0.48\linewidth}
\centering
\zoomin{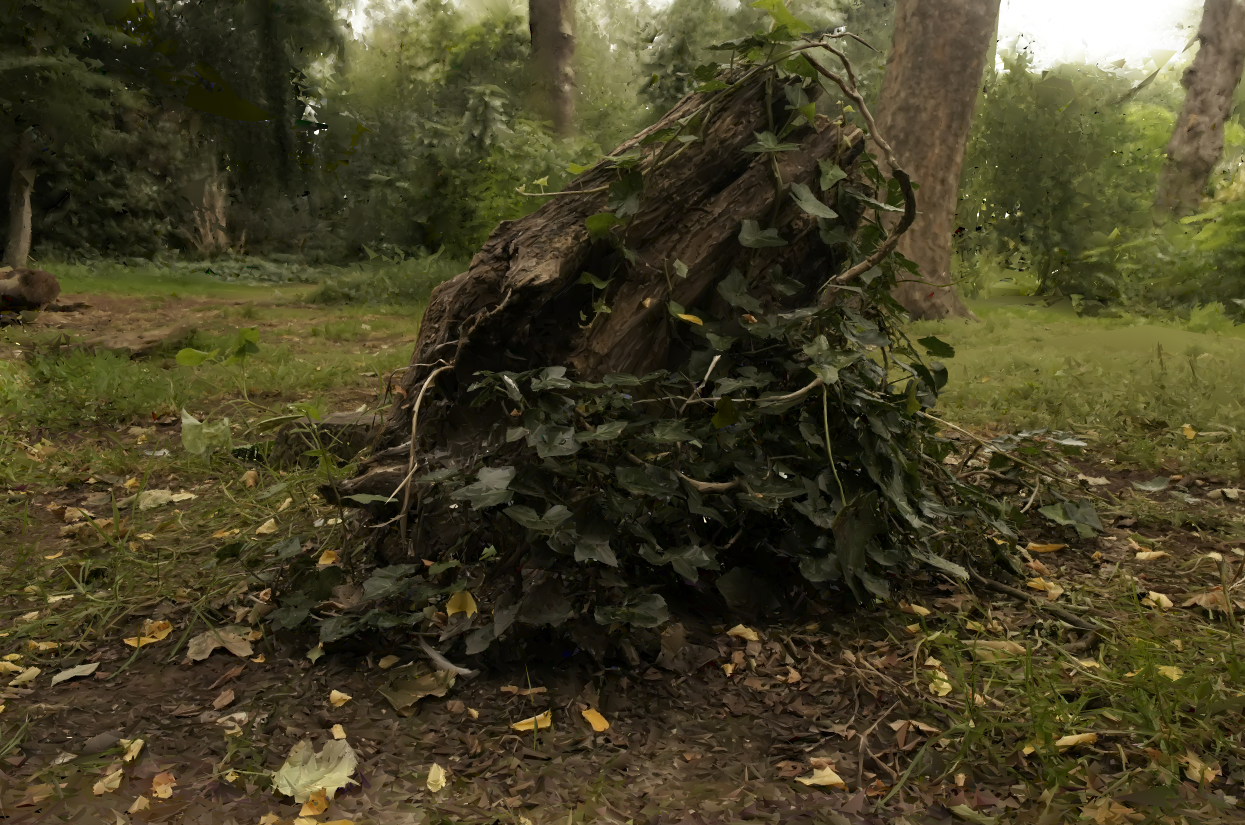}{0.3\mytmplen}{0.6\mytmplen}{0.81\mytmplen}{0.19\mytmplen}{1.5cm}{\mytmplen}{3.5}{red} \hfill
\zoomin{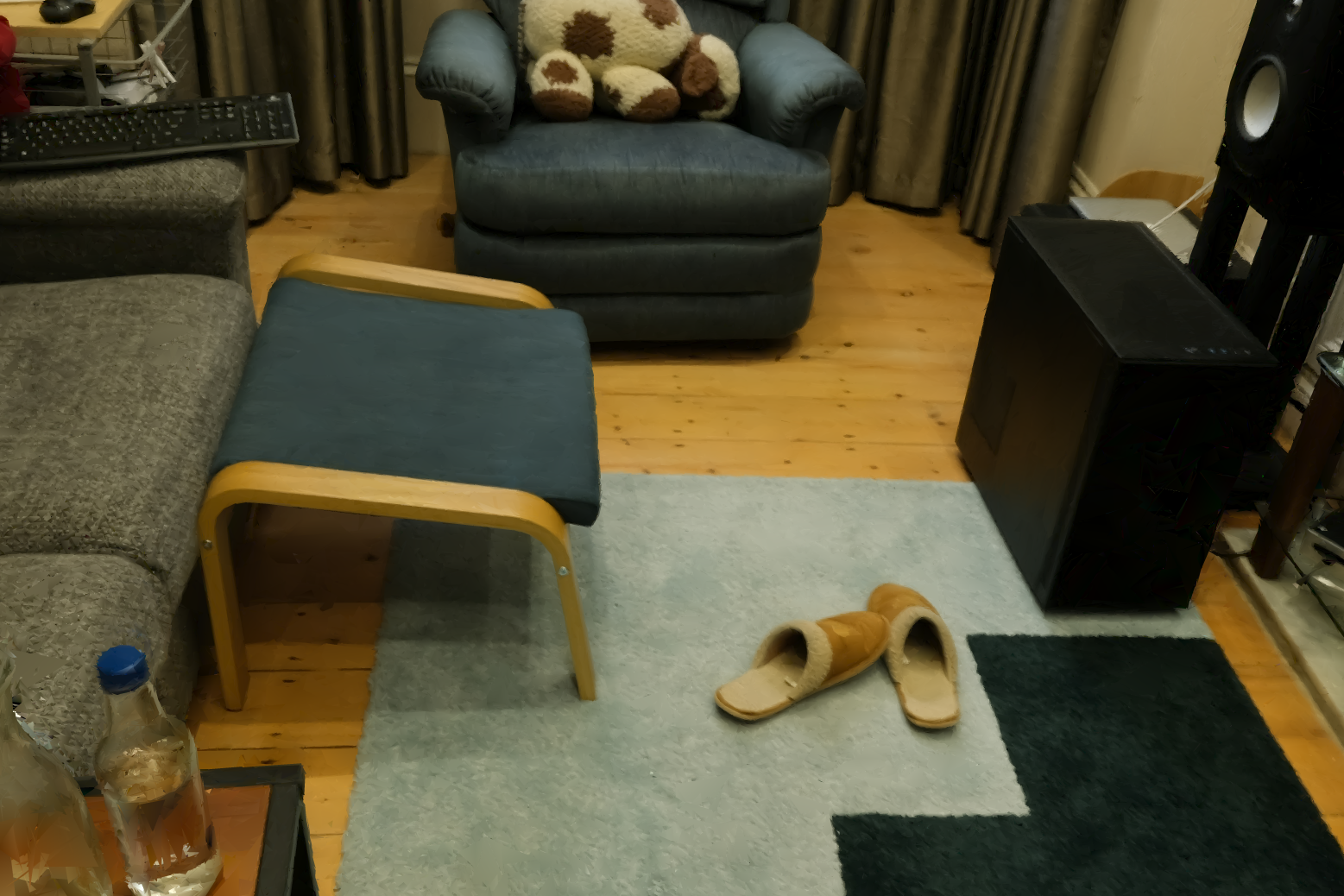}{0.1\mytmplen}{0.1\mytmplen}{0.81\mytmplen}{0.19\mytmplen}{1.5cm}{\mytmplen}{3.5}{red} 
\captionof{figure}{\small
\textbf{Limitations.} Accurately recovering backgrounds (left), particularly under limited viewpoints, and handling transparent objects (right) remain challenging.
\label{fig:limitations}
}
\end{figure}

\subsection{Limitations}

\paragraph{Limitations}
\methodname achieves high visual quality and accurate reconstruction in regions where the initial point cloud is dense. In contrast, background areas with sparse coverage still exhibit incomplete geometry and reduced fidelity. 
Moreover, when moving outside the orbit of training views, the visual quality degrades. While the softness and opacity of Gaussian-based approaches may still provide slightly plausible results in such cases, our use of opaque triangles makes the artifacts more pronounced.
Future work could address those limitation by initializing with a more complete point cloud, or by incorporating alternative additional representation such as a triangulated sky dome.
Furthermore, transparent objects such as glasses or bottles remain difficult to represent using only opaque triangles, as illustrated in \cref{fig:limitations}.
Finally, \methodname does not explicitly enforce watertightness; however, the resulting meshes can be directly used in many downstream applications, offering an effective trade-off between simplicity, usability, and high visual quality. Future work could incorporate additional regularization terms to constrain vertex motion and prevent self-intersections or overlaps.

\end{document}